\documentclass{egpubl}

 \JournalSubmission    
 \electronicVersion 

\ifpdf \usepackage[pdftex]{graphicx} \pdfcompresslevel=9
\else \usepackage[dvips]{graphicx} \fi

\PrintedOrElectronic

\usepackage{t1enc,dfadobe}
\usepackage{mathptmx}   
\usepackage{amsmath}   
\usepackage[ruled,lined,boxed,commentsnumbered]{algorithm2e} 
\usepackage{subfigure}   
\usepackage{graphicx}   
\usepackage{epstopdf}   
\usepackage{egweblnk}
\usepackage{cite}
\usepackage{amsfonts}
\usepackage{color}

\DeclareMathOperator*{\argmin}{arg\,min}

\title%
      {Automatic Semantic Style Transfer using Deep Convolutional Neural Networks and Soft Masks}

\author[Huihuang Zhao \& Paul L. Rosin \& Yu-Kun Lai]
{\parbox{\textwidth}{\centering Huihuang Zhao\thanks{happyday.huihuang@gmail.com}$^{1}$, Paul L. Rosin$^{2}$ and Yu-Kun Lai$^{2}$
        }
        \\
{\parbox{\textwidth}{\centering $^1$School of Computer Science and Technology, Hengyang Normal University, Hengyang, Hunan, China\\
        $^2$ School of Computer Science and Informatics, Cardiff University, Cardiff, UK
      }
}
}

\begin{document}

\maketitle
\begin{abstract}
This paper presents an automatic image synthesis method to transfer the style of an example image to a content image.
When standard neural style transfer approaches are used, the textures and colours in different semantic regions of the style image are often applied inappropriately to the content image, ignoring its semantic layout, and ruining the transfer result. In order to reduce or avoid such effects, we propose a novel method based on automatically segmenting the objects and extracting their soft semantic masks from the style and content images, in order to preserve the structure of the content image while having the style transferred. Each soft mask of the style image represents a specific part of the style image, corresponding to the soft mask of the content image with the same semantics. Both the soft masks and source images are provided as multichannel input to an augmented deep CNN framework for style transfer which incorporates a generative Markov random field (MRF) model. Results on various images show that our method outperforms the most recent techniques.


\end{abstract}  
\section{Introduction}

Style transfer is a process of migrating a style from a ``style image'' to a ``content image''. The goal is to be able to generate different renditions of the same scene according to different style images. Image style transfer has become a popular problem in computer vision and graphics, and can generate impressive results covering a wide variety of styles for both images~\cite{gatys2015neural}
and videos~\cite{ruder2016artistic}.
It has also been widely employed to solve problems such as texture synthesis \cite{efros2001image}, inpainting \cite{criminisi2004region}, head portraits \cite{selim2016painting} and super-resolution \cite{johnson2016perceptual}.

When existing neural style transfer methods are applied to images with complex structures, visual elements from the style image are often transferred to semantically irrelevant areas of the content image. 
In order to achieve good results, users must pay attention to the composition and/or the selection of the style image, because for example the background colours or textures will often ruin the style transfer results,
especially for portraits where the artifacts can be particularly off-putting. Addressing this problem, \cite{champandard2016semantic} (and subsequently~\cite{gatys2016controlling}) recently proposed a method which uses a manually generated semantic map to help control the style transfer, and can achieve better results than some common methods.

\begin{figure}
\centering
    \includegraphics[width=0.23 \linewidth]{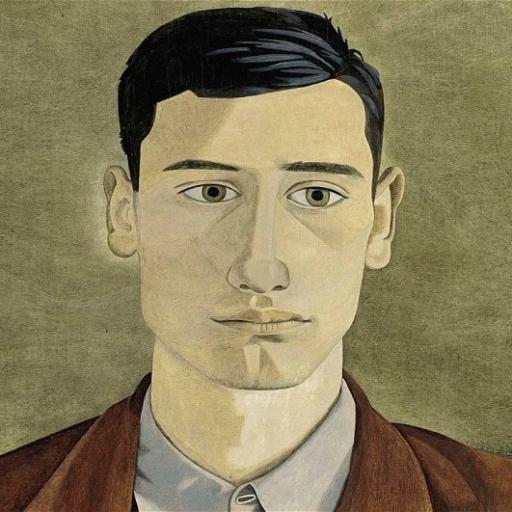}
   \includegraphics[width=0.23 \linewidth]{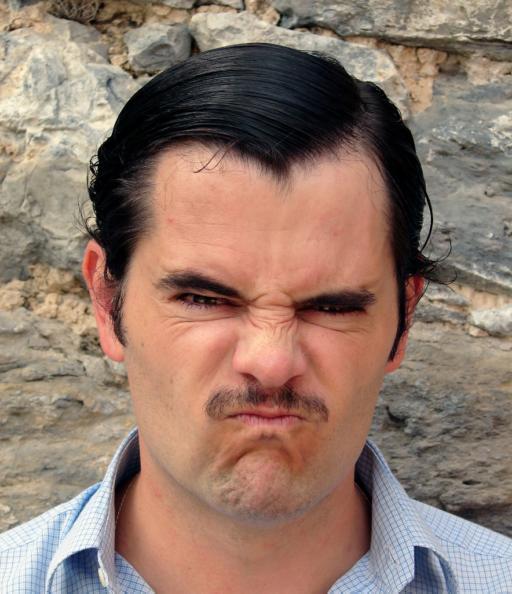}
   \includegraphics[width=0.23 \linewidth]{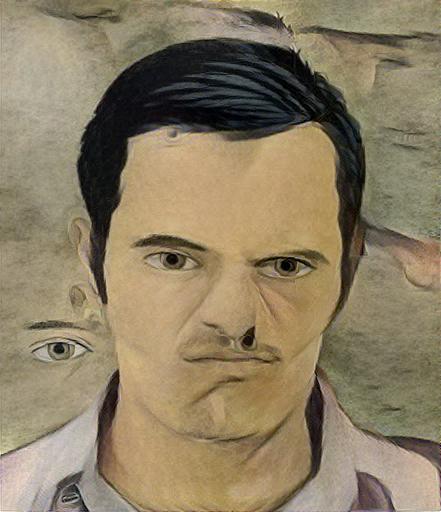}
   \includegraphics[width=0.23 \linewidth]{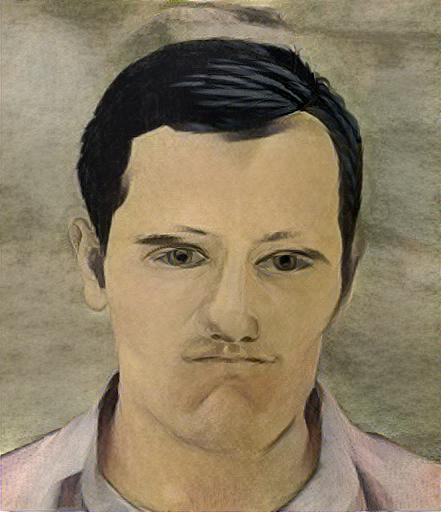}
  \\   
    \subfigure[style]{\includegraphics[width=0.23 \linewidth]{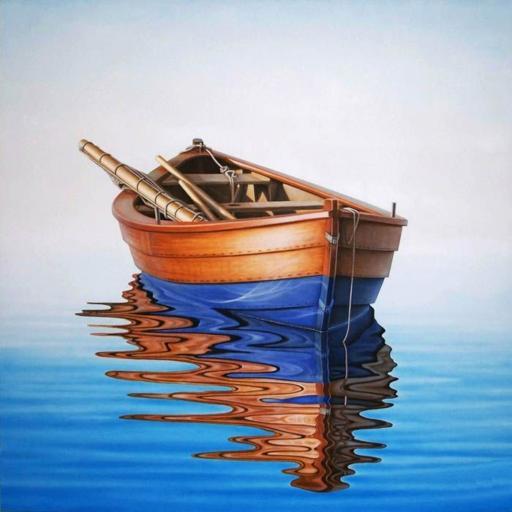}} 
    \subfigure[content]{\includegraphics[width=0.23 \linewidth]{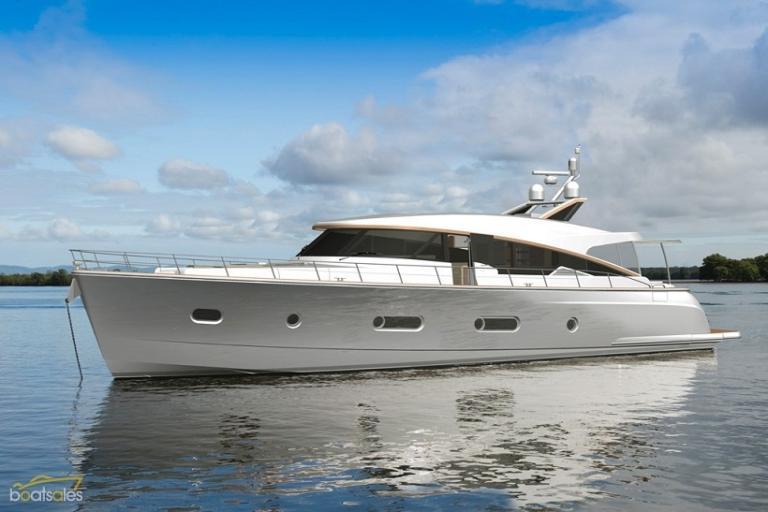}}    
    \subfigure[CNNMRF]{\includegraphics[width=0.23 \linewidth]{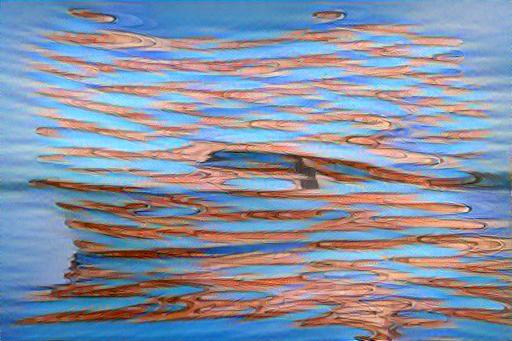}} 
    \subfigure[our method]{\includegraphics[width=0.23 \linewidth]{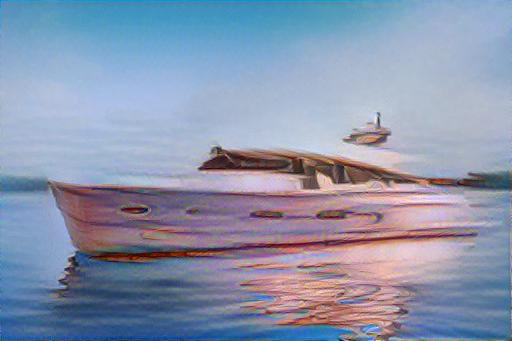}}    
    \caption{Automatic semantic style transfer using deep convolutional neural networks.}
    \label{fig:top_figure} 
\end{figure}

In this paper, we specifically consider the problem of image style transfer guided by \emph{automatically} extracted \emph{soft} semantic masks. 
To achieve this, we adapt various semantic segmentation and labelling techniques to extract soft masks associated with specific semantics. 
By deploying the semantic masks to control the transfer, it is possible to avoid errors such as those shown in
figure~\ref{fig:top_figure}(c) generated using the CNNMRF method~\cite{li2016combining} in which stylised foreground objects are
contaminated by the background texture, and vice versa.
%
%

The main contributions of the paper are as follows:

We adapt a state-of-the-art semantic segmentation method~\cite{zheng2015conditional} to generate semantic masks automatically. Instead of using hard segmentation as~\cite{zheng2015conditional}, we propose to use soft masks containing the probabilities of occurrence of different objects in the image, since they preserve more information and is more robust when image regions have similar chances of belonging to multiple object categories. They are used to capture elements of the styles for objects in the style image and to preserve the structure of the content image. For the human face in particular we use a more detailed segmentation, in which different facial parts such as the nose, eyes and mouth are also automatically segmented, providing fine-grained control in perceptually crucial areas; these are also treated as semantic masks.

We augment a trained deep convolutional neural network by concatenating $K$ soft mask channels and $N$ channels of regular filters. This is further combined with a generative Markov random field (MRF) model~\cite{li2016combining} for image style transfer. Both the style and content images and their semantic maps are input into the augmented deep convolutional neural network. Extensive experiments show that such higher-level semantic information improves the quality of style transfer.

\section{Related Work}

\textbf{Style transfer using deep networks.} The success of deep CNNs (DCNNs) in image processing has also raised interest in image style transfer. \cite{shih2014style} proposed a new style transfer method for headshot portraits. During their method, they presented a new multiscale technique based on deep networks to robustly transfer the local statistics of an example portrait onto a new one. \cite{gatys2015neural,gatys2015texture} showed remarkable results by using the VGG-19 deep neural network for style transfer. Their approach was employed in unguided settings and taken up by various follow-up papers. \cite{gatys2016controlling} in particular extended the Gram matrix method beyond the paradigm of transferring global style information between pairs of images, and they introduced control over spatial location, colour information and spatial scale.
\cite{ulyanov2016texture} presented an alternative approach which trained compact feed-forward convolutional networks. The resulting networks are extremely light-weight and can generate images faster than \cite{gatys2015neural}. By combining the benefits of training feedforward convolutional neural networks and perceptual loss functions, \cite{johnson2016perceptual} presented a novel approach for image style transfer. \cite{li2016combining} suggested an approach to preserving local patterns of the style image. Instead of using a global representation of the style computed as a Gram matrix, they used patches of the neural activation from the style image. \cite{ruder2016artistic} presented an approach that transfers the style from one image (for example, a painting) to a whole video sequence. 

%
Two main types of methods are used in deep learning based style transfer: global approaches based on the Gram matrix, and local approaches based on patch matching. Compared to the global methods, methods based on patch matching are more flexible and better cope with images with spatial variation of visual styles or elements. However, they could also produce visible artefacts when there are local matching errors. 
In order to control the region of application of the style image, \cite{gatys2016controlling} used several manually specified spatial guidance channels, containing values in [0,1], for both the content and style images.
Their experiments showed that the guidance channels can ensure that the style is transferred between regions of similar scene content in the content and style images. 
It is however time-consuming to produce masks. As a result, for examples in their paper, they just used a mask to separate two parts of the image (e.g. sky and non-sky) for simple spatial control, and did not distinguish more detailed content in the images. 

\textbf{MRF-based image synthesis.} 
Markov Random Fields (MRFs) are a famous framework for non-parametric image synthesis \cite{efros1999texture}, \cite{freeman2000learning}. \cite{kwatra2003graphcut}, \cite{kwatra2003graphcut} and \cite{kwatra2005texture} modelled the texture as an MRF and computed some approximation to the optimal solution. \cite{zhang2013style} formulated the patch mapping problem as a labelling problem modelled by a discrete MRF. Moreover, \cite{frigo2016split} proposed a novel unsupervised method for texture and colour transfer based on MRFs. In their approach an adaptive patch partition is used to capture the style of the example image and preserve the structure of the source image. MRF models suffer from a limitation that local image statistics are usually not sufficient for capturing complex image layouts at a global scale. \cite{wei2000fast} and \cite{kwatra2005texture} proposed a multi-resolution synthesis approach to improve this.
We adapt this in our method. 
\cite{li2016combining} presented a combination of generative Markov random field (MRF) models for image synthesis.
Unlike other MRF-based texture synthesis approaches, their combined system can both match and adapt local features with considerable variability, and therefore our paper is based on this method.

\textbf{Semantic segmentation.}
Recently, CNN architectures have been shown to be capable of providing semantic segmentation \cite{girshick2014rich,thoma2016survey}.
\cite{girshick2014rich} proposed a method called R-CNN, which combined region proposals with CNNs.
\cite{noh2015learning} applied a trained network (VGG 16-layer net) to each proposal in an input image, and constructed the final semantic segmentation map by combining the results from all the proposals. \cite{shelhamer2017fully} proposed a fully convolutional network for semantic segmentation. For producing accurate and detailed segmentations, they defined a skip architecture which combines semantic information from a deep, coarse layer with appearance information from a shallow, fine layer. In order to achieve better results, some existing face detection methods are also used in style transfer. By searching a database using Face++ \cite{faceplusplus.com} to find images with similar poses to a given source portrait image, \cite{yang2017semantic} presented a novel colour transfer approach for portraits. \cite{zheng2015conditional} introduced a new form of convolutional neural network that combines the strengths of Convolutional Neural Networks and Conditional Random Fields based probabilistic graphical modelling. These models rely primarily on convolutional layers to extract high-level patterns, then use deconvolution to label the individual pixels. Currently they have trained this model to recognise 20 classes, and our paper uses this method to obtain some semantic content from images.

\textbf{Limitations of current methods.}
%
Images are typically composed of regions corresponding to different (foreground) objects and background. Most existing methods either use Gram matrices which treat images globally, or for methods based on local patch matching, can often match regions of one object in the style image to regions of a different object in the content image, causing artefacts such as those shown in figure~\ref{fig:top_figure}. This is more critical for human faces as subtle mismatches can be detrimental to the quality of synthesised images. 
To address this, existing methods~\cite{champandard2016semantic,gatys2016controlling} use manual segmentation to improve style transfer. However, manual segmentation is time-consuming and laborious. 
In contrast, our method automatically performs a partial soft semantic segmentation of the content and style images. We augment the CNNMRF model used in~\cite{li2016combining} to further incorporate soft semantic masks, which can better capture features from the style image and preserve the structure of the content image.

We first briefly introduce our augmented DCNN architecture in section~\ref{sec:arch}, followed by details for the style transfer algorithm in section~\ref{sec:alg}. We then provide details for automatic semantic mask extraction in section~\ref{sec:masks}. Experimental results and discussions are presented in section~\ref{sec:results} and finally conclusions are drawn in section~\ref{sec:conc}.

 
\section{Architecture}\label{sec:arch}

\begin{figure*}[tb]
  \center
\includegraphics[width=0.80\linewidth]{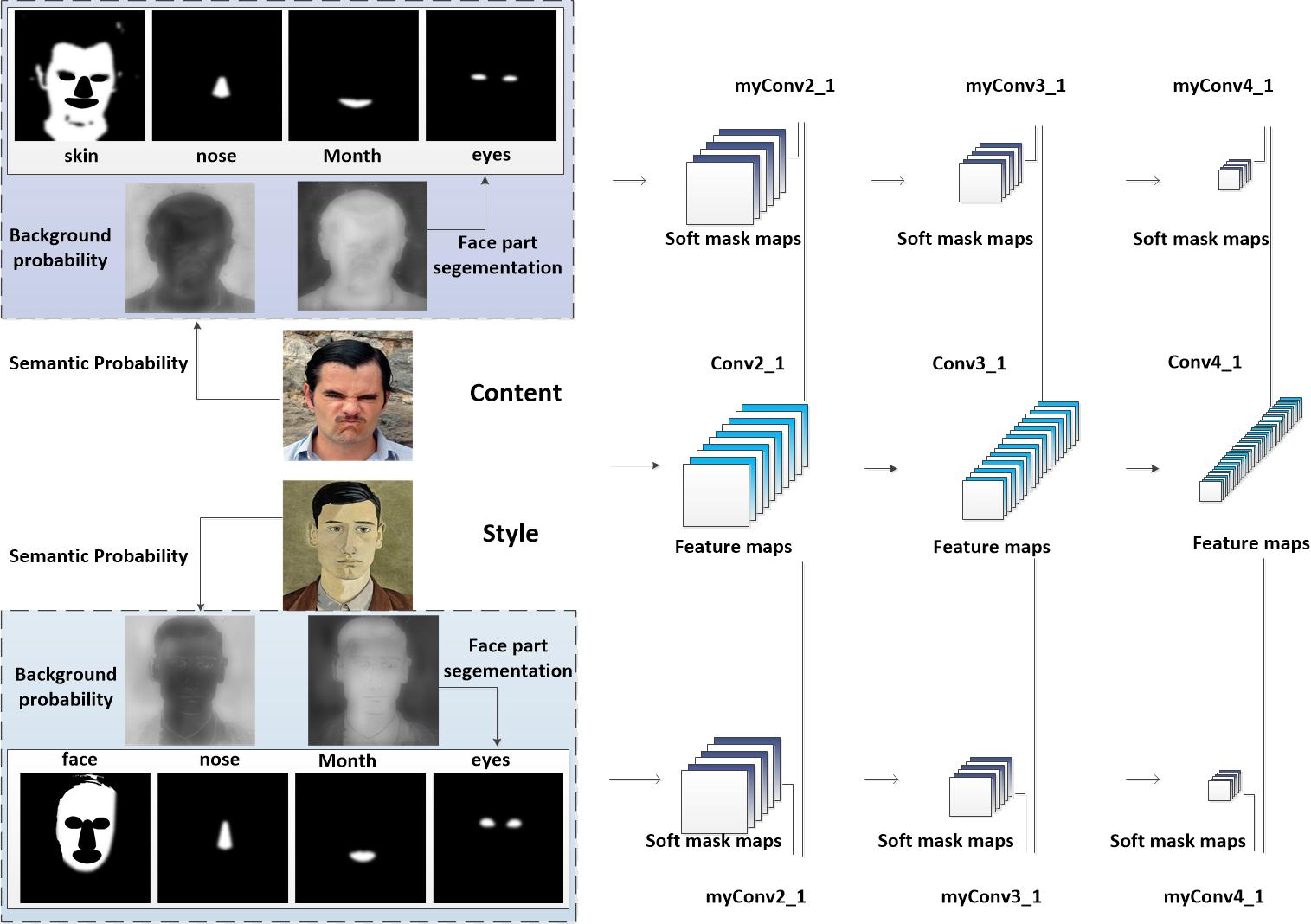}
  \caption{Style transfer framework with deep neural networks and soft masks.}
  \label{fig:framework}
\end{figure*}

We now discuss our augmented DCNN architecture which is based on VGG \cite{simonyan2014very} for style transfer. It takes as input a content image and a style image, both of which are fed into the VGG net.
The DCNN architecture combines pooling and convolution layers $l$ with $3 \times 3$ filters (for example, the first layer after second pooling is named $Conv3\_1$). Like common DCNNs, intermediate post-activation results denoted as $x^l$ for the layer $l$ consist of $N$ channels, which capture patterns from the source images for each region of the image. Our augmented network is shown in figure~\ref{fig:framework}.

Our augmented network also takes $K$ semantic soft masks as input, which are down-sampled to produce semantic channels $p^l$ at layer $l$ with the same resolution as $x^l$.  
We concatenate them to form the new output with $N + K$ channels, defined as $m^l$ and labeled accordingly for each layer (e.g. $mycov4\_1$). Before concatenation, the semantic channels are weighted by parameter $\beta$ to balance their importance:
\begin{equation}
\label{mrf_eq00}
m^l= (x^l, \beta p^l).
\end{equation}

We set $\beta=20$ which we have found experimentally to provide interesting results.

\section{Semantic Style Transfer Algorithm}\label{sec:alg}
\subsection{Optimisation function}



Next, we introduce our style transfer model. We use an augmented loss function which is based on a patch-based approach \cite{li2016combining} for style transfer, using optimisation to minimise content reconstruction error $E_c$ and style remapping error $E_s$, which combines an MRF and a DCNN model,
given a style image $x_s \in \mathbb{R}^{w_s \times h_s} $, a content image
$x_c \in \mathbb{R}^{w_c \times h_c}$, and semantic maps $m_{c_k} \in \mathbb{R}^{w_c \times h_c}$ and $m_{s_k} \in \mathbb{R}^{w_c \times h_c}$ associated with the content and style images, respectively ($k = 1, 2, \dots, K$). 
For simplicity, the semantic masks for the content and style images are also collectively represented as $m_c \in \mathbb{R}^{w_c \times h_c \times K}$ and $m_s \in \mathbb{R}^{w_s \times h_s \times K}$. 
The style transfer result image is denoted by $x \in \mathbb{R}^{w_c \times h_c}$. 
Since the synthesised image $x$ is expected to have the same semantic layout as the content image, we treat $m_{c}$ also as the semantic masks for the synthesised image.
During our method, we make the high-level neural encoding of $x$ similar to $x_c$ and use the local patches similar to patches in $x_s$. As a result, the style of $x_s$ is transferred onto the layout of $x_c$. 
Meanwhile, we penalise patch matches with inconsistent semantic masks. 
We define an energy function as follows and seek $x$ that minimises it:
\begin{equation}
\label{mrf_eq01}
E(x) = \alpha_1 E_s(\Phi(x),\Phi(x_s),\Phi(m_{c}), \Phi(m_{s}))+ \\ \alpha_2 E_c(\Phi(x),\Phi(x_c)).
\end{equation}
$E_s$ and $E_c$ are defined as the style loss function and content loss function respectively, where $\Phi(x)$ is $x$'s feature map (activation) that the network outputs in some layer, $\Phi(x_s)$ is the feature map (activation) of the style image $x_s$ in the same layer, and $\Phi(m_{c})$ and $\Phi(m_{s})$ are the semantic masks of the content and style images downsampled to the same resolution as $\Phi(x)$ and $\Phi(x_s)$.   
For our method, $E_s$ aims to penalise inconsistencies in neural activations and/or semantic masks between $x$ and $x_s$.
$E_c$ computes the squared distance between the feature map of the synthesised image and that of the content source image $x_c$. Since $x$ is assumed to have the same content layout as $x_c$, $E_c$ does not involve the semantic masks. 

\textbf{Style loss function:} We extract all the local patches from $ \Phi(x)$, denoted as $ \Psi(\Phi(x)) $. For a given layer, assuming $N$ is the number of channels, each patch in $\Psi_i(\Phi(x)) $ has size $t \times t \times N$, where $t$ is the width and height of the patch. Similarly, $\tilde\Psi (\Phi(m_{c_k}))$ and $\tilde\Psi (\Phi(m_{s_k}))$ are the down-sampled semantic masks of extracted patches, each of size $t \times t$. 
We define the modified energy function $E_s$ incorporating semantic masks as
\begin{equation}
\label{mrf_eq02}
\begin{split}
E_s(\Phi(x),\Phi(x_s),\Phi(m_s))=\sum_{i=1}^{P}{\parallel \Psi_i(\Phi(x))- \Psi_{NN(i)}(\Phi(x_s))\parallel^2 }\\ + 
 \beta\sum_{i=1}^{P}{\sum_{k=1}^{K}{\parallel \tilde\Psi_i(\Phi(m_{c_k}))- \tilde\Psi_{NN(i)}(\Phi(m_{s_k}))\parallel^2 }},
\end{split}
\end{equation}
where $P = |\Psi(\Phi(x))|$ is the number of patches in the synthesised image. For each patch $\Psi_i(\Phi(x))$ with semantic masks $\tilde\Psi_i(\Phi(m_{c_k}))$ we find its best matching patch $\Psi_{NN(i)}(\Phi(m_s))$ using normalised cross-correlation over all $P_s$ example patches in $\Psi(\Phi(x_s))$:
\begin{equation}
\label{mrf_eq03}
\centering
\begin{split}
NN(i):=\argmin\limits_{j=1,\dots,P_s}{\dfrac{\Psi_i^*(\Phi(x))\cdot\Psi_j^*(\Phi(x_s))}{\mid \Psi_i^*(\Phi(x))\mid \cdot \mid\Psi_j^*(\Phi(x_s))\mid}},
\end{split}
\end{equation} 
where $\Psi_i^*(\Phi(x)) = \left( \Psi_i(\Phi(x)), \beta \tilde \Psi_i (\Phi(m_{c})) \right)$ is the concatenation of neural activation and semantic masks for the $i^{\textrm {th}}$ patch of the synthesised image, and $\Psi_j^*(\Phi(x_s)) = \left( \Psi_j(\Phi(x_s)), \beta \tilde \Psi_j (\Phi(m_{s})) \right)$ is the concatenation of neural activation and semantic masks for the $j^{\textrm {th}}$ patch of the style image. The nearest patch thus takes both style similarity and semantic consistency into account.

\textbf{Content loss function:} In order to control the content of the synthesised image, we define $E_c$ as the squared Euclidean distance between $\Phi(x)$ and $\Phi(x_c)$:
\begin{equation}
\label{mrf_eq04}
E_c(\Phi(x),\Phi(x_c),\Phi(m_c))=\| (\Phi(x)) - \Phi(x_c) \|^2.
\end{equation}

Like method~\cite{li2016combining}, we also minimise Equation~\ref{mrf_eq01} using backpropagation with L-BFGS. During Equation~\ref{mrf_eq01}, $ \alpha_1$ and $ \alpha_2$ are weights for the style image and the content image constraints, respectively. According to our experiments, we set $\alpha_1=10^{-4}$ and $\alpha_2=20$, and these values can be fine tuned to interpolate between the content and the style preservation.
 
\section{Automatic Soft Semantic Mask Extraction}\label{sec:masks}
\cite{champandard2016semantic} manually generated the semantic masks that they used in their work to control the style transfer. Each image used one mask containing semantic labels, where each component (not necessarily connected) was indicated by a particular pixel value in the image. Often these values were carefully chosen so that components with similar appearance such as ear and nose would be assigned similar mask values.
Not only is it tedious to manually segment the image, but for most images some parts cannot be partitioned accurately. Therefore, instead of using a single crisp mask to control an image stylization, we propose instead to use a set of soft masks. Such soft masks provide more information than a single crisp mask, and do not require potentially unreliable boundaries to be set in the semantic mask, which is especially beneficial at ill-defined object boundaries.

In this paper we aim to automatically generate soft masks. Obviously this would make mask-based style transfer more convenient for the user.
However, generating appropriate masks is challenging.
Ideally, the segmentation of the style and content images should be consistent, e.g. using co-segmentation~\cite{vicente2011object}. However, such approaches have not been developed for semantic segmentation. Moreover, the different appearance of photographs compared to artwork (typically used for style images) leads to the cross-depiction problem~\cite{Hall2015}, so that semantic segmentation techniques trained on photographs will fail on paintings.
In this paper we not only demonstrate our approach for the domain of portraits, which are a popular topic for style transfer~\cite{selim2016painting}, and non-photorealistic rendering in general, but also show stylisation of scenes containing other objects, such as cars and trains. Portrait style transfer allows us to leverage state-of-the-art techniques for face detection, that are more robust than general segmentation methods, and are effective even for many artworks.
During our method, facial component masks are automatically extracted using a combination of semantic segmentation, facial landmark detection, and skin detection. 
 
\subsection{Semantic Image Prediction}
\begin{figure}
\centering
    \subfigure[content]{\includegraphics[width=0.23\linewidth]{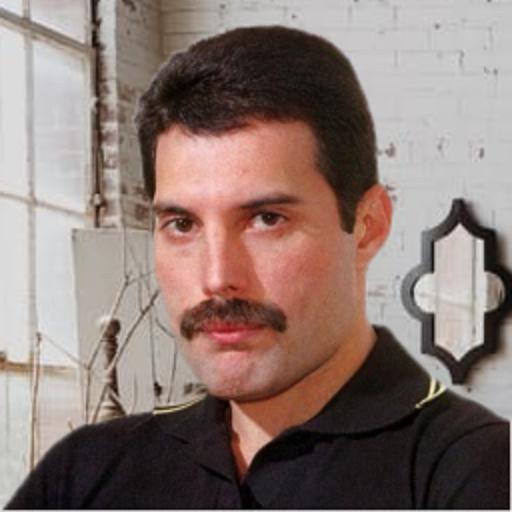}}
    \subfigure[style]{\includegraphics[width=0.23\linewidth]{style_man_09.jpg}}
    \subfigure[probability map for (a)]{\includegraphics[width=0.23\linewidth]{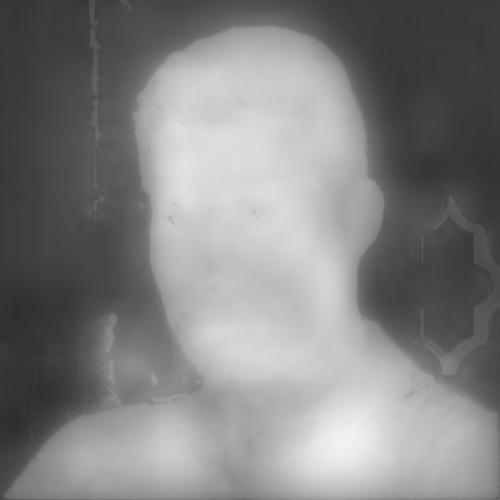}}
    \subfigure[probability map for (b)]{\includegraphics[width=0.23\linewidth]{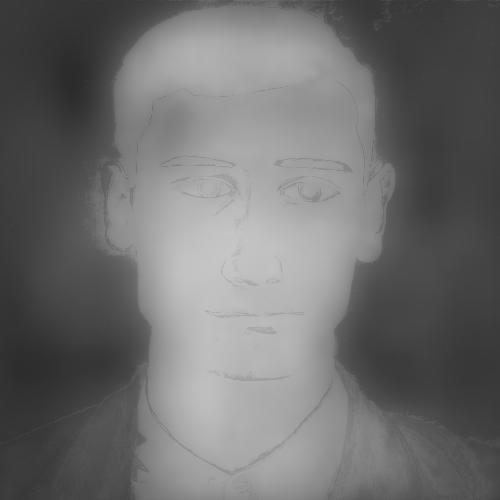}}    
\caption{Probabilistic semantic segmentations using CRF-RNN~\cite{zheng2015conditional} for person prediction.}
\label{fig:Semanticimg}       
\end{figure}

\cite{zheng2015conditional} proposed a semantic segmentation method named CRF-RNN which can segment 20 different objects. CRF-RNN achieves a good result on the popular Pascal VOC segmentation benchmark. This improvement can be attributed to the uniting of the strengths of CNNs and CRFs in a single deep network. In our work, we use CRF-RNN to produce semantic probability maps. 
Instead of labelling each pixel with an object category, we skip over the max pooling stage and extract the neural activations before that and rescale them to [0, 1]. These are treated as probability maps predicting the chance of each pixel belonging to each object category.
An example is shown in figure \ref{fig:Semanticimg}.

This provides 20 probability masks which represent different objects. Since most images only contain a small number of object types,
rather than use all 20 semantic masks we just use a subset of five so as to reduce memory requirements and improve efficiency.
For a given content and style image pair the five semantic masks are automatically selected as the five masks maximising their average probability.

We have found that the CRF-RNN is mostly reliable for photographs. For paintings its performance degrades, especially as the style of the artwork becomes more extreme. However, it is still capable of producing adequate extractions of people, cars, etc. for many paintings (used as style images) that we have tested.

\subsection{Skin Detection}

Skin detection is performed on the photographic images~\cite{brancati2017human}, using a rule-based analysis of pixels in YCbCr colour space. The skin mask is then intersected with the person mask provided by the CRF-RNN, so as to subdivide the person into skin and non-skin (e.g. hair, clothing). An example is shown in figure \ref{fig:facial_part_seg}.

Since skin detection is primarily colour based, it is not in general effective on artwork due to the typical colour shifts, as well as distortions caused by strong brush stroke textures. Therefore, for paintings
the facial region is detected using the face detector, rather than using skin detection.

\subsection{Face and Facial Part Segmentation}

\begin{figure}
\centering
    \includegraphics[width=0.18 \linewidth]{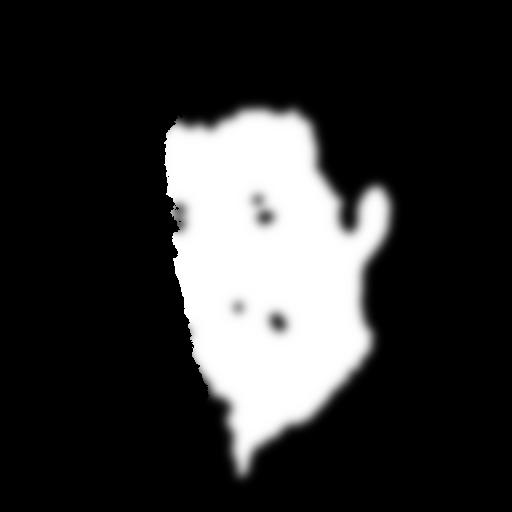}
   \includegraphics[width=0.18 \linewidth]{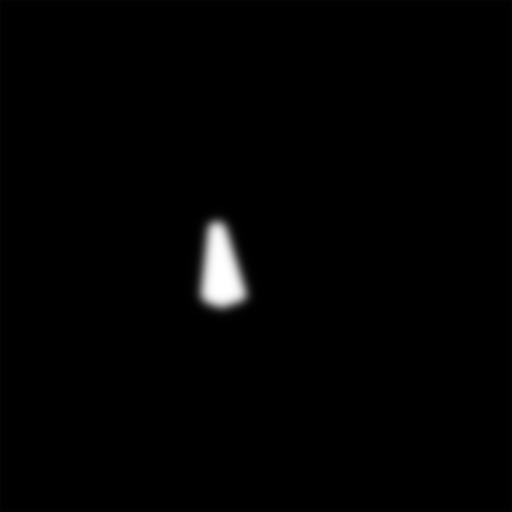}
   \includegraphics[width=0.18 \linewidth]{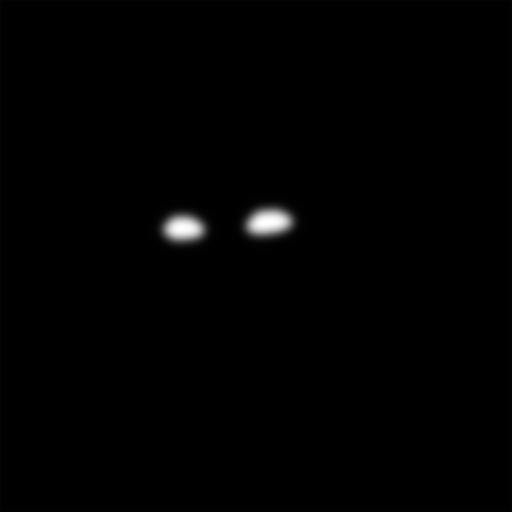}
  \includegraphics[width=0.18 \linewidth]{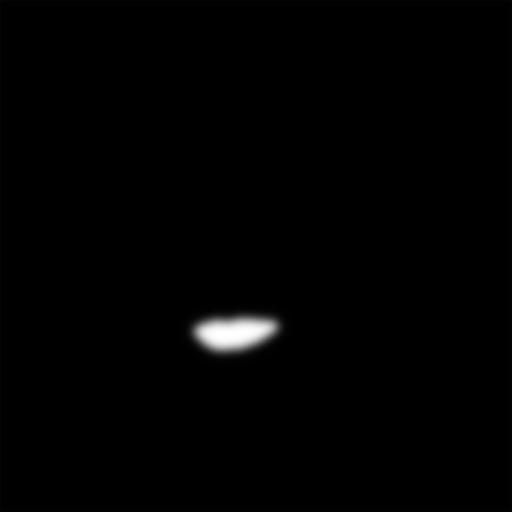}
    \includegraphics[width=0.18 \linewidth]{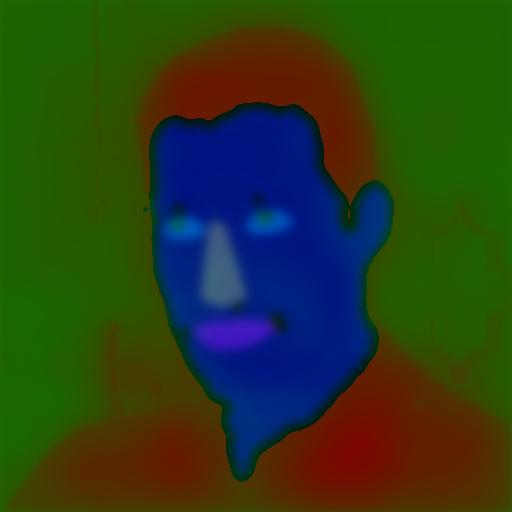}  \\
    
    \subfigure[skin]{\includegraphics[width=0.18 \linewidth]{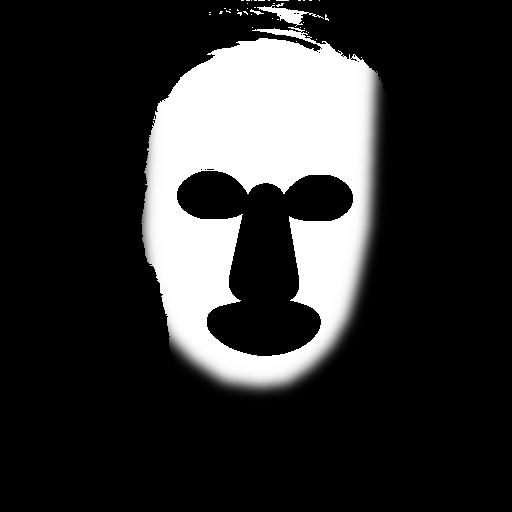}} 
    \subfigure[eyes]{\includegraphics[width=0.18 \linewidth]{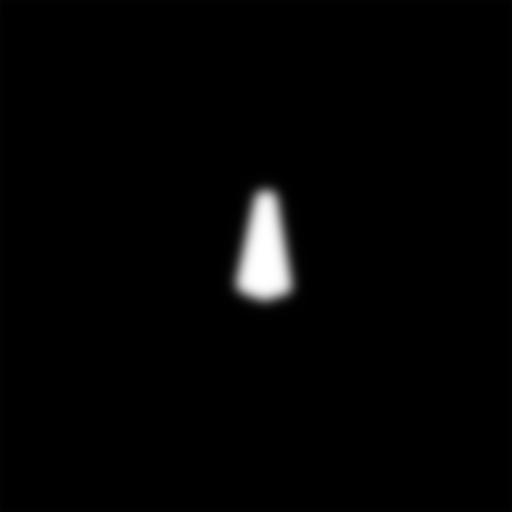}}    
    \subfigure[mouth]{\includegraphics[width=0.18 \linewidth]{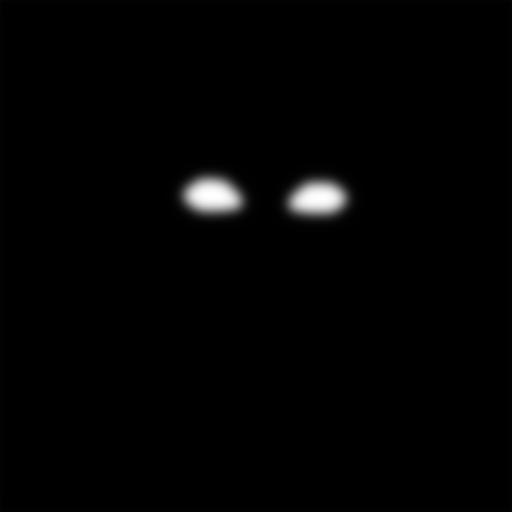}} 
    \subfigure[nose]{\includegraphics[width=0.18 \linewidth]{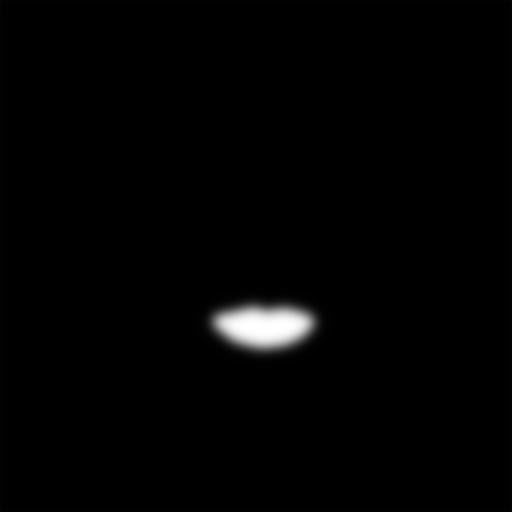}} 
    \subfigure[masks]{\includegraphics[width=0.18 \linewidth]{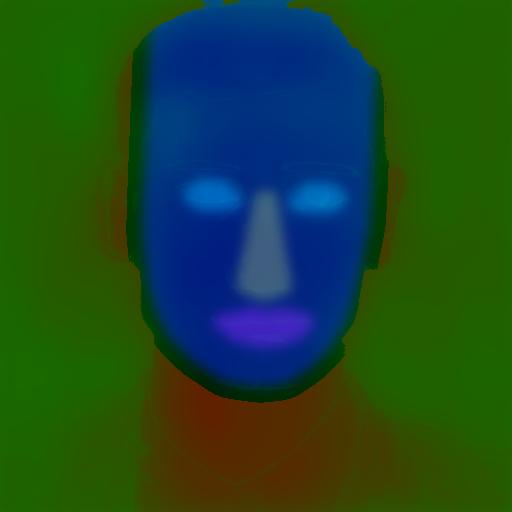}}      
\caption{Segmentation of facial components for the images in figure~\ref{fig:Semanticimg}.}
\label{fig:facial_part_seg}       
\end{figure}

Facial landmark detection is performed using OpenFace \cite{baltruvsaitis2016openface}, which is based on Conditional Local Neural Fields, a version of the well known Constrained Local Model approach. Sixty-eight facial landmarks are located, from which the eye, nose, inner and outer mouth regions are determined -- see figure \ref{fig:facial_part_seg}.

Since the facial landmarks only cover the lower half of the face, the outline of the face is extended upwards, and intersected with the person mask provided by the semantic segmentation to produce a good approximation to the head region. This mask is used for artwork. For photographs the skin mask is used instead of the extended facial region as it is more accurate (although prone to noise).

The above steps result in a set of masks that are blurred to produce soft masks identifying the following objects: face/skin, nose, eye, mouth, see figure~\ref{fig:facial_part_seg} for an example.
To provide a more compact visualisation we also combine the set of soft masks into a single colour image, see figure~\ref{fig:facial_part_seg}. The soft masks for body, background and face/skin are mapped to red, green, blue respectively, while the eyes, nose and mouth values are mapped to cyan, yellow, magenta respectively.
(Note that when performing style transfer the multiple soft image masks are used instead.)

\section{Results}\label{sec:results}

We use the pre-trained 19-layer VGG-Network with the augmented layers $myConv3\_1 $ and $myConv4\_1 $. For layers $relu3\_1$, $relu4\_1$, $myConv3\_1 $ and $myConv4\_1 $ we use $3\times 3$ patches, and we set the stride to one. 
Following the patch-based approach of \cite{li2016combining}, we synthesise at multiple increasing resolutions, and randomly initialise the optimisation. On a GTX TIT with 12Gb of GPU RAM, synthesis takes from 5 to 30 minutes depending on the quality and resolution.

We will now compare the proposed method with several popular methods: \cite{gatys2015neural,li2016combining} which are representative global and local neural style transfer methods, and \cite{gatys2016controlling,champandard2016semantic} which use manual segmentation to improve style transfer. 

Note that for our method multiple soft masks were used; the single colour mask is just shown for illustrative purposes. For~\cite{champandard2016semantic}, we set the content weight to 10, style weight to 25, semantic weight to 100, and we use the masks from \cite{champandard2016semantic} when available and otherwise manually draw them ourselves. For \cite{gatys2016controlling}, we used two image maps of values in the range [0,1] for content and style images like figure~\ref{fig:Semanticimg} (c, d), similar to the examples used in their paper, which are also used in our method. To partially overcome orientation and scale differences between the style and the content images, we also allow a range of rotations and scalings to be considered in the CNNMRF, following the settings in \cite{li2016combining}.


\begin{figure*}
\centering
   \includegraphics[width=0.135\linewidth]{Freddie5.jpg}
   \hspace{0.5mm}
    \includegraphics[width=0.135\linewidth]{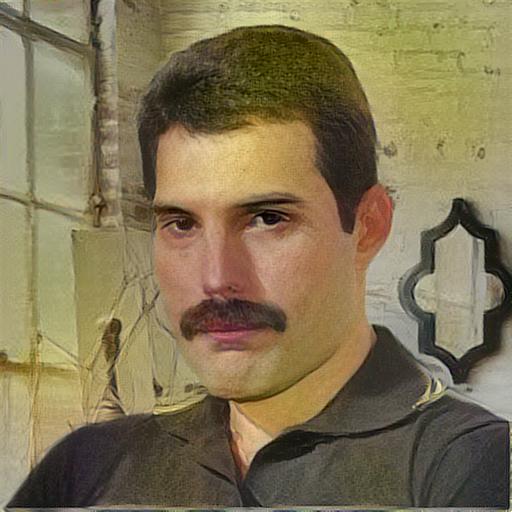}
    \includegraphics[width=0.135\linewidth]{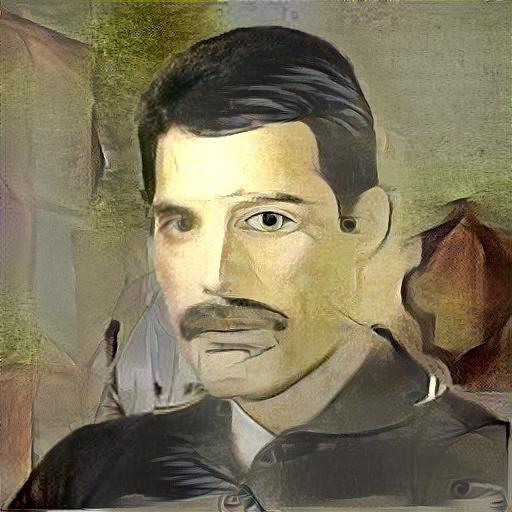}    
    \includegraphics[width=0.135\linewidth]{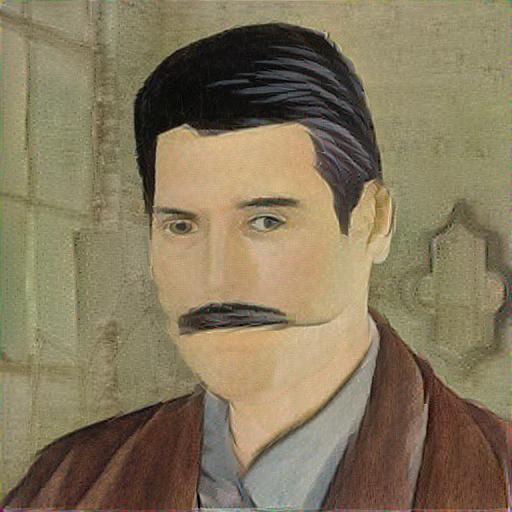}
    \includegraphics[width=0.135\linewidth]{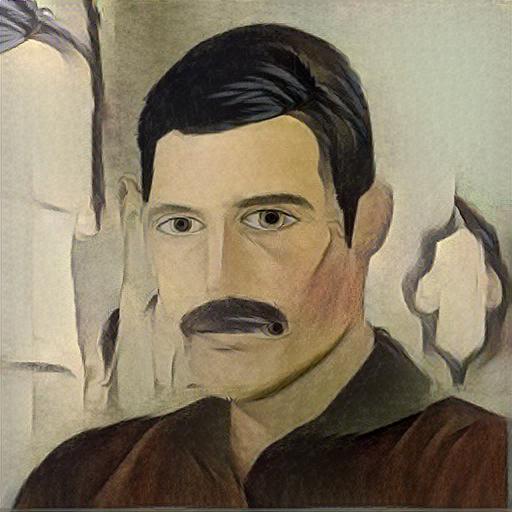}    
    \includegraphics[width=0.135\linewidth]{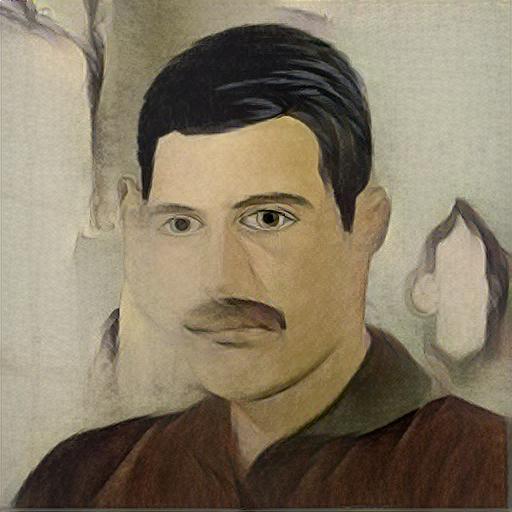}\\
    \includegraphics[width=0.135\linewidth]{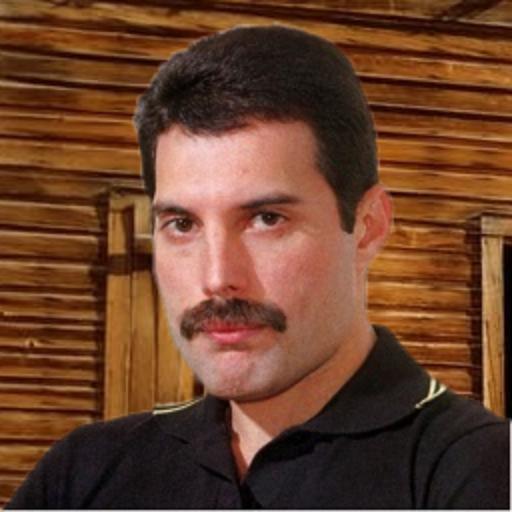}
    \hspace{0.5mm}
    \includegraphics[width=0.135\linewidth]{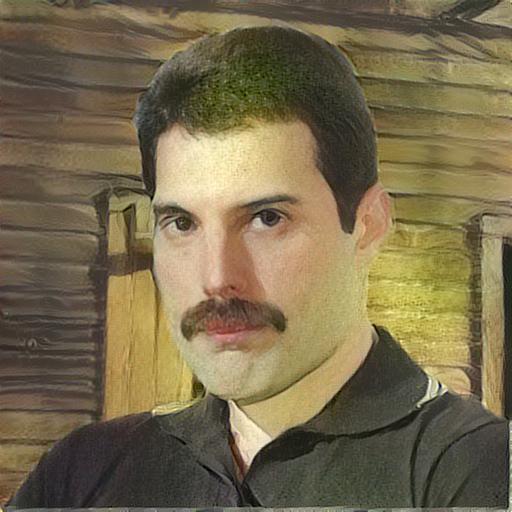}
    \includegraphics[width=0.135\linewidth]{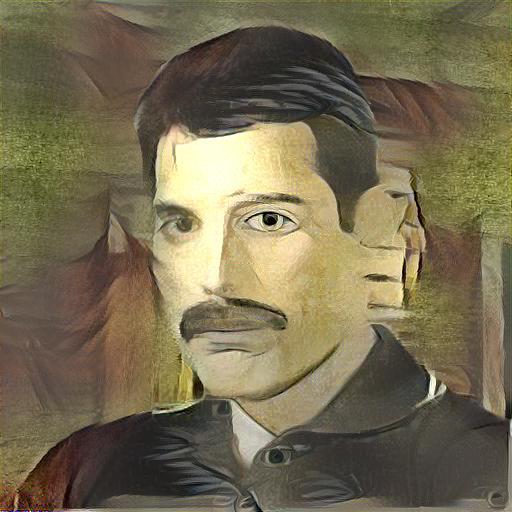}    
    \includegraphics[width=0.135\linewidth]{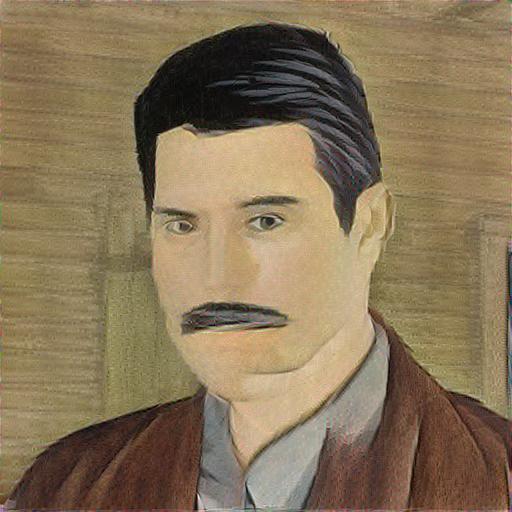}    
    \includegraphics[width=0.135\linewidth]{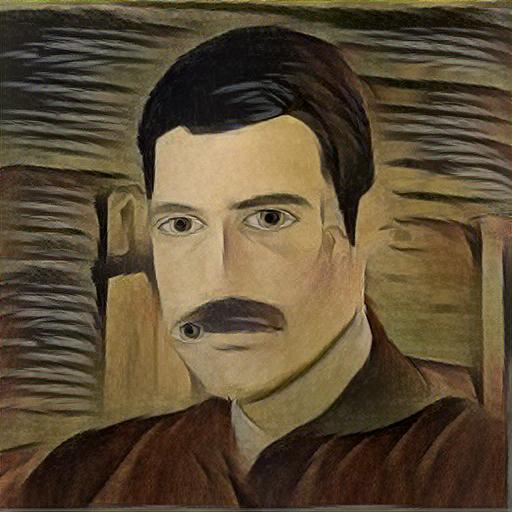}
    \includegraphics[width=0.135\linewidth]{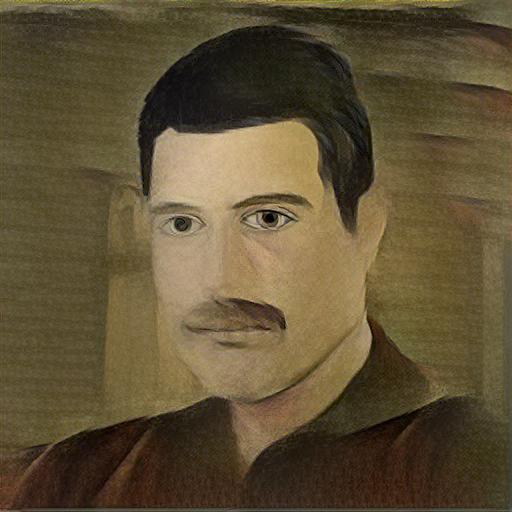}\\
    \subfigure[Content]{\includegraphics[width=0.135\linewidth]{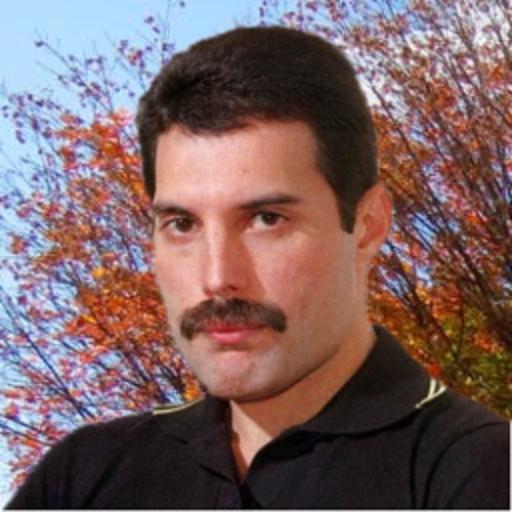}}
    \hspace{0.5mm}
    \subfigure[\cite{gatys2015neural}]{\includegraphics[width=0.135\linewidth]{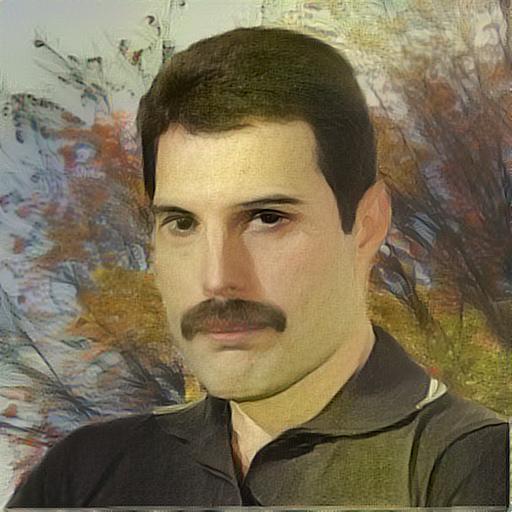}}
    \subfigure[\cite{gatys2016controlling}]{\includegraphics[width=0.135\linewidth]{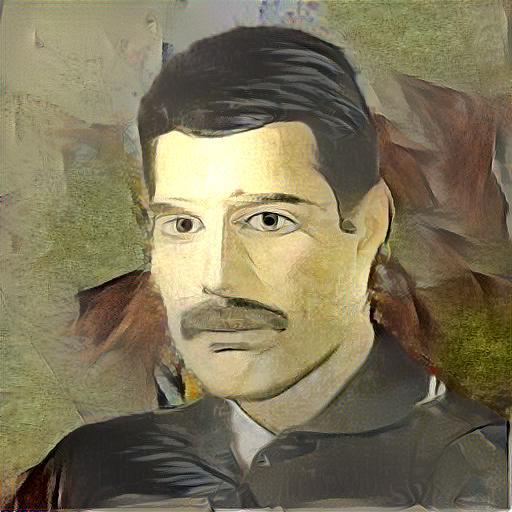}}    
    \subfigure[\cite{champandard2016semantic}]{\includegraphics[width=0.135\linewidth]{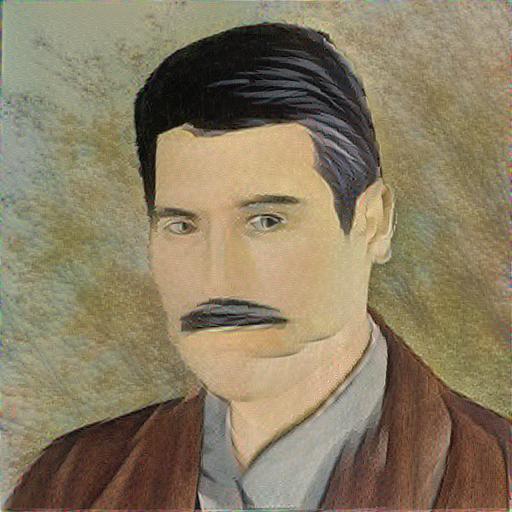}}
    \subfigure[\cite{li2016combining}]{\includegraphics[width=0.135\linewidth]{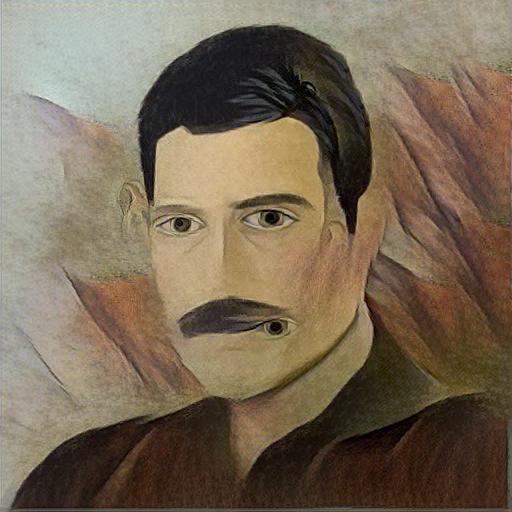}}    
    \subfigure[Our]{\includegraphics[width=0.135\linewidth]{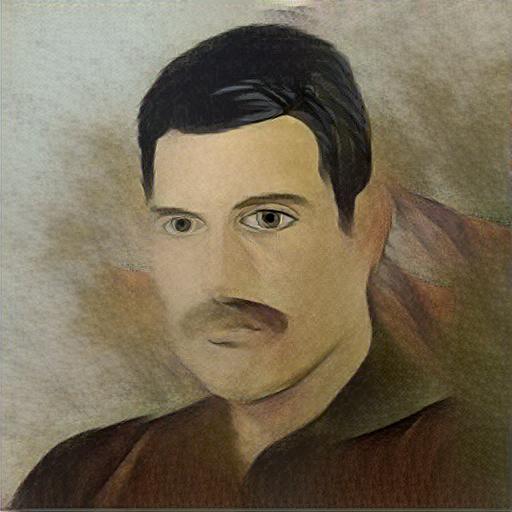}}
\caption{Style transfer results from several methods using versions of a content image with varied backgrounds.}
\label{fig:Freddie_style09_02}       
\end{figure*}
  
We use figure~\ref{fig:Semanticimg}(a) with several different backgrounds as the content image, and choose figure~\ref{fig:Semanticimg}(b) as the style image. Style transfer results obtained by the different methods are shown in figure~\ref{fig:Freddie_style09_02}. 
Considering the four existing methods and by comparing the results in figure~\ref{fig:Freddie_style09_02}, it seems that \cite{gatys2015neural} and \cite{li2016combining} cannot transfer the background texture well. \cite{champandard2016semantic} achieves better background texture transfer, comparable to our method, but some key facial parts (nose and mouth) are lost. \cite{gatys2016controlling} can control the spatial texture very well, but the human style transfer is not so good. It also generates errors in rows 1 and 2 of figure~\ref{fig:Freddie_style09_02}(c). 

Because both our method and \cite{li2016combining} are based on the MRF regulariser, and \cite{li2016combining} has previously demonstrated better results than \cite{gatys2015neural}, we mainly compare our result with \cite{li2016combining}.

\begin{figure}
\centering
  \includegraphics[width=0.13\linewidth]{Freddie5.jpg} 
  \includegraphics[width=0.13\linewidth]{man_12.jpg}   
  \includegraphics[width=0.13\linewidth]{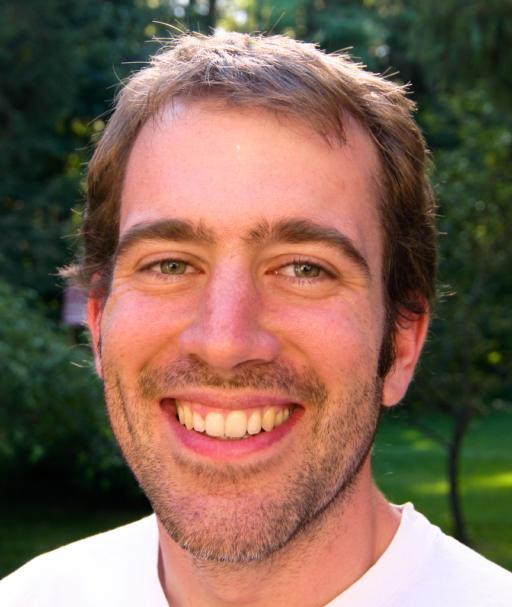}  
  \includegraphics[width=0.13\linewidth]{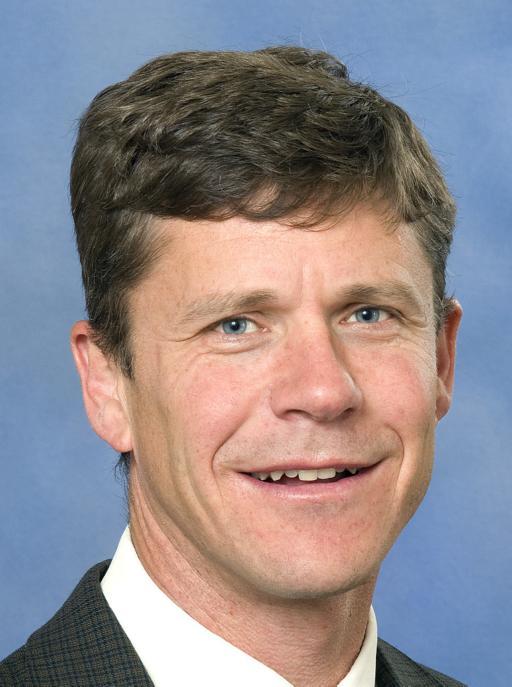}
   \includegraphics[width=0.13 \linewidth]{style_man_09.jpg}
   \includegraphics[width=0.13 \linewidth]{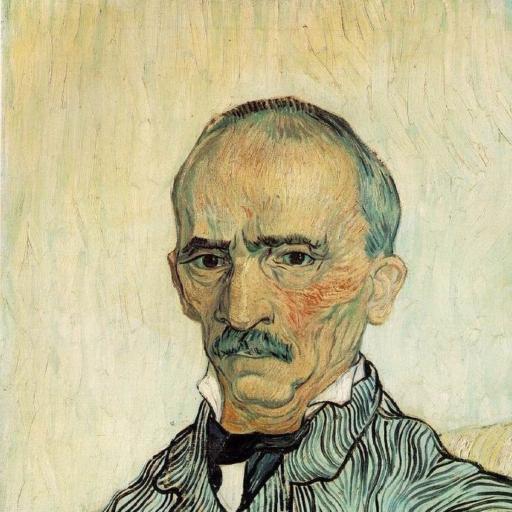} 
   \includegraphics[width=0.13 \linewidth]{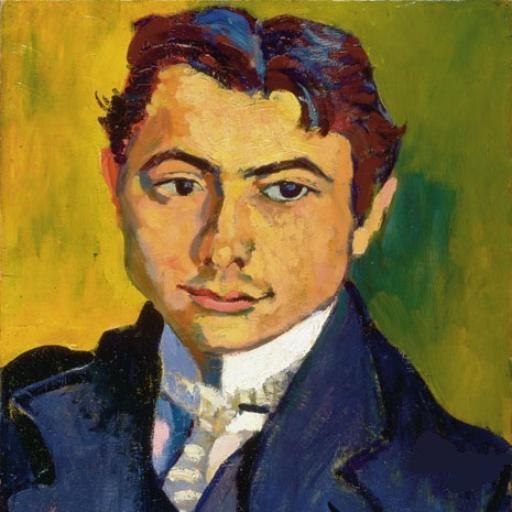}   
   \\  
  \subfigure[]{\includegraphics[width=0.13\linewidth]{Freddie5_pb_06sem_v5_masks_all.jpg}}
  \subfigure[]{\includegraphics[width=0.13\linewidth]{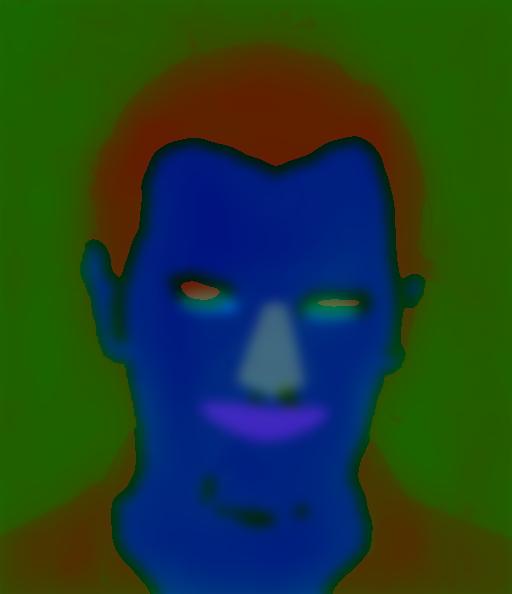}}
  \subfigure[]{\includegraphics[width=0.13\linewidth]{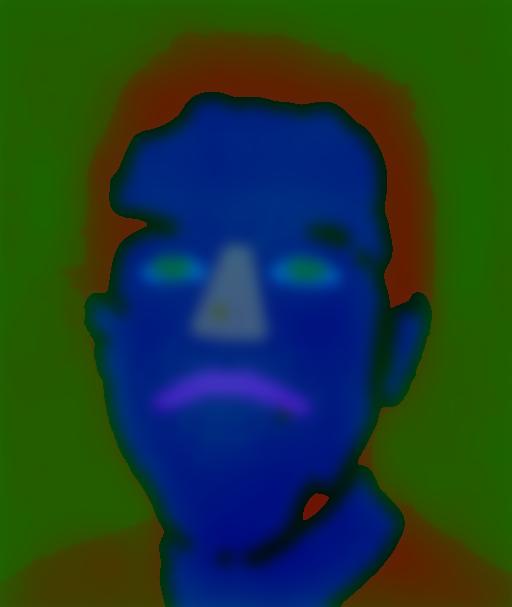}}
  \subfigure[]{\includegraphics[width=0.13\linewidth]{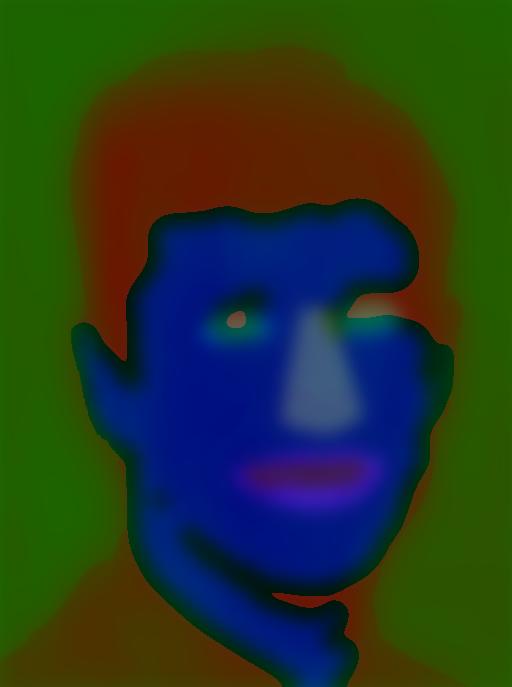}}
  \subfigure[]{\includegraphics[width=0.13\linewidth]{style_man_09_pb_06sem_v5_masks_all.jpg}}
  \subfigure[]{\includegraphics[width=0.13\linewidth]{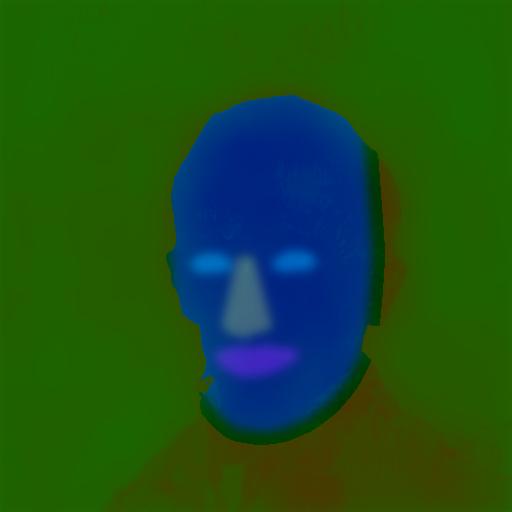}}
  \subfigure[]{\includegraphics[width=0.13\linewidth]{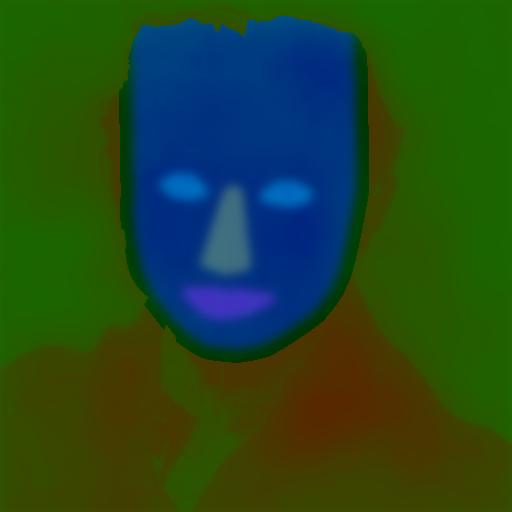}}  
\caption{Content (a--d) and style (e--g) images and visualisations of their soft masks.}  
\label{fig:content_man_2}    
\end{figure}

\begin{figure}
\centering
    \includegraphics[width=0.23\linewidth]{Freddie5_TO_style_man_09_cnnmrf.jpg}
    \includegraphics[width=0.23\linewidth]{man_12_TO_style_man_09_cnnmrf.jpg}  
    \includegraphics[width=0.23\linewidth]{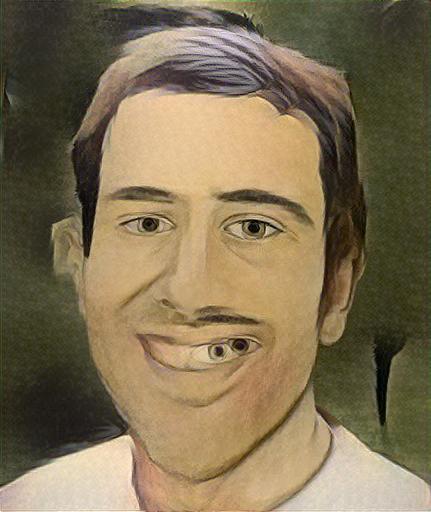}
    \includegraphics[width=0.23\linewidth]{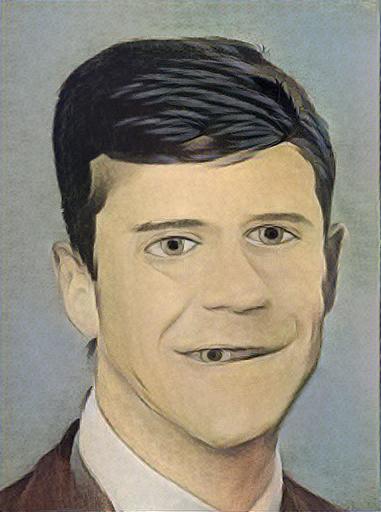}    
\\
    \includegraphics[width=0.23\linewidth]{Freddie5_TO_style_man_09_pb_hh.jpg}
    \includegraphics[width=0.23\linewidth]{man_12_TO_style_man_09_pb_hh.jpg}  
    \includegraphics[width=0.23\linewidth]{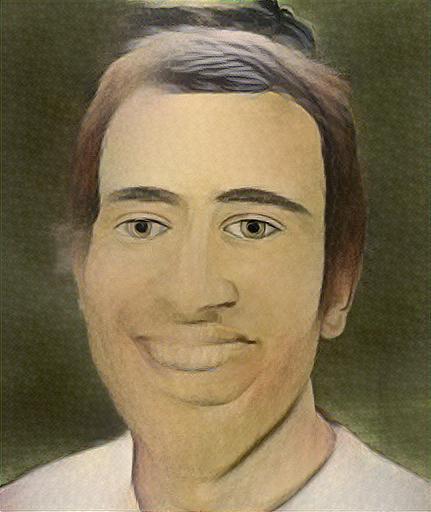}  
    \includegraphics[width=0.23\linewidth]{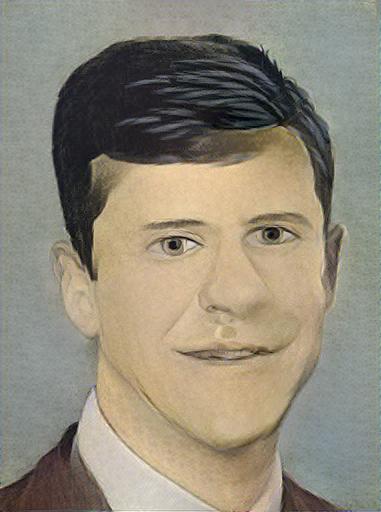}
    \\
    \includegraphics[width=0.23\linewidth]{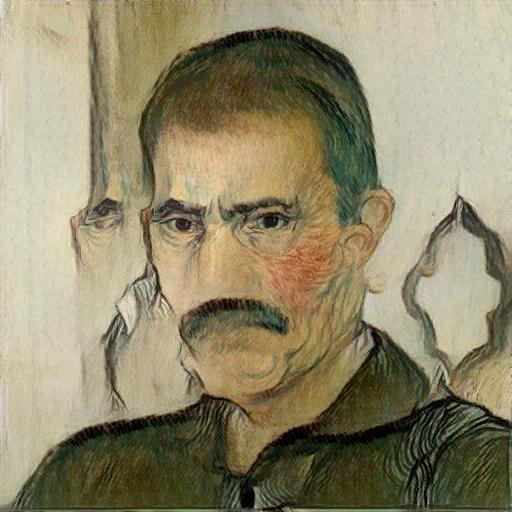}
    \includegraphics[width=0.23\linewidth]{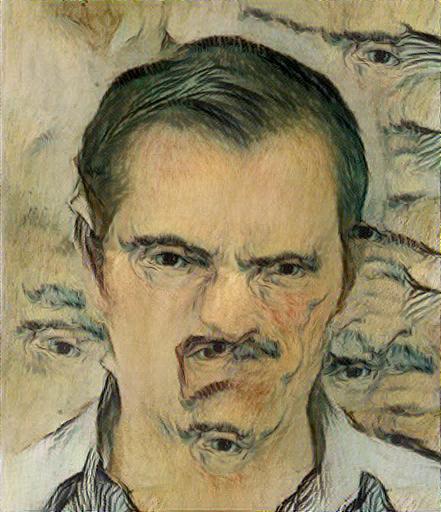}  
    \includegraphics[width=0.23\linewidth]{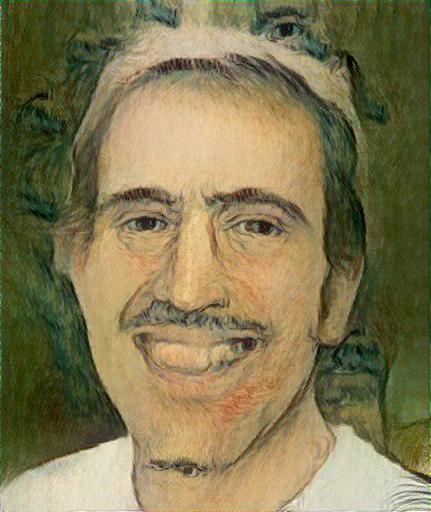}
    \includegraphics[width=0.23\linewidth]{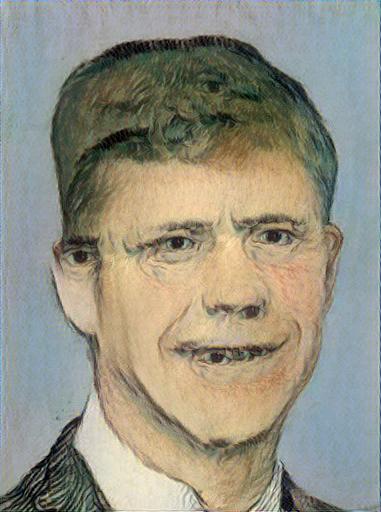}    
    \\  
    \includegraphics[width=0.23\linewidth]{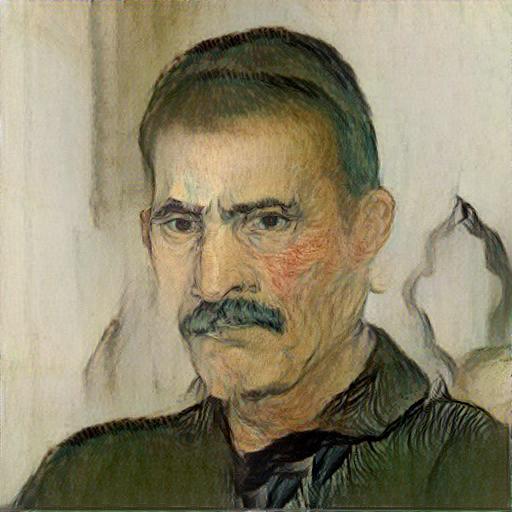} 
    \includegraphics[width=0.23\linewidth]{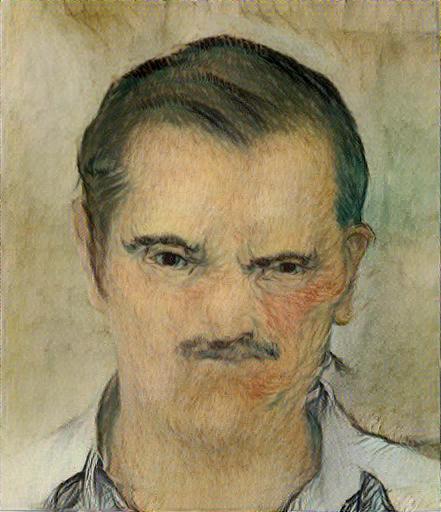} 
    \includegraphics[width=0.23\linewidth]{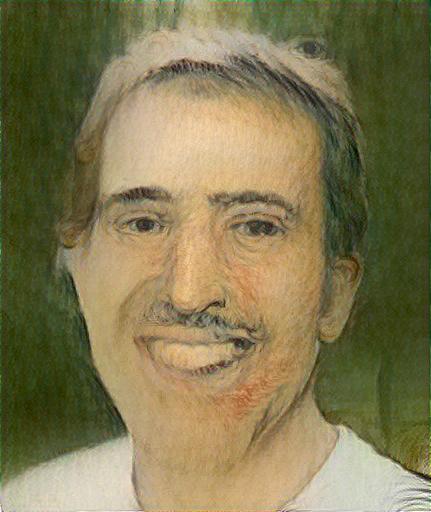}     
    \includegraphics[width=0.23\linewidth]{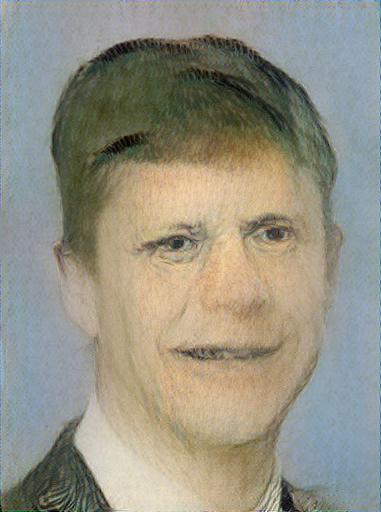} 
    \\    
    \includegraphics[width=0.23\linewidth]{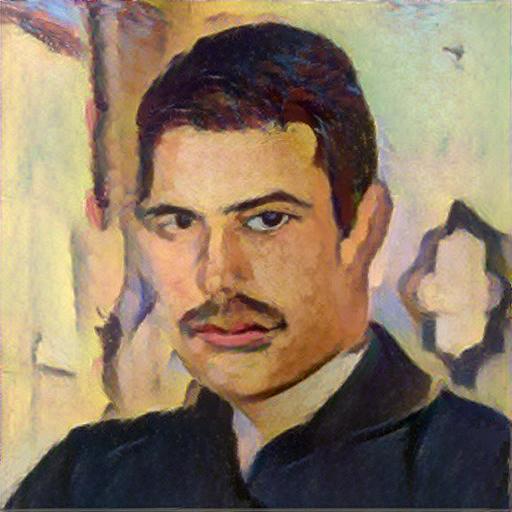}
    \includegraphics[width=0.23\linewidth]{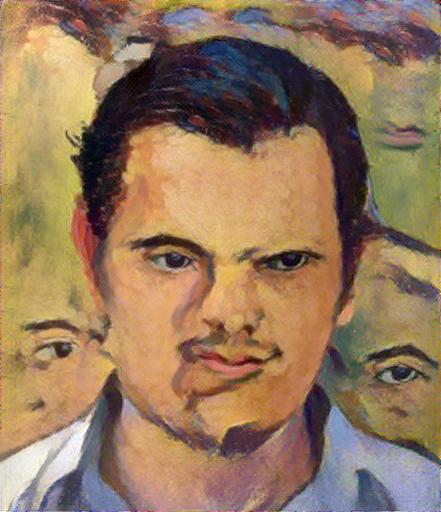}  
    \includegraphics[width=0.23\linewidth]{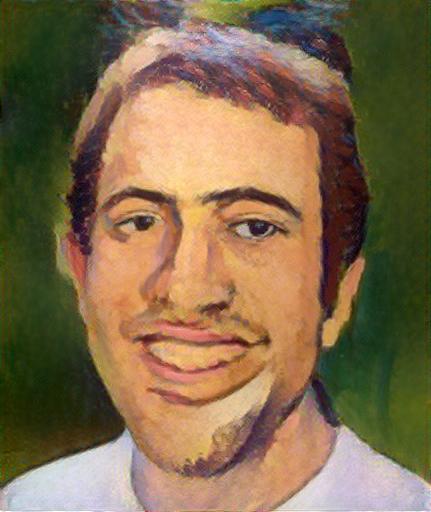}
    \includegraphics[width=0.23\linewidth]{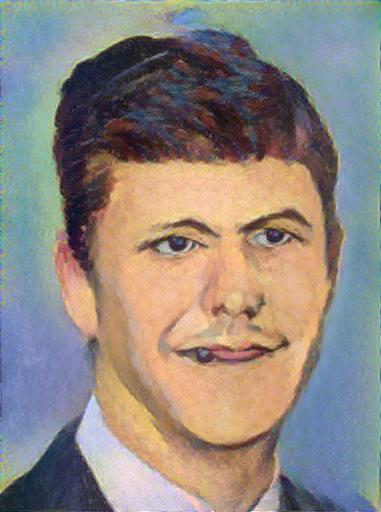}    
    \\ 
    \includegraphics[width=0.23\linewidth]{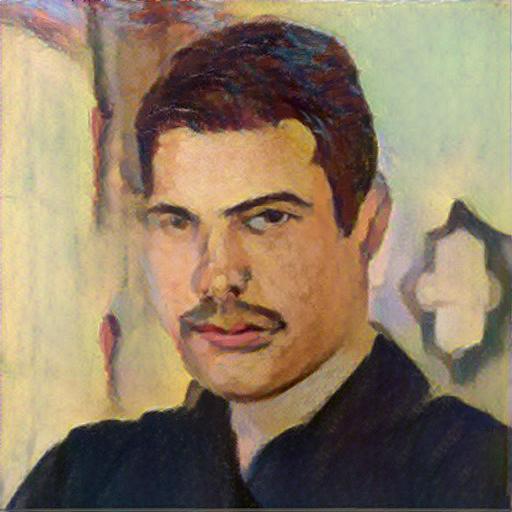} 
    \includegraphics[width=0.23\linewidth]{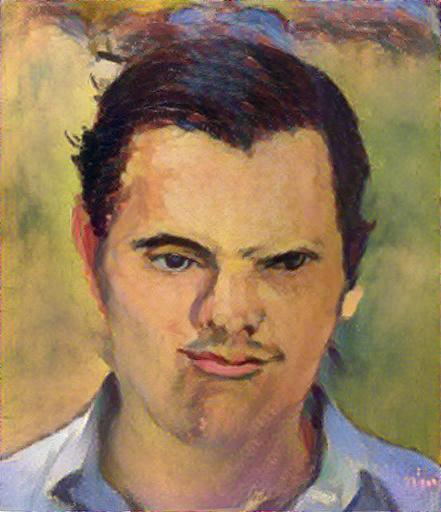}
    \includegraphics[width=0.23\linewidth]{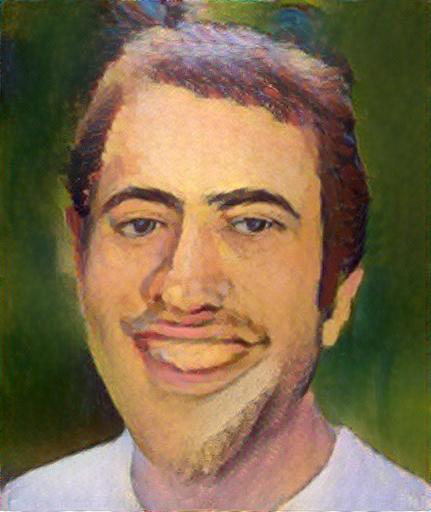}   
    \includegraphics[width=0.23\linewidth]{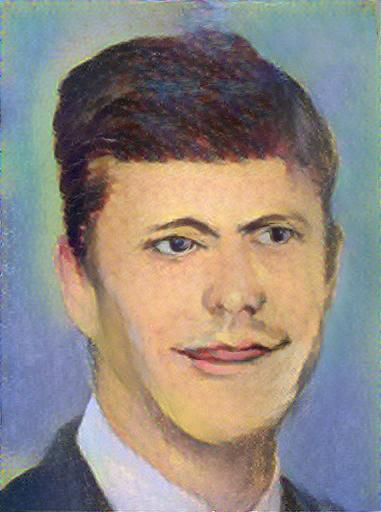}            
\caption{Comparison of style transfer results (odd lines CNNMRF method, even lines our method).}
\label{fig:com_more02}       
\end{figure}

\begin{figure}
\centering
    \includegraphics[width=0.13 \linewidth]{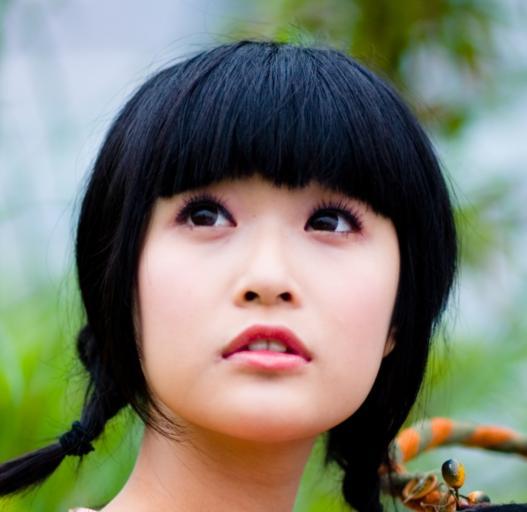} 
    \includegraphics[width=0.13 \linewidth]{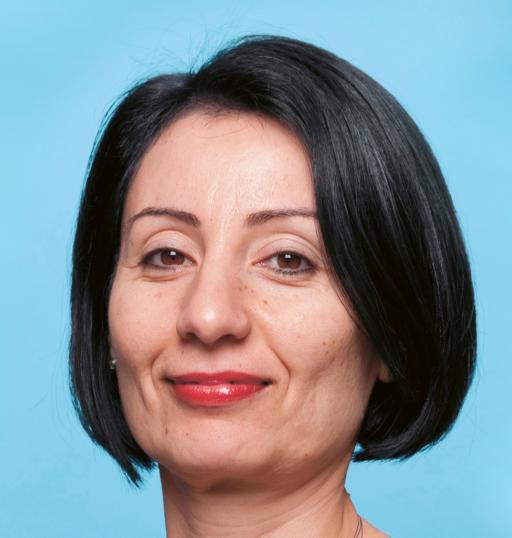}
    \includegraphics[width=0.13 \linewidth]{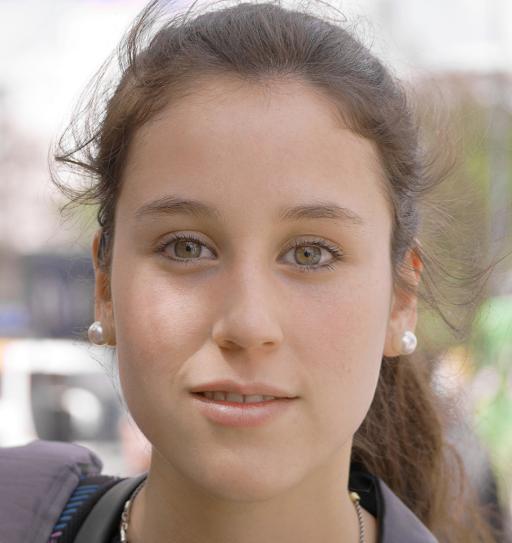}
    \includegraphics[width=0.13 \linewidth]{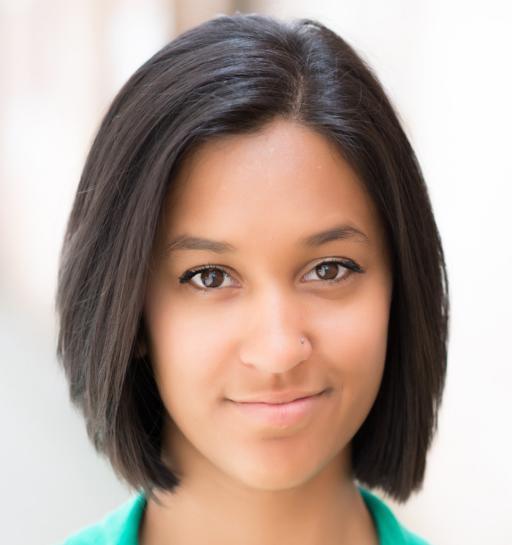}
    \includegraphics[width=0.13 \linewidth]{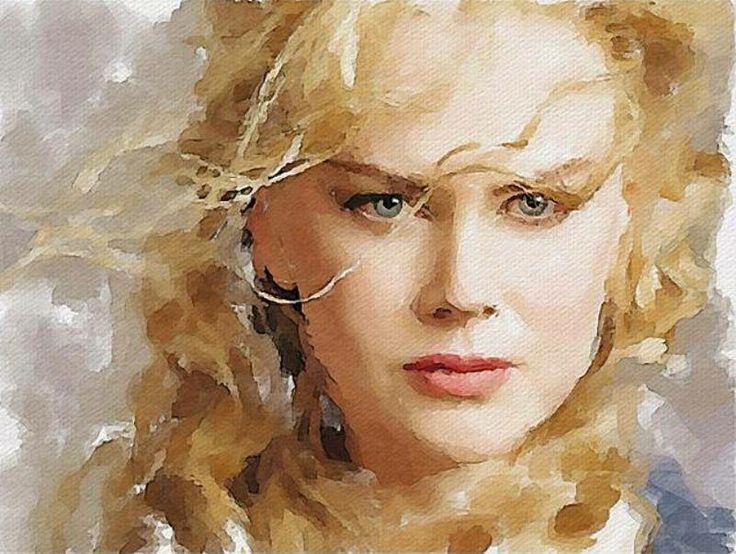} 
    \includegraphics[width=0.13 \linewidth]{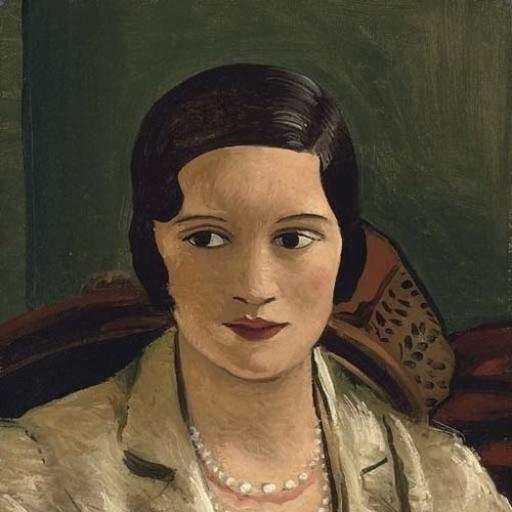}
    \includegraphics[width=0.13 \linewidth]{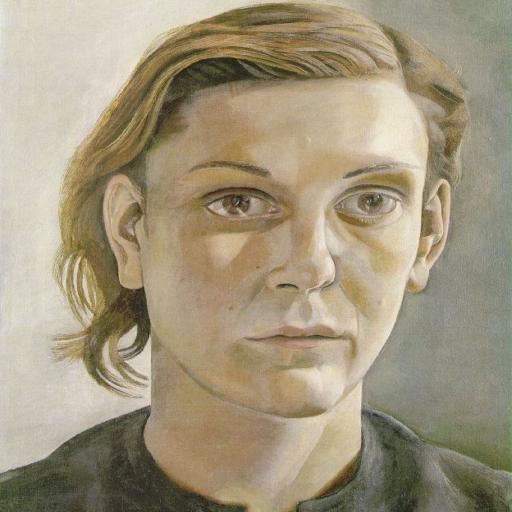}
    \\
    \subfigure[]{\includegraphics[width=0.13 \linewidth]{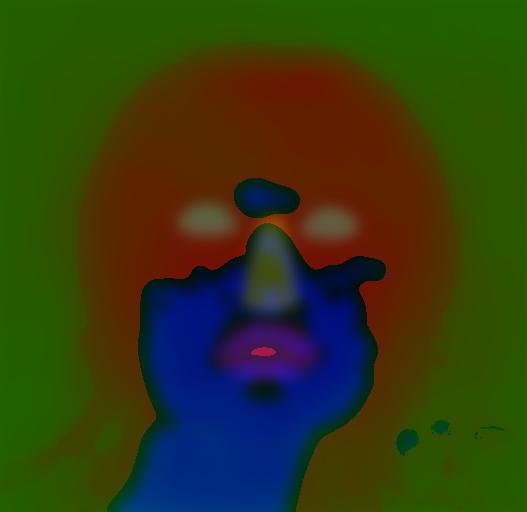}}
    \subfigure[]{\includegraphics[width=0.13 \linewidth]{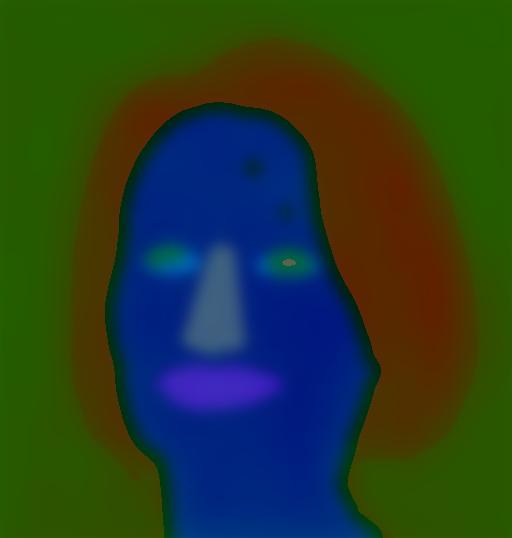}}
    \subfigure[]{\includegraphics[width=0.13 \linewidth]{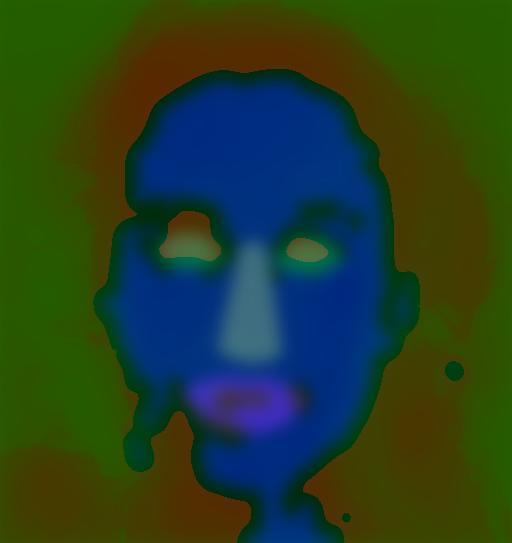}}  
    \subfigure[]{\includegraphics[width=0.13 \linewidth]{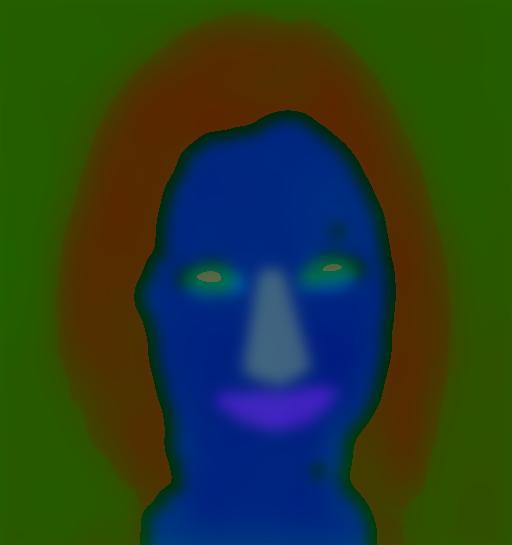}}
    \subfigure[]{\includegraphics[width=0.13\linewidth]{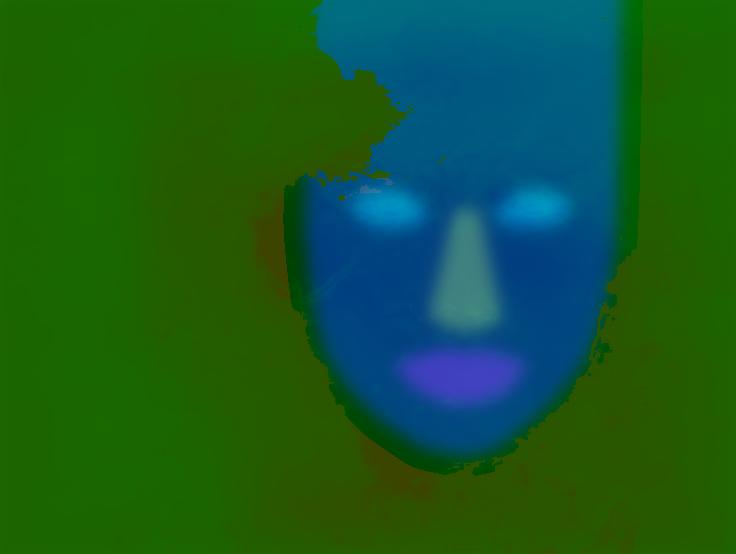}}
     \subfigure[]{\includegraphics[width=0.13\linewidth]{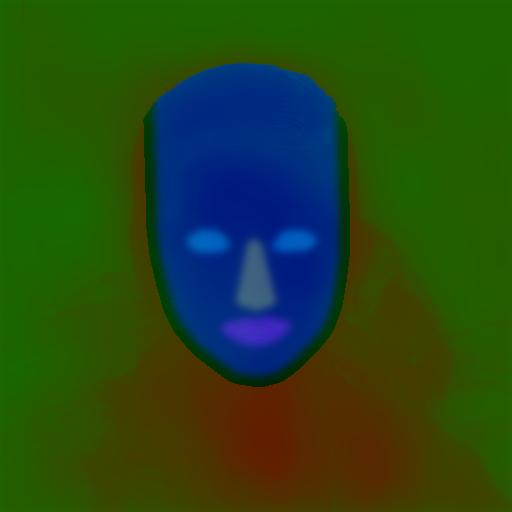}}
     \subfigure[]{\includegraphics[width=0.13\linewidth]{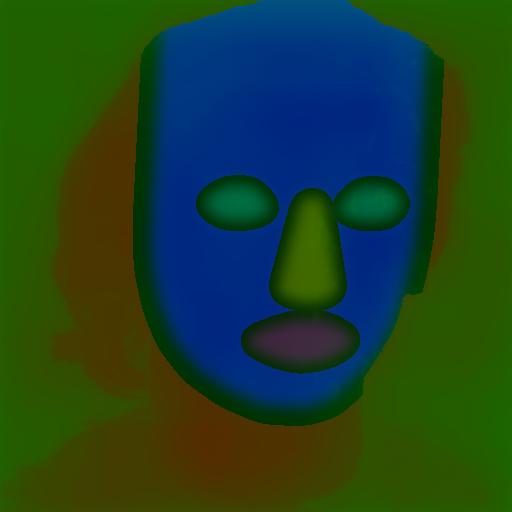}}
\caption{Content (a--d) and style (e--g) images and visualisations of their soft masks.}  
\label{fig:style_woman_02} 
\end{figure}

\begin{figure}
\centering 
    \includegraphics[width=0.23\linewidth]{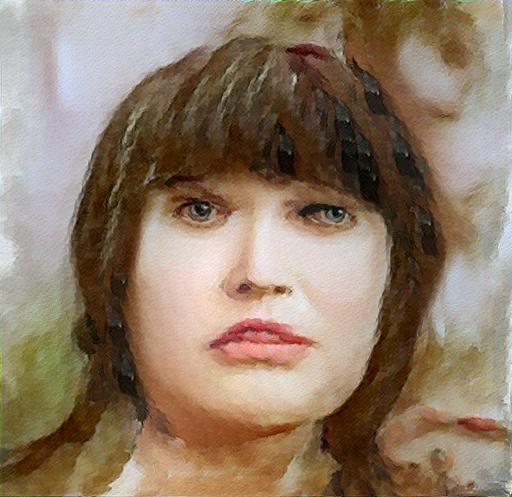}  
    \includegraphics[width=0.23\linewidth]{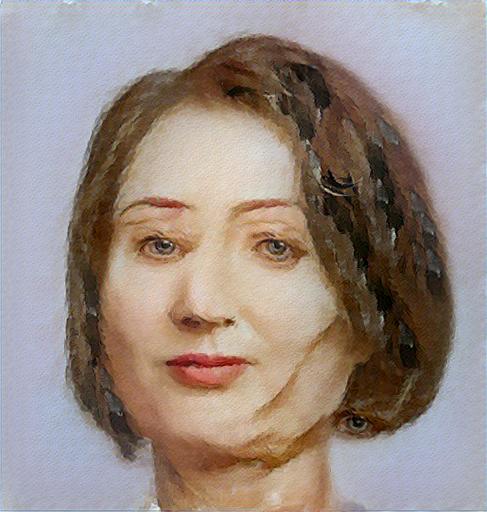}
    \includegraphics[width=0.23\linewidth]{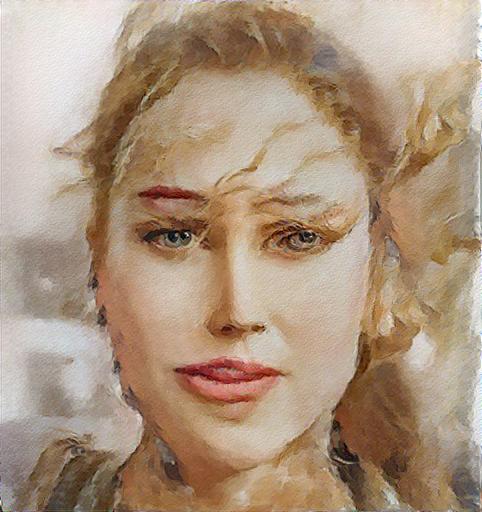}
    \includegraphics[width=0.23\linewidth]{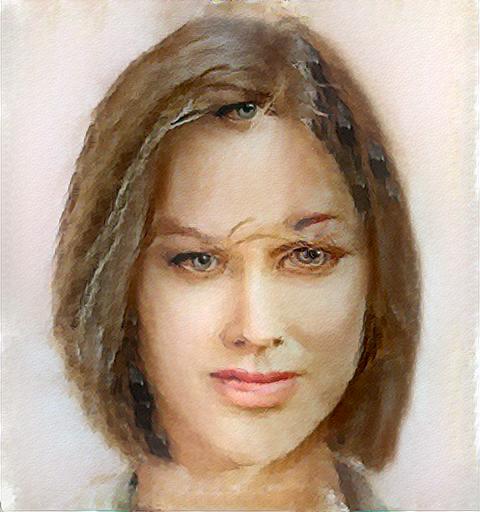}
    \\
    \includegraphics[width=0.23\linewidth]{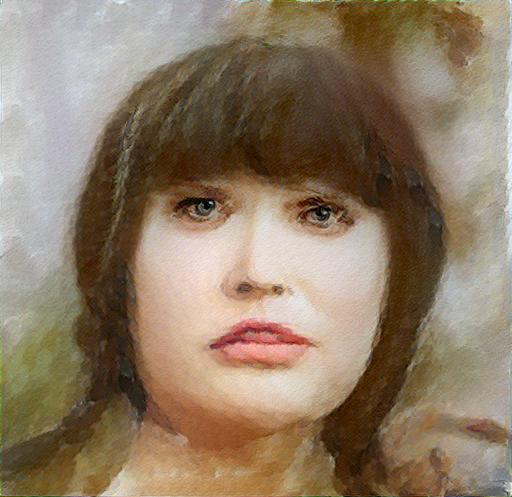}    
    \includegraphics[width=0.23\linewidth]{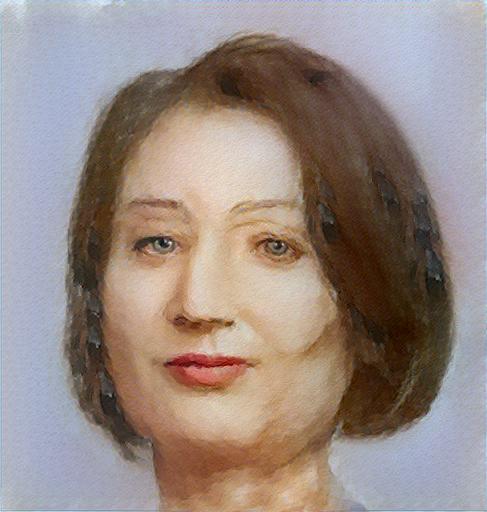}    
    \includegraphics[width=0.23\linewidth]{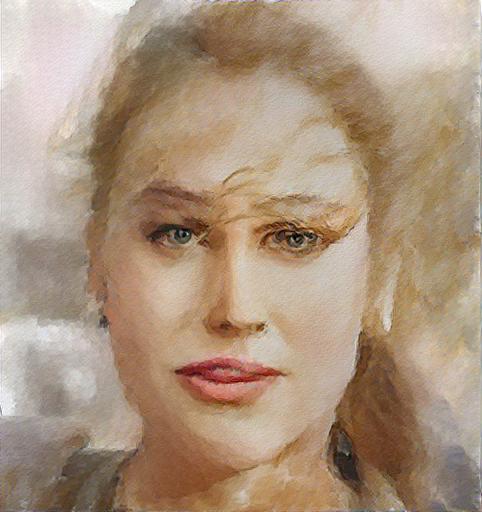}    
    \includegraphics[width=0.23\linewidth]{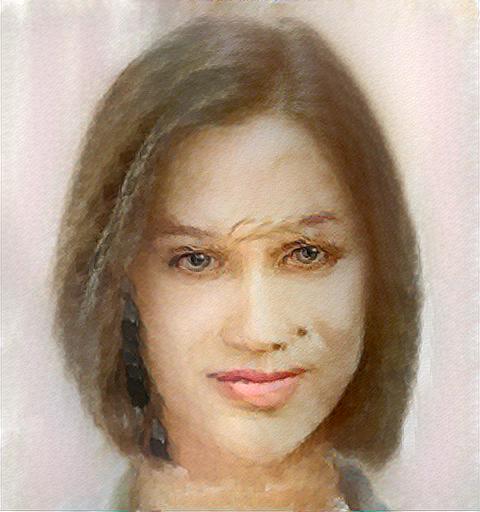}    
     \\
   
    \includegraphics[width=0.23\linewidth]{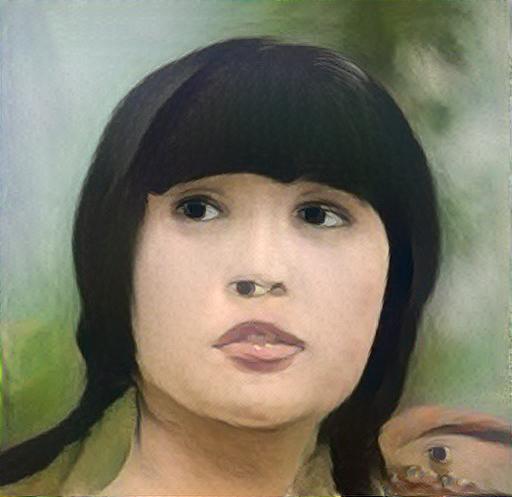}    
    \includegraphics[width=0.23\linewidth]{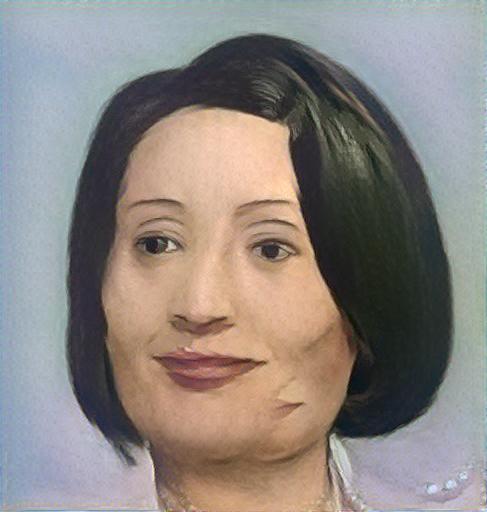}
    \includegraphics[width=0.23\linewidth]{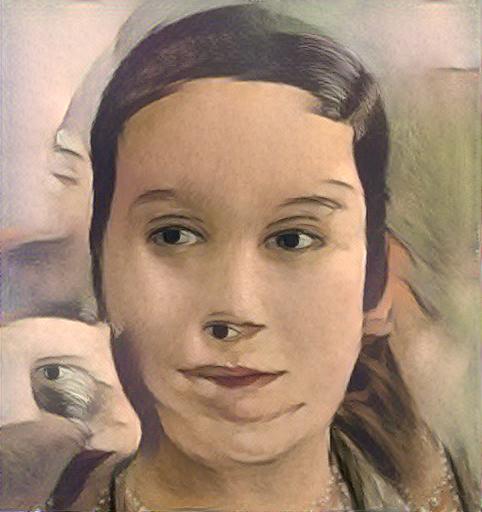}
    \includegraphics[width=0.23\linewidth]{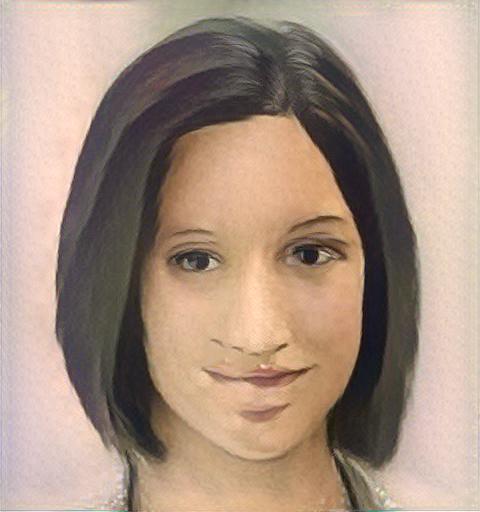}
 \\
    \includegraphics[width=0.23\linewidth]{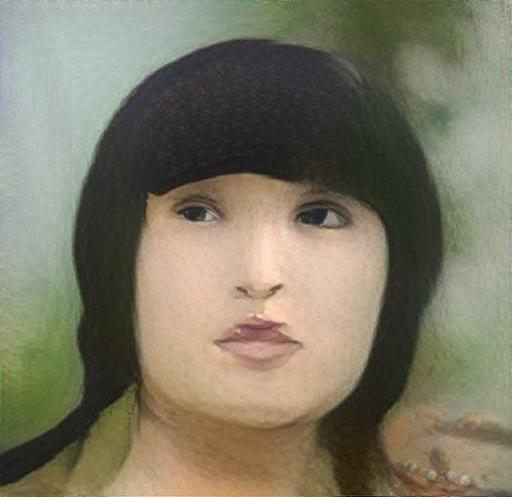}    
    \includegraphics[width=0.23\linewidth]{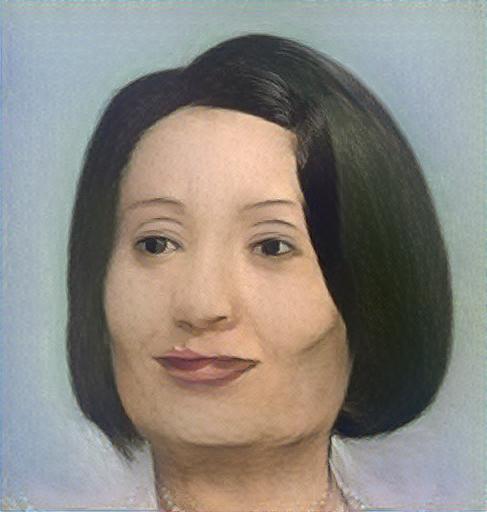}
    \includegraphics[width=0.23\linewidth]{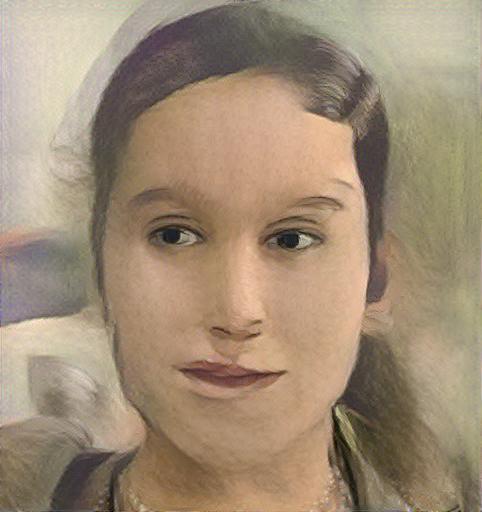}
    \includegraphics[width=0.23\linewidth]{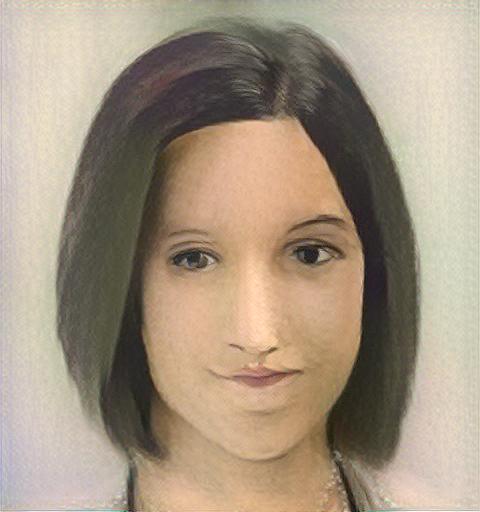}
\\   
    \includegraphics[width=0.23\linewidth]{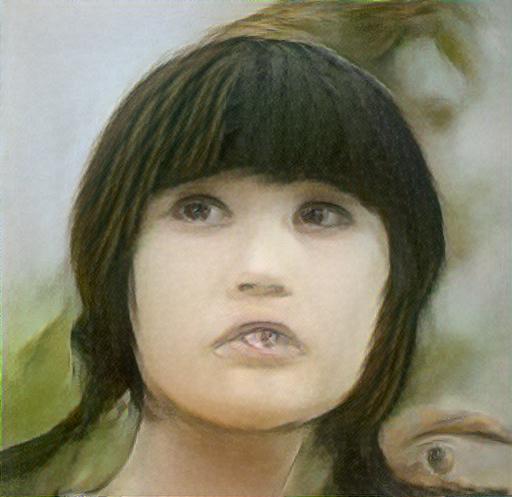}    
    \includegraphics[width=0.23\linewidth]{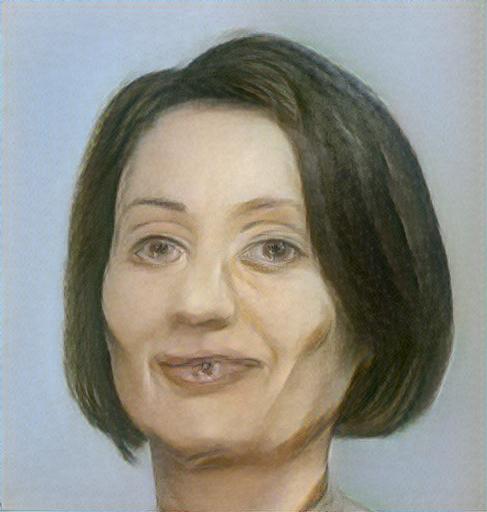}
    \includegraphics[width=0.23\linewidth]{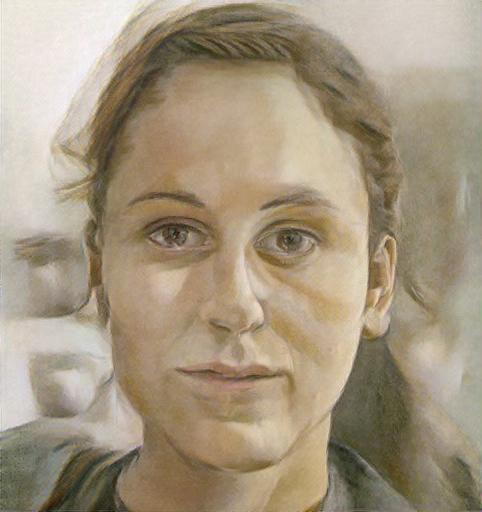}
    \includegraphics[width=0.23\linewidth]{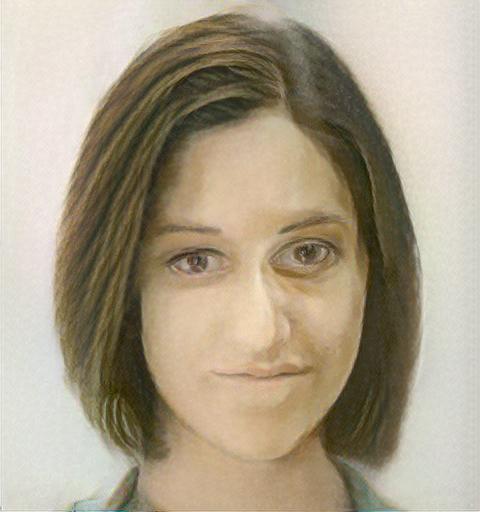}
\\
    \includegraphics[width=0.23\linewidth]{woman_10_TO_style_woman_07_res_3_100_hh.jpg}    
    \includegraphics[width=0.23\linewidth]{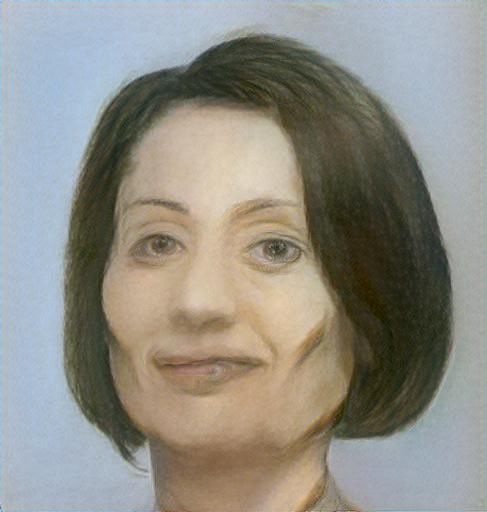}
    \includegraphics[width=0.23\linewidth]{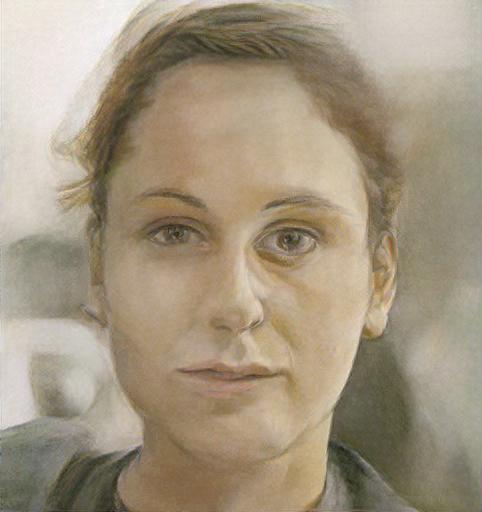}
    \includegraphics[width=0.23\linewidth]{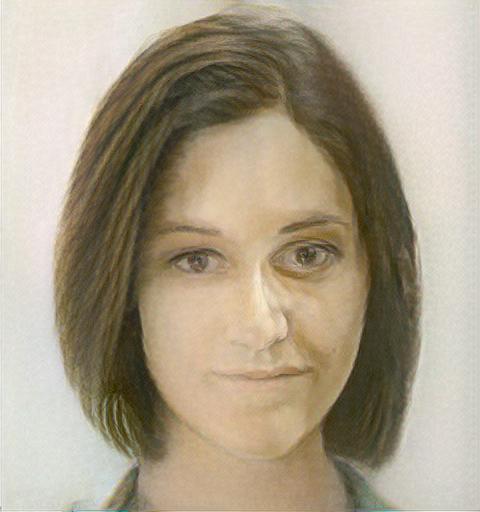}
   \\     
    \caption{Comparison of style transfer results (odd lines CNNMRF method, even lines our method).}
\label{fig:com_more01}       
\end{figure}

\begin{figure*}
\centering
    \includegraphics[width=0.15 \linewidth]{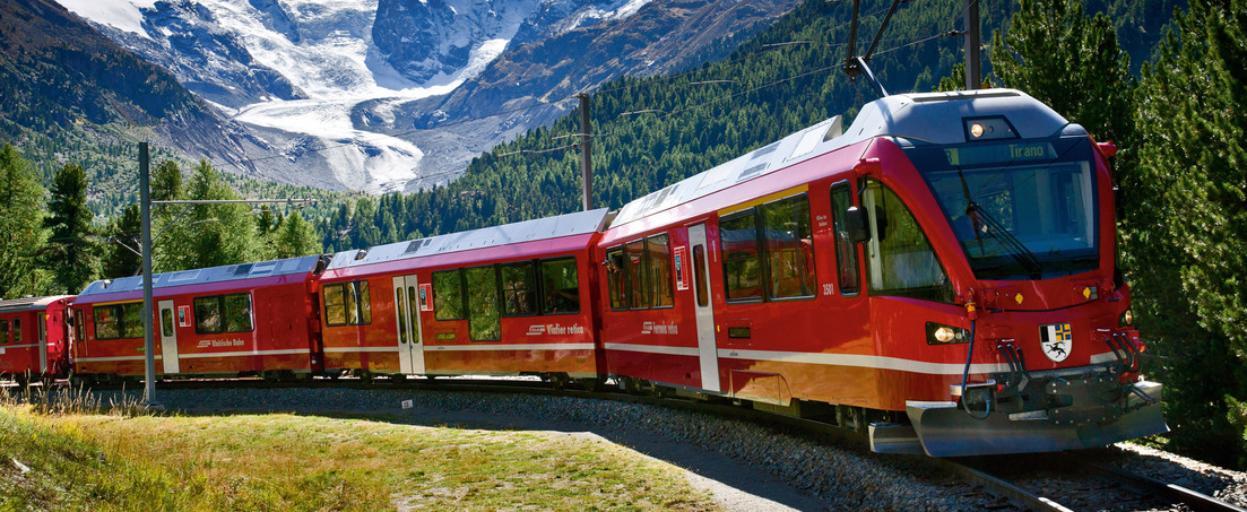}
    \includegraphics[width=0.15 \linewidth]{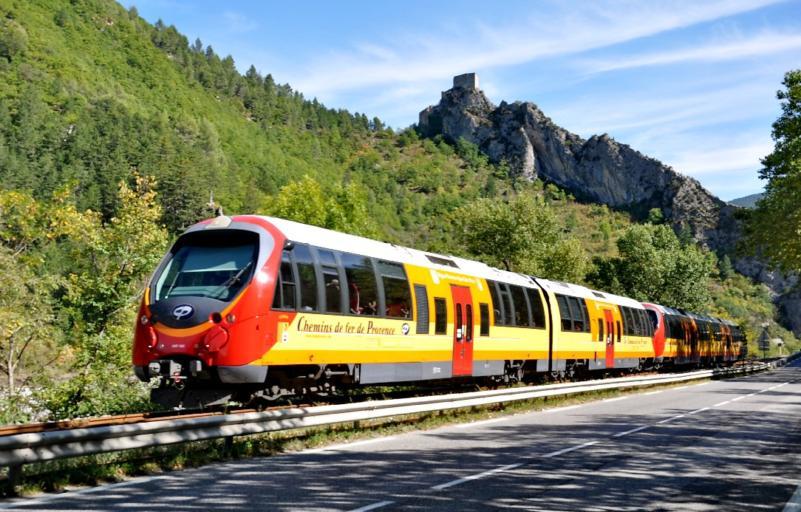}
    \includegraphics[width=0.15 \linewidth]{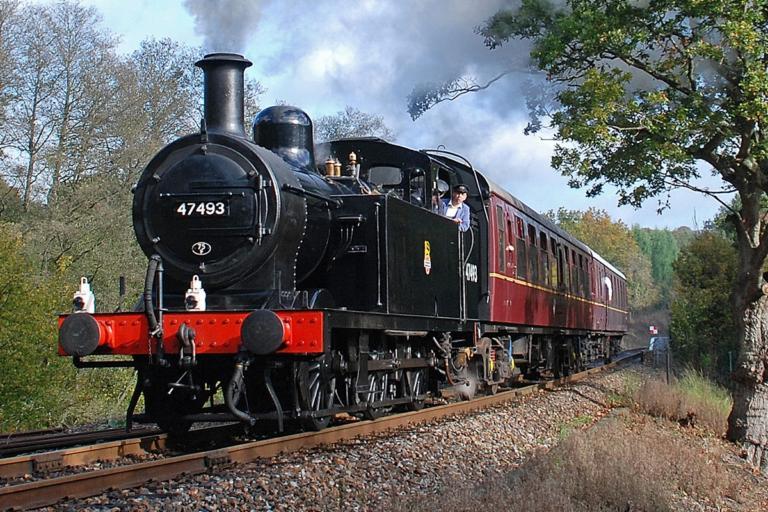} 
    \includegraphics[width=0.15 \linewidth]{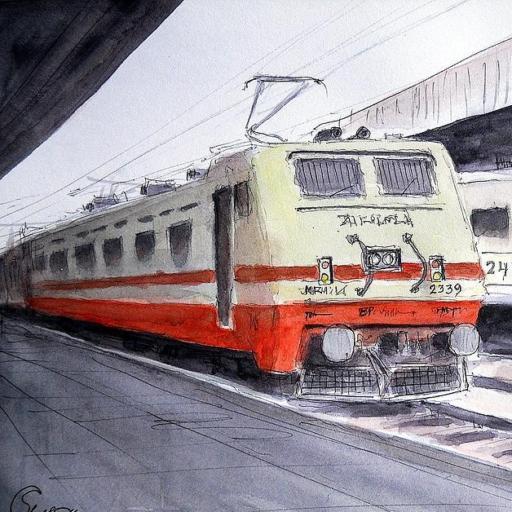}    
    \includegraphics[width=0.15 \linewidth]{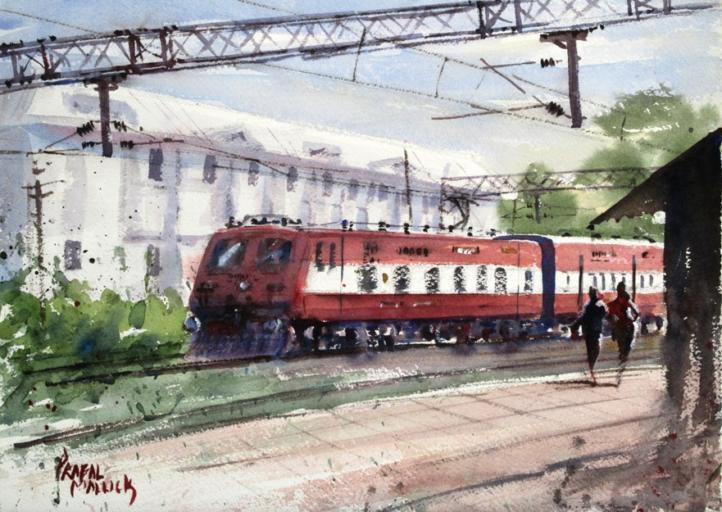}  
    \includegraphics[width=0.15 \linewidth]{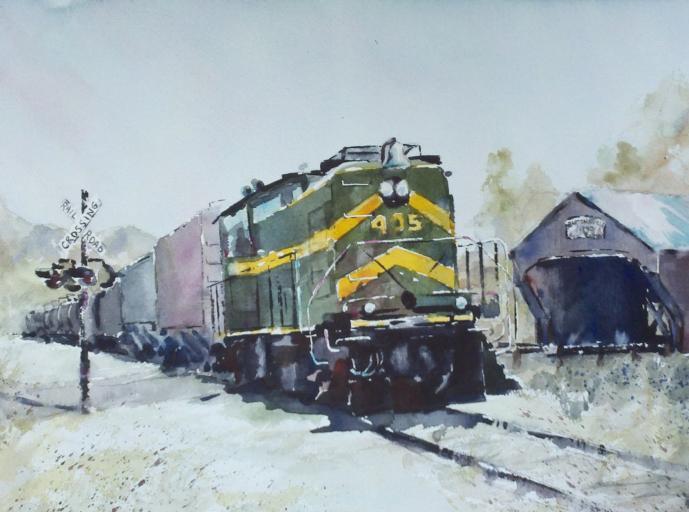}    
 \\ 
    \subfigure[]{\includegraphics[width=0.15 \linewidth]{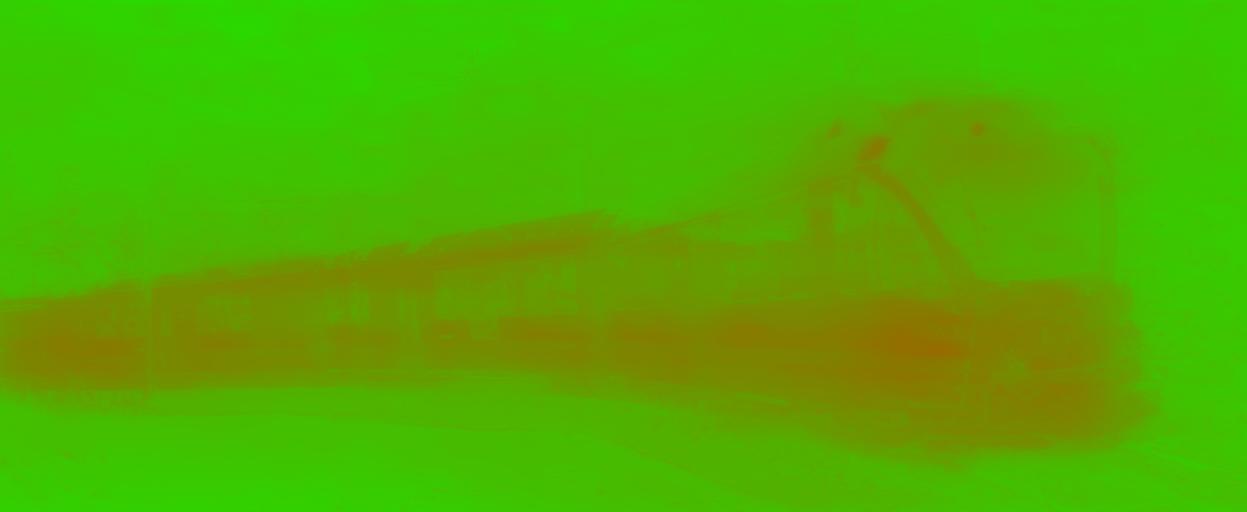}} 
    \subfigure[]{\includegraphics[width=0.15 \linewidth]{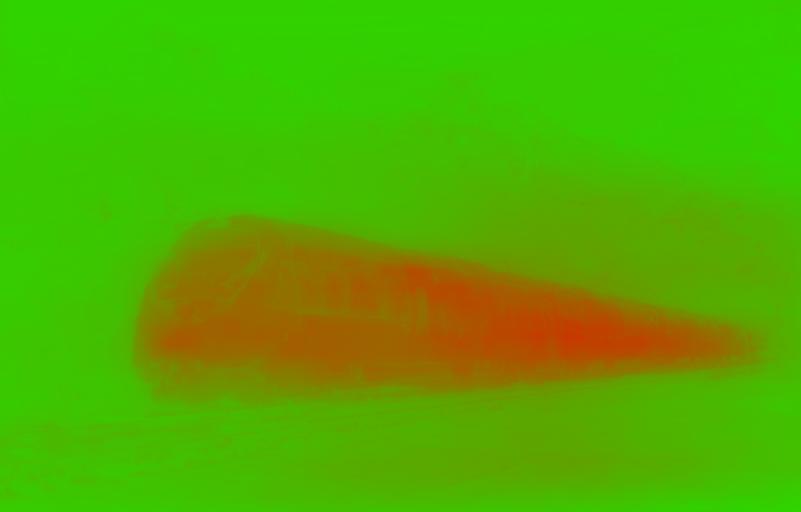}}     
    \subfigure[]{\includegraphics[width=0.15 \linewidth]{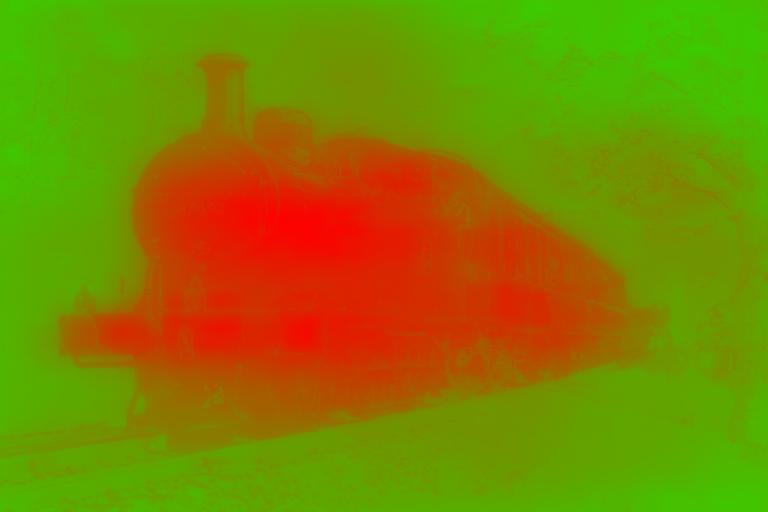}}
    \subfigure[]{\includegraphics[width=0.15\linewidth]{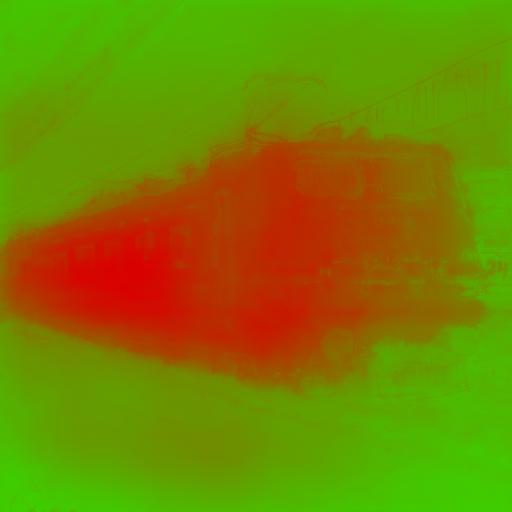}}
    \subfigure[]{\includegraphics[width=0.15\linewidth]{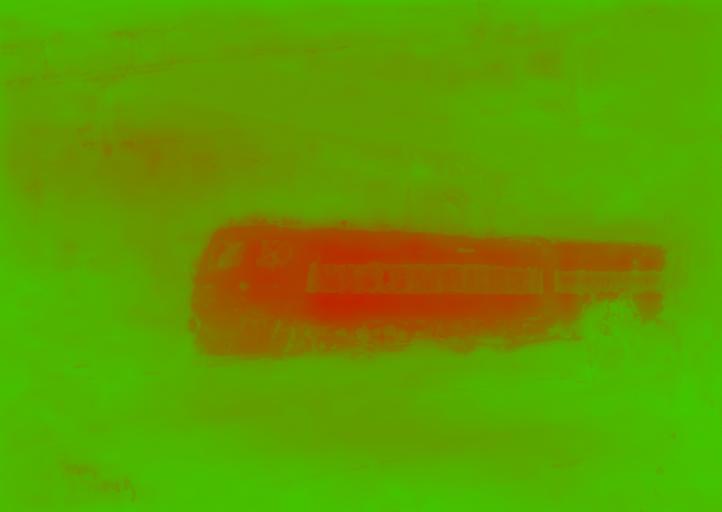}}    
    \subfigure[]{\includegraphics[width=0.15\linewidth]{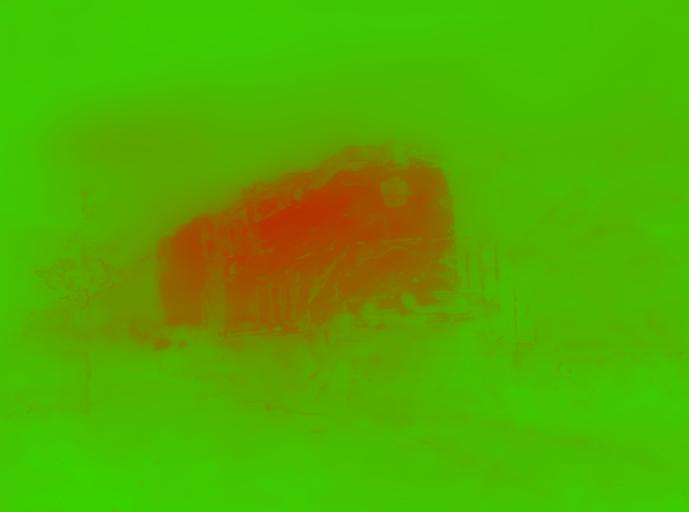}}
    
\caption{Content (a--c) and style (d--f) images and their soft masks.}  
\label{fig:content_trains}    
\end{figure*}

\begin{figure*}
\centering
    \includegraphics[width=0.153\linewidth]{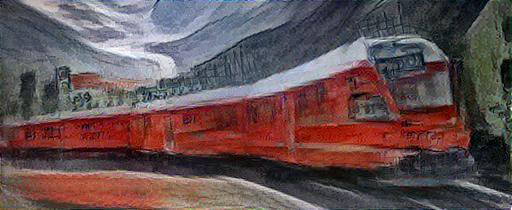}  
    \includegraphics[width=0.153\linewidth]{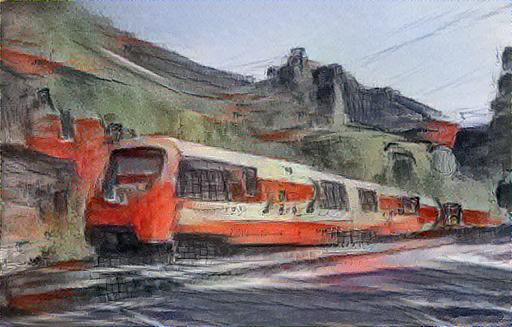}
    \includegraphics[width=0.153\linewidth]{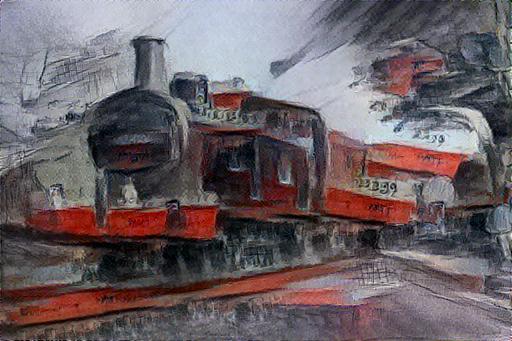} 
    \includegraphics[width=0.153\linewidth]{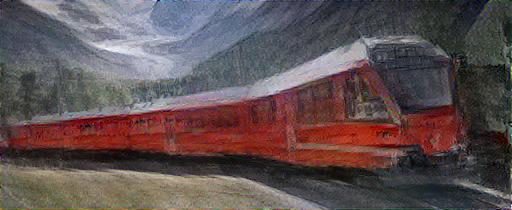}
    \includegraphics[width=0.153\linewidth]{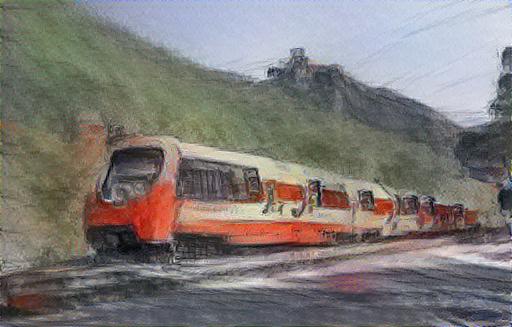} 
    \includegraphics[width=0.153\linewidth]{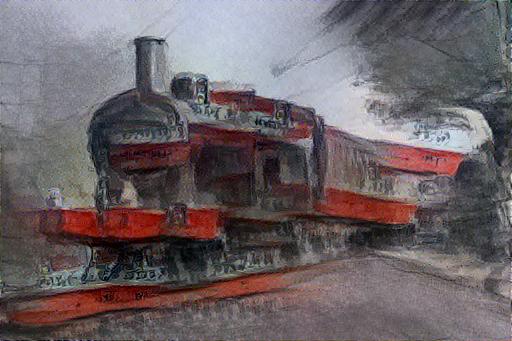}\\

    \includegraphics[width=0.153\linewidth]{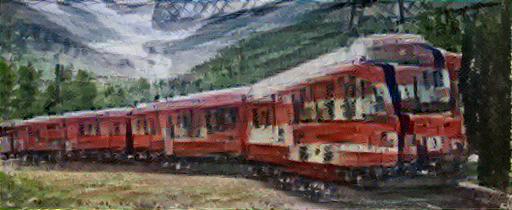}  
    \includegraphics[width=0.153\linewidth]{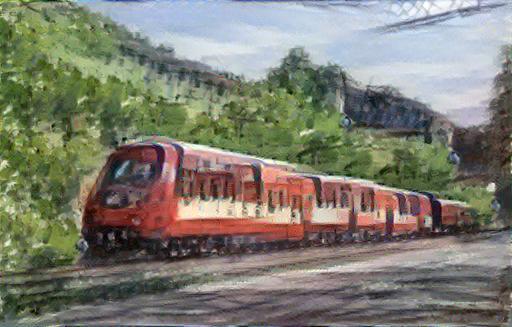}
    \includegraphics[width=0.153\linewidth]{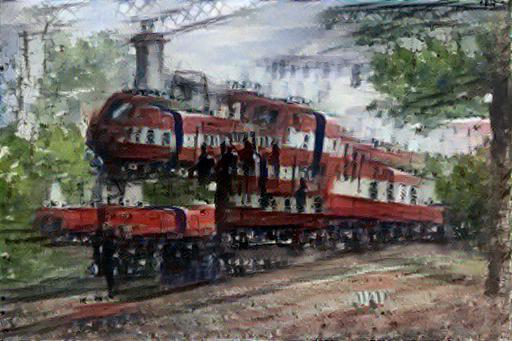} 
    \includegraphics[width=0.153\linewidth]{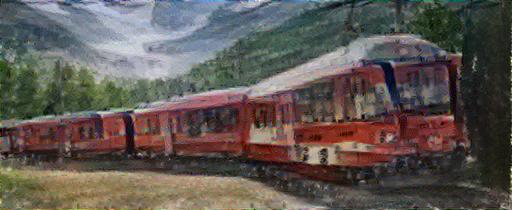}
    \includegraphics[width=0.153\linewidth]{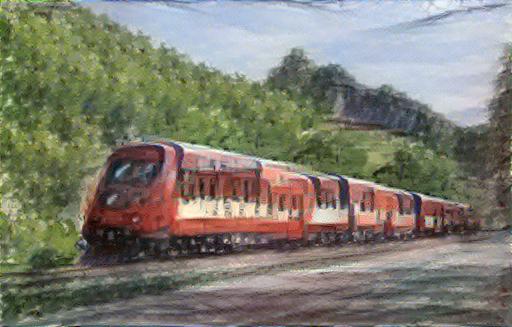}   
    \includegraphics[width=0.153\linewidth]{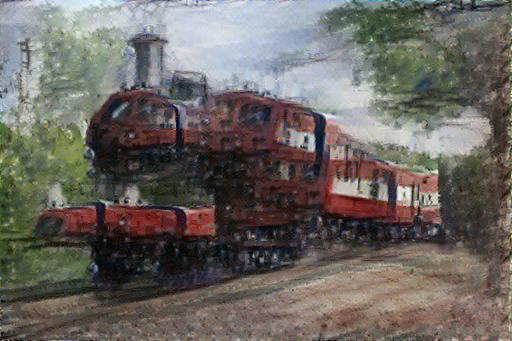}\\

    \includegraphics[width=0.153\linewidth]{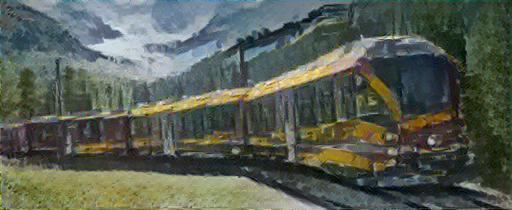}  
    \includegraphics[width=0.153\linewidth]{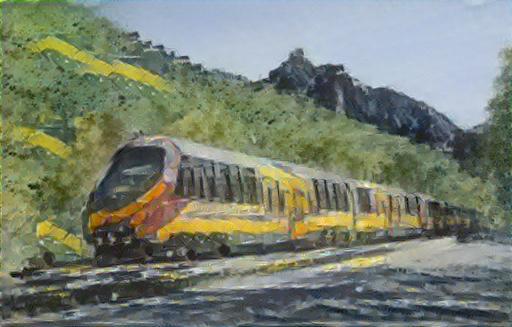}
    \includegraphics[width=0.153\linewidth]{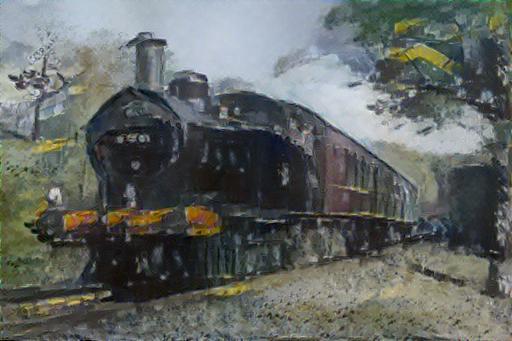} 
    \includegraphics[width=0.153\linewidth]{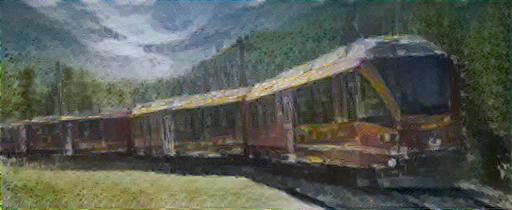}
    \includegraphics[width=0.153\linewidth]{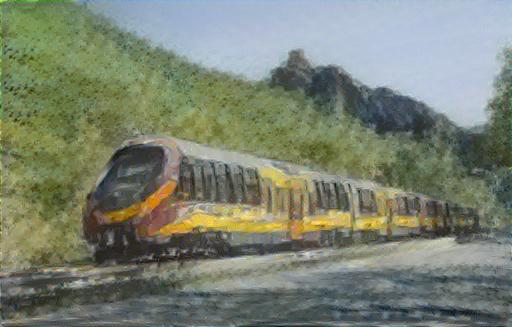} 
    \includegraphics[width=0.153\linewidth]{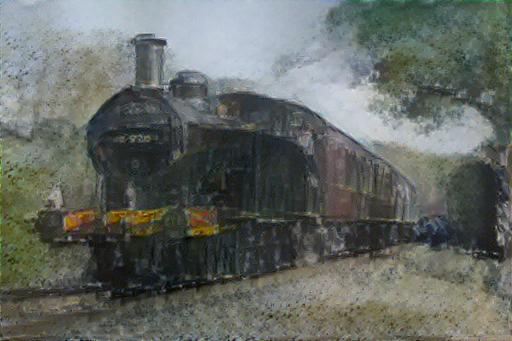}
  
\caption{Style transfer comparison (columns 1--3 CNNMRF method, columns 4--6 our method).}
\label{fig:com_trains}       
\end{figure*}

\begin{figure*}
\centering
    \includegraphics[width=0.15 \linewidth]{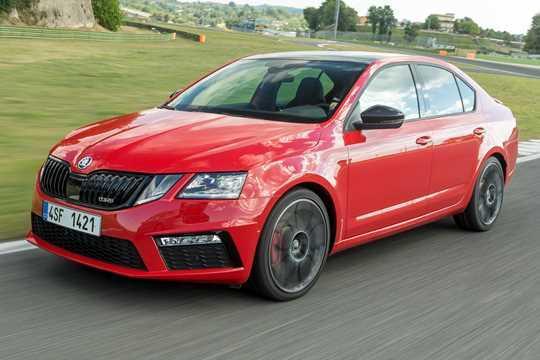} 
    \includegraphics[width=0.15 \linewidth]{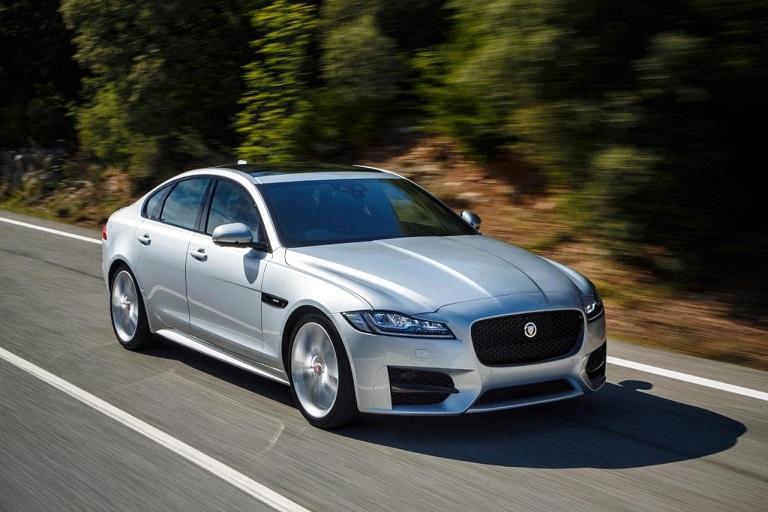} 
    \includegraphics[width=0.15 \linewidth]{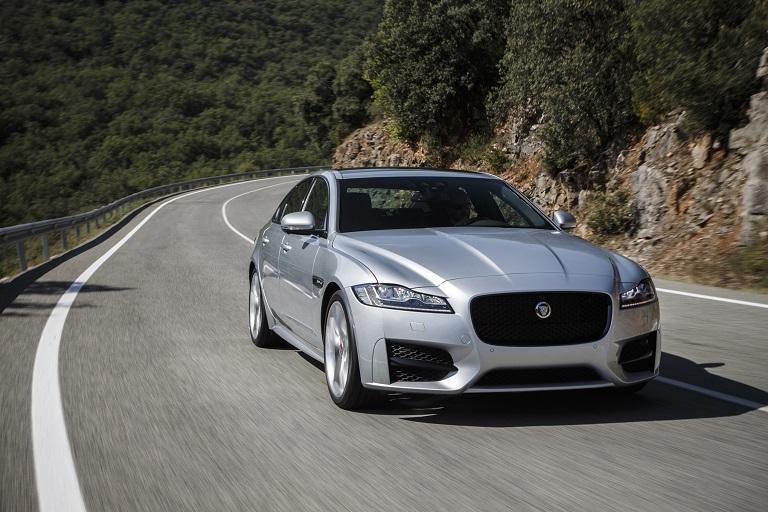}  
    \includegraphics[width=0.15 \linewidth]{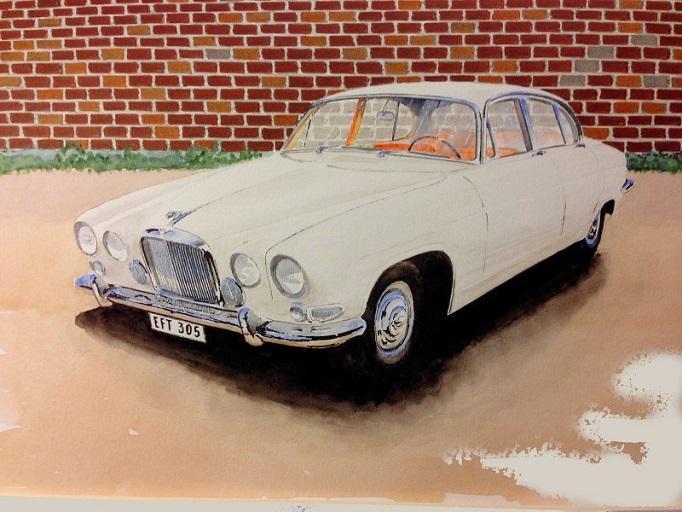}    
     \includegraphics[width=0.15 \linewidth]{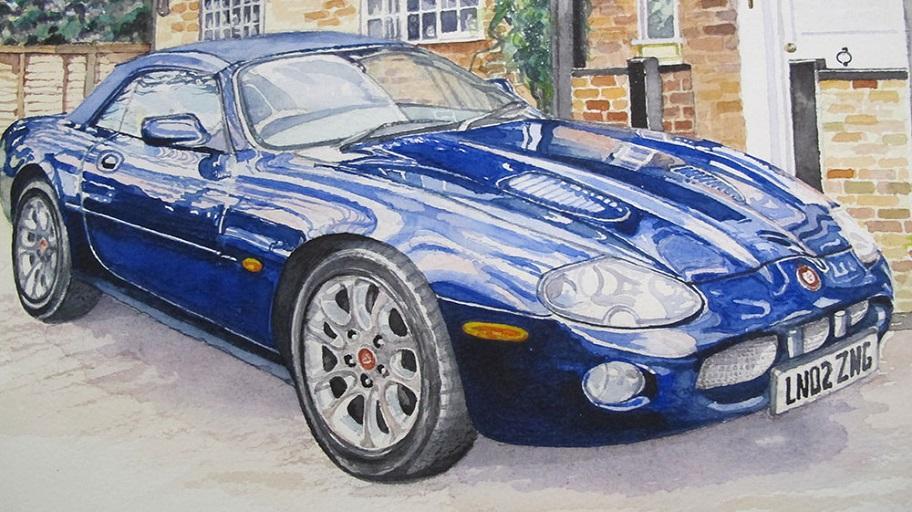}
     \includegraphics[width=0.15 \linewidth]{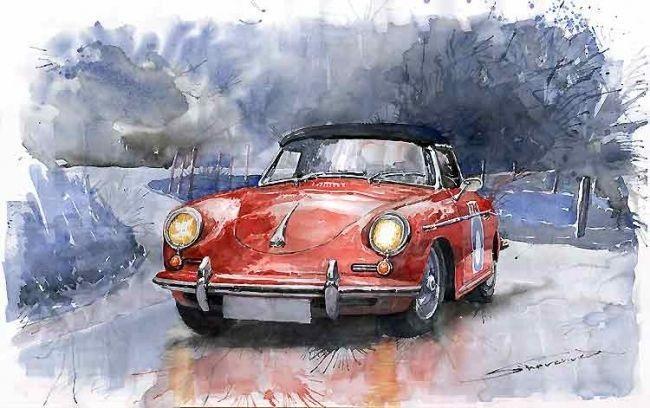} 
    \\ 
    \subfigure[]{\includegraphics[width=0.15 \linewidth]{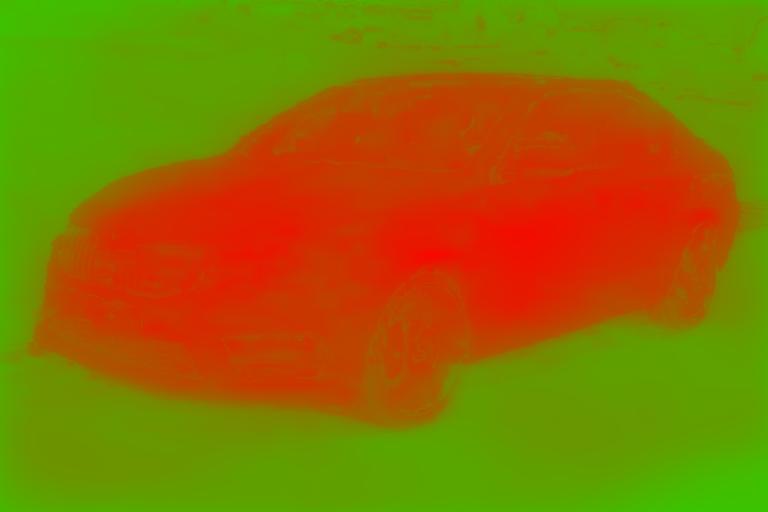}}
    \subfigure[]{\includegraphics[width=0.15 \linewidth]{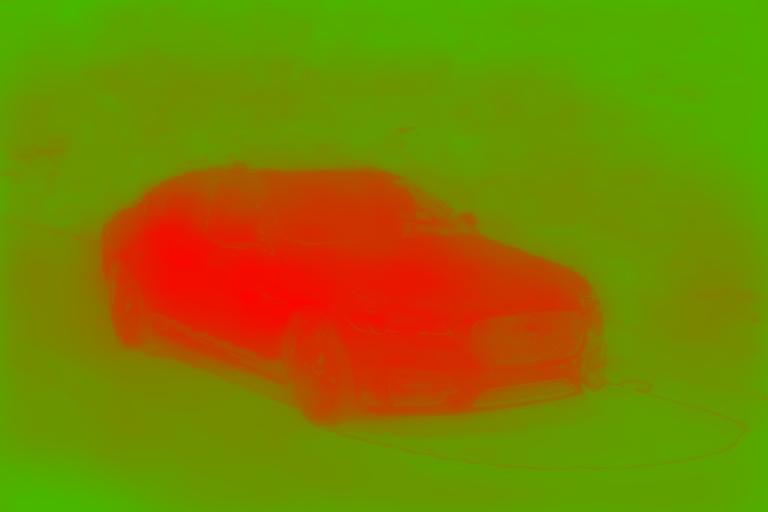}}      
    \subfigure[]{\includegraphics[width=0.15 \linewidth]{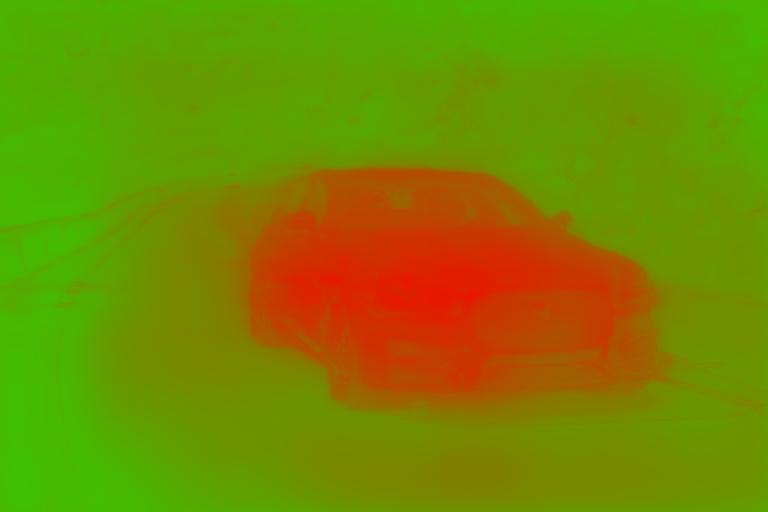}}   
    \subfigure[]{\includegraphics[width=0.15\linewidth]{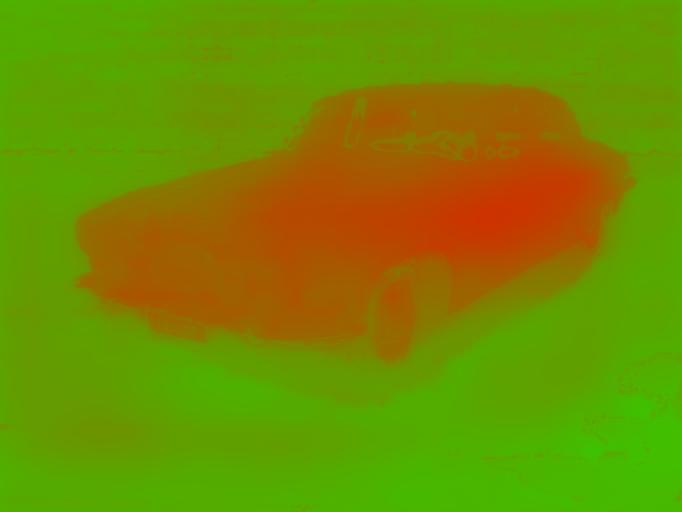}}
    \subfigure[]{\includegraphics[width=0.15\linewidth]{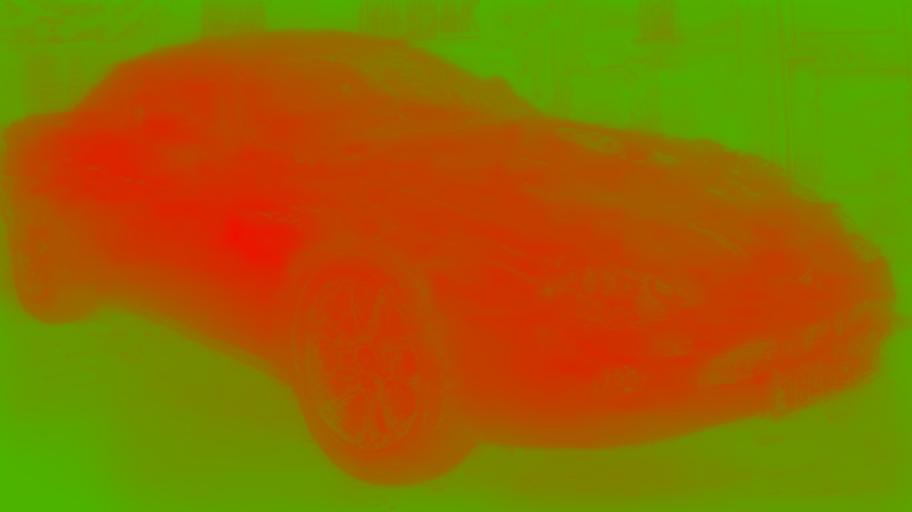}}
    \subfigure[]{\includegraphics[width=0.15\linewidth]{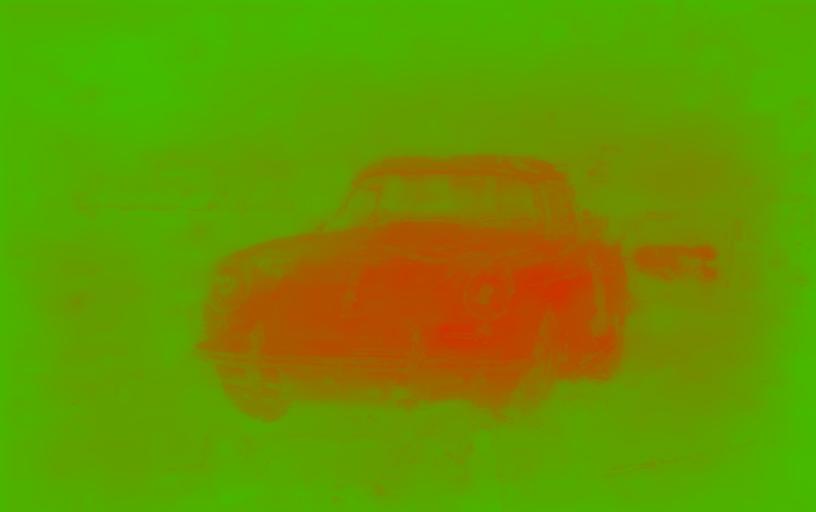}}
    
\caption{Content (a--c) and style (d--f) images and their soft masks.}  
\label{fig:content_car}    
\end{figure*}

\begin{figure*}
\centering
    \includegraphics[width=0.153\linewidth]{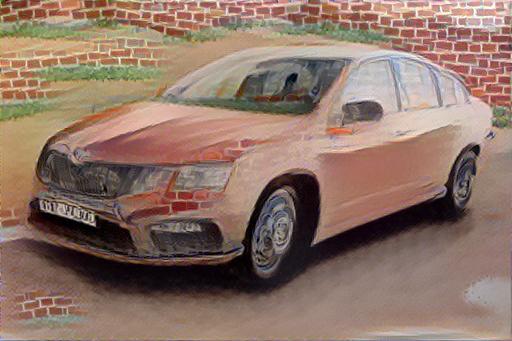}
    \includegraphics[width=0.153\linewidth]{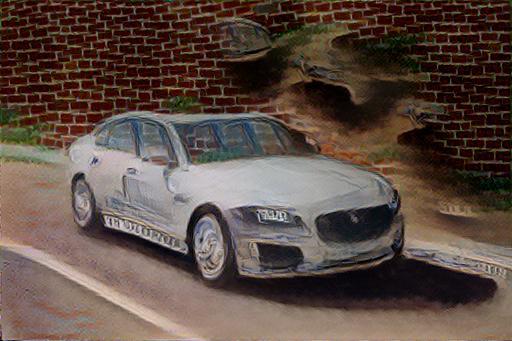}
    \includegraphics[width=0.153\linewidth]{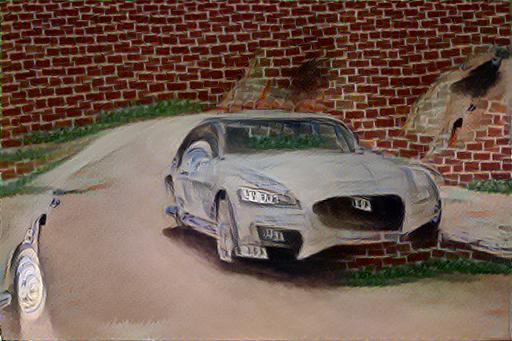} 
    \includegraphics[width=0.153\linewidth]{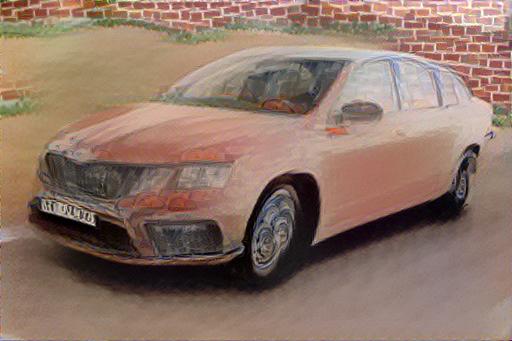}
    \includegraphics[width=0.153\linewidth]{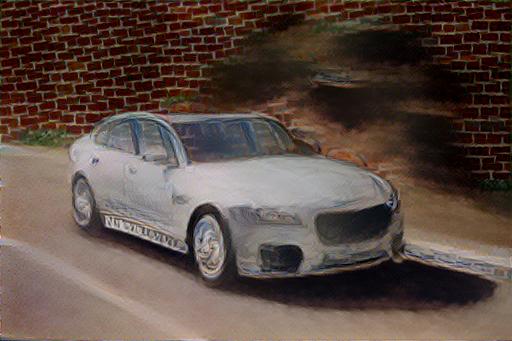}   
    \includegraphics[width=0.153\linewidth]{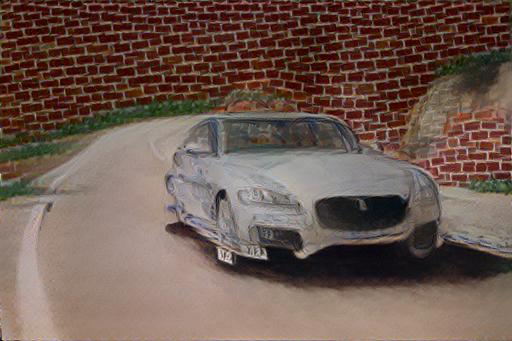}\\

    \includegraphics[width=0.153\linewidth]{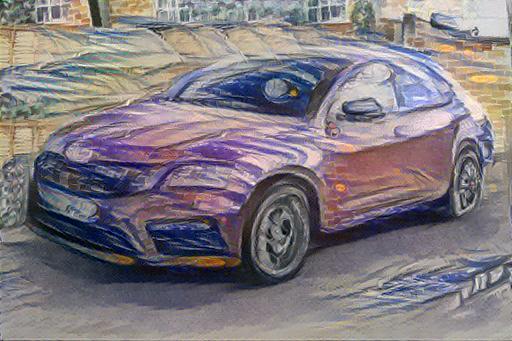}
    \includegraphics[width=0.153\linewidth]{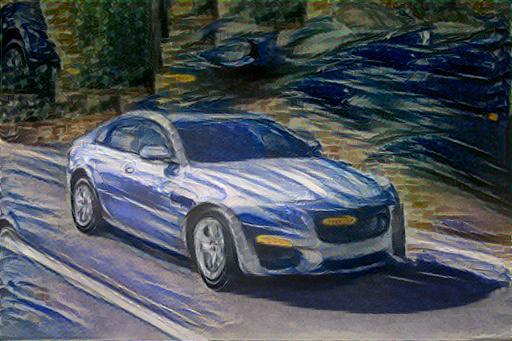}
    \includegraphics[width=0.153\linewidth]{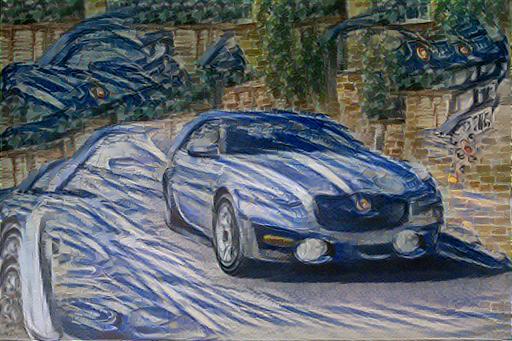} 
    \includegraphics[width=0.153\linewidth]{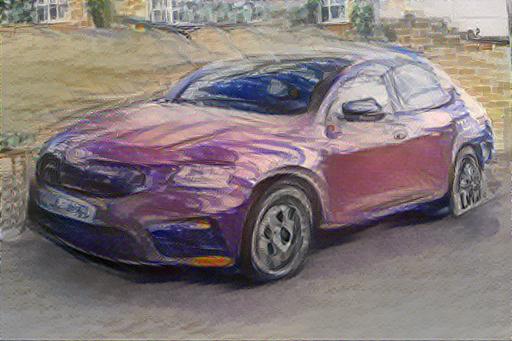}
    \includegraphics[width=0.153\linewidth]{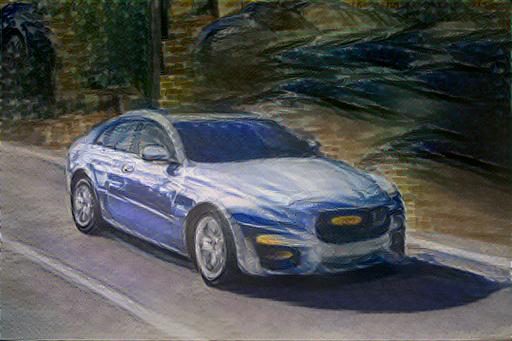}   
    \includegraphics[width=0.153\linewidth]{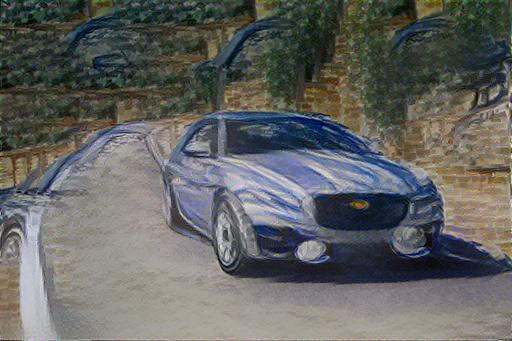}\\

    \includegraphics[width=0.153\linewidth]{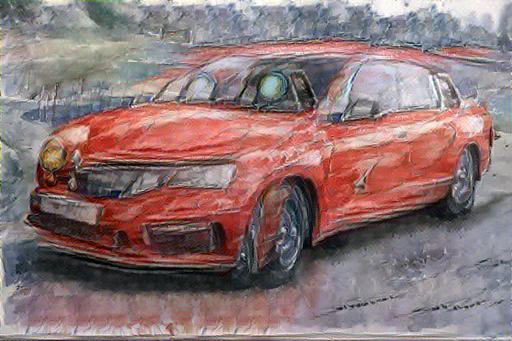}
    \includegraphics[width=0.153\linewidth]{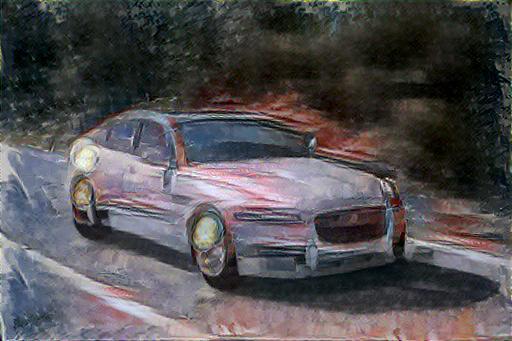}
    \includegraphics[width=0.153\linewidth]{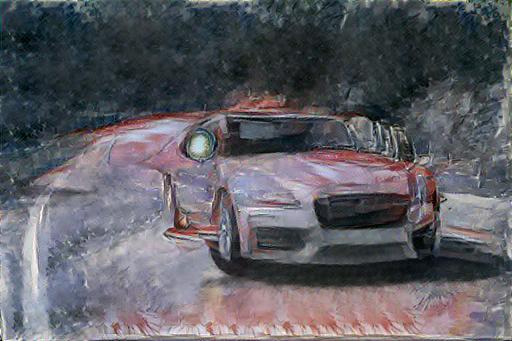} 
    \includegraphics[width=0.153\linewidth]{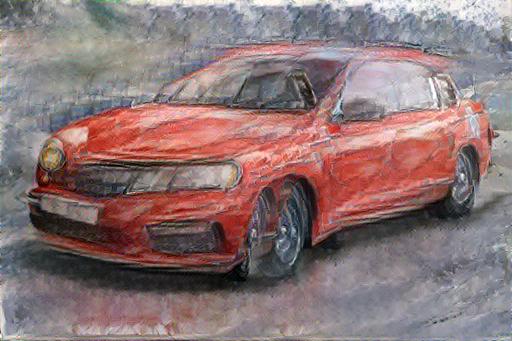}
    \includegraphics[width=0.153\linewidth]{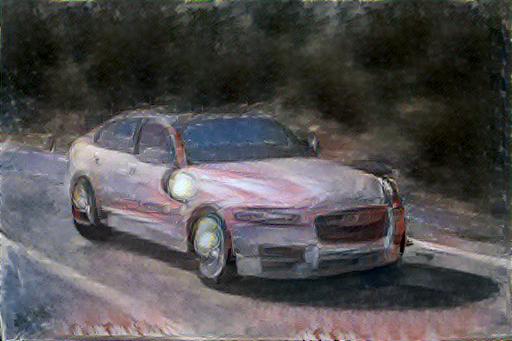}   
    \includegraphics[width=0.153\linewidth]{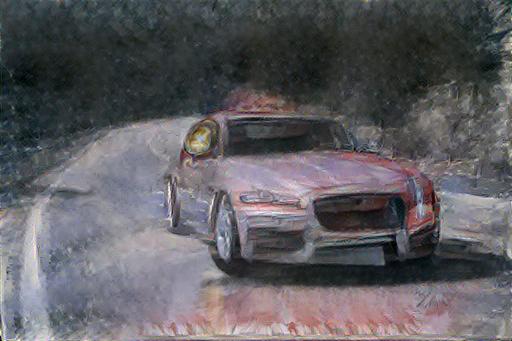}
        
\caption{Style transfer comparison (columns 1--3 CNNMRF method, columns 4--6 our method).}
\label{fig:com_cars}       
\end{figure*}

\begin{figure*}
\centering
    \includegraphics[width=0.15 \linewidth]{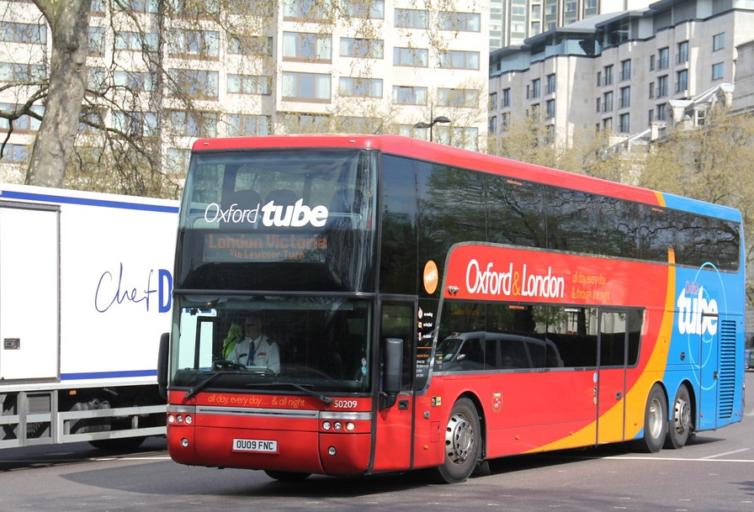} 
    \includegraphics[width=0.15 \linewidth]{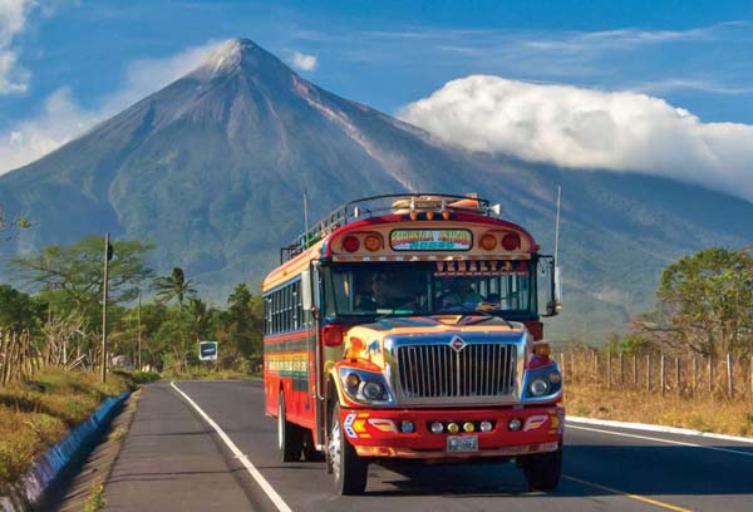} 
    \includegraphics[width=0.15 \linewidth]{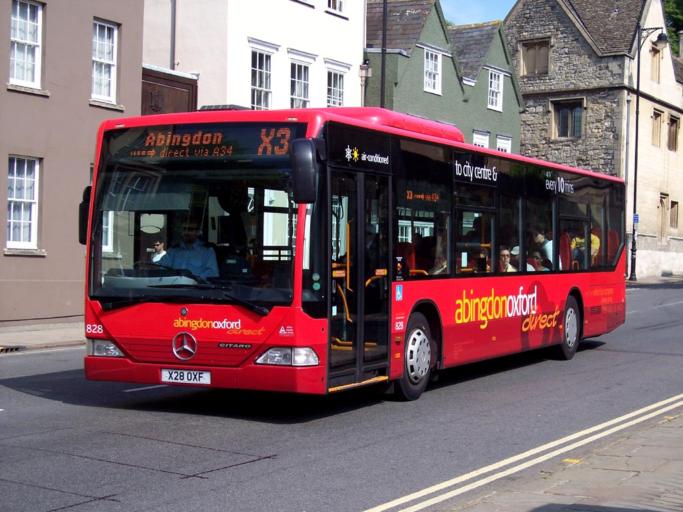}  
    \includegraphics[width=0.15 \linewidth]{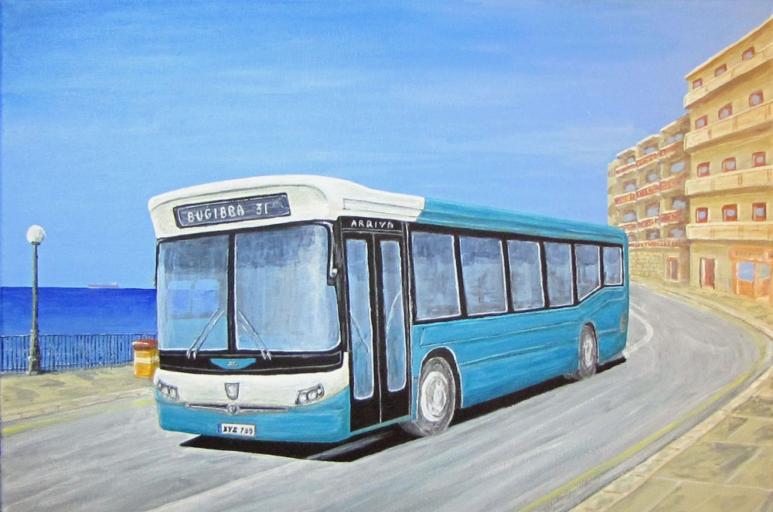}
    \includegraphics[width=0.15 \linewidth]{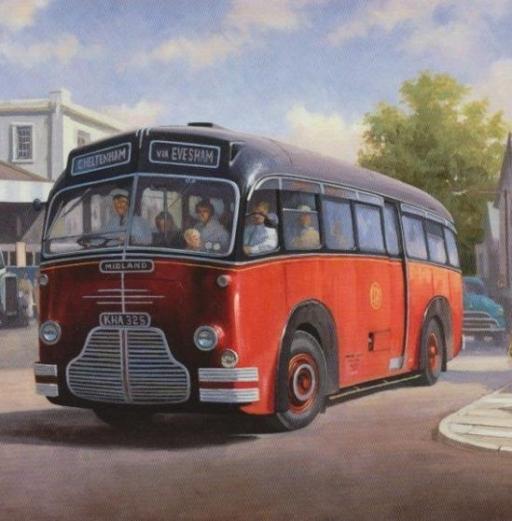}
    \includegraphics[width=0.15 \linewidth]{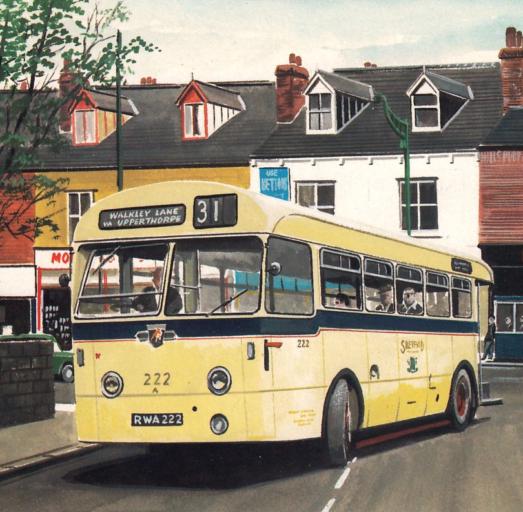}
    \\ 
    \subfigure[]{\includegraphics[width=0.15 \linewidth]{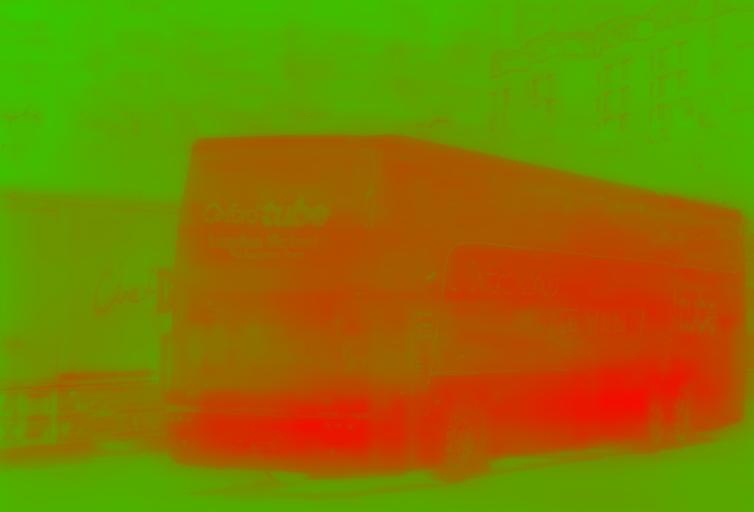}}      
    \subfigure[]{\includegraphics[width=0.15 \linewidth]{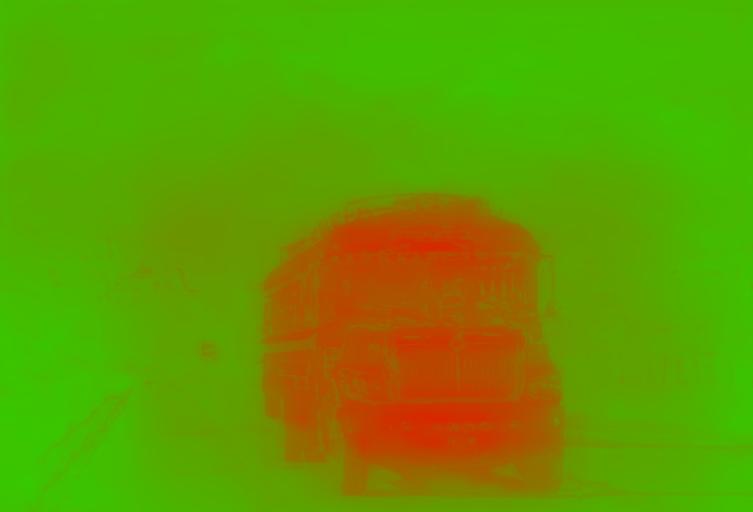}}   
    \subfigure[]{\includegraphics[width=0.15 \linewidth]{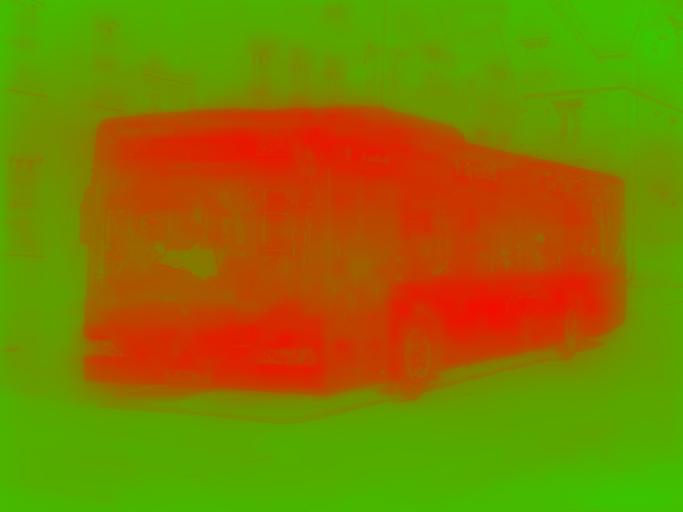}}  
    \subfigure[]{\includegraphics[width=0.15\linewidth]{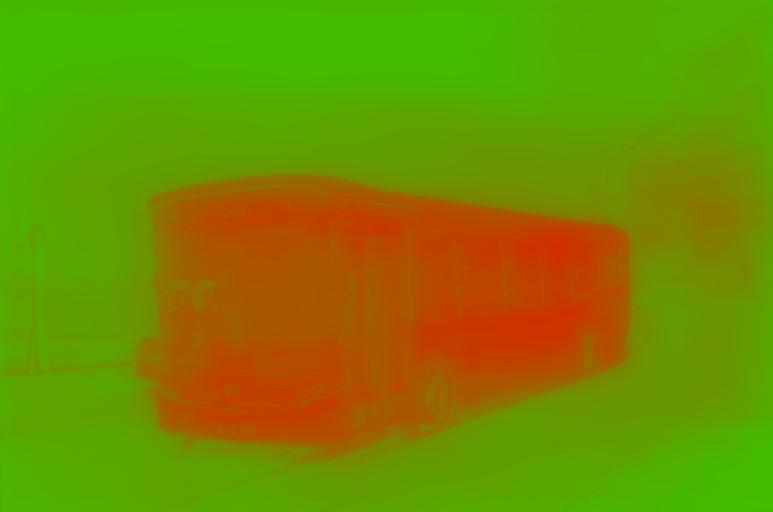}}
    \subfigure[]{\includegraphics[width=0.15\linewidth]{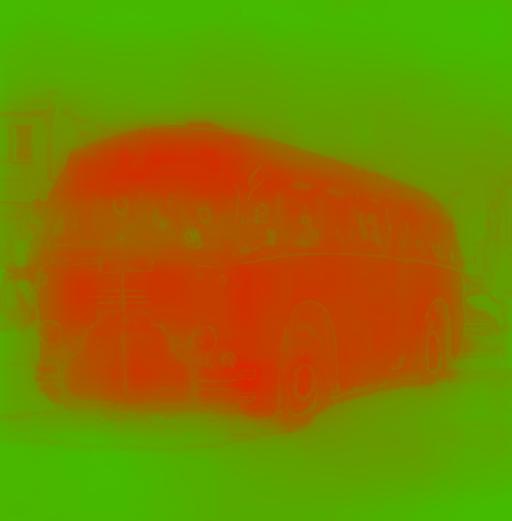}}
    \subfigure[]{\includegraphics[width=0.15\linewidth]{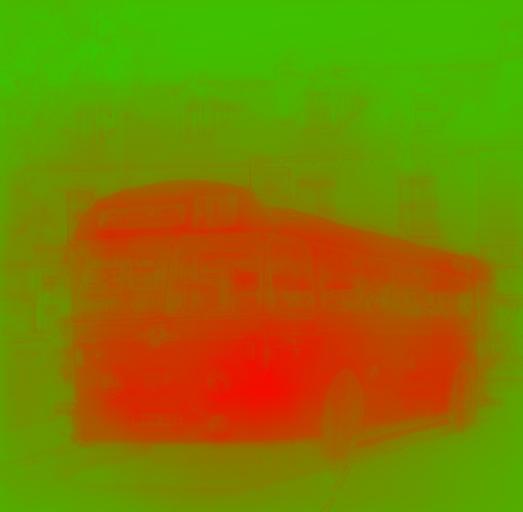}}       
\caption{Content (a--c) and style (d--f) images and their soft masks.}  
\label{fig:content_bus}    
\end{figure*}

\begin{figure*}
\centering
    \includegraphics[width=0.153\linewidth]{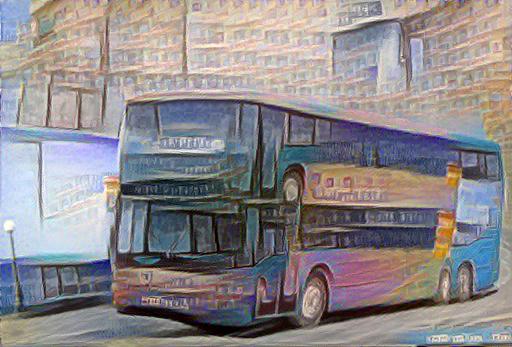}
    \includegraphics[width=0.153\linewidth]{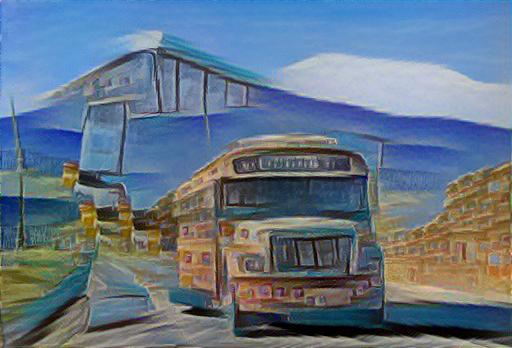} 
    \includegraphics[width=0.153\linewidth]{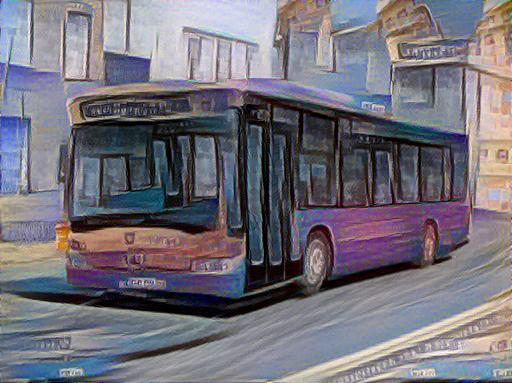}
    \includegraphics[width=0.153\linewidth]{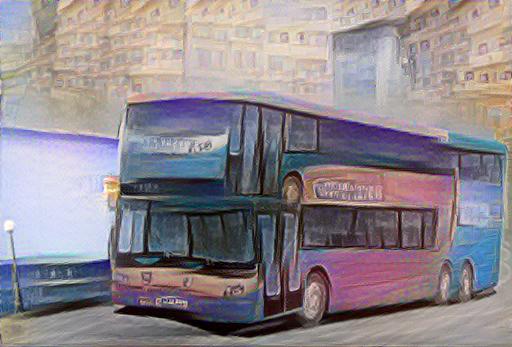}   
    \includegraphics[width=0.153\linewidth]{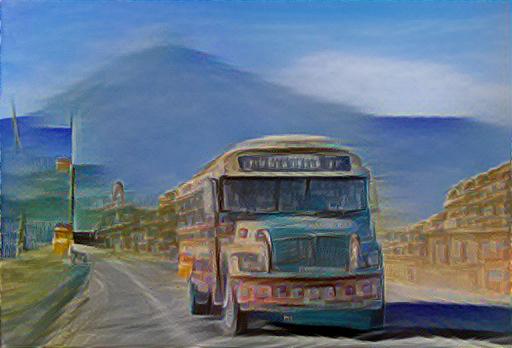}
    \includegraphics[width=0.153\linewidth]{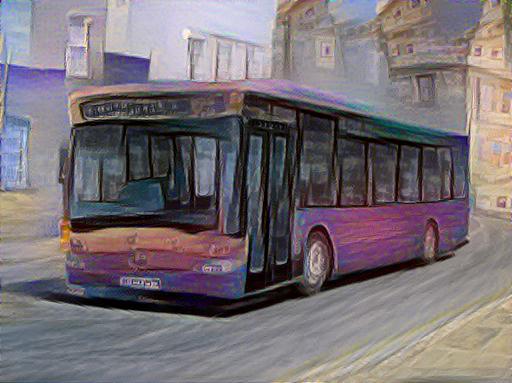}
    \\
    \includegraphics[width=0.153\linewidth]{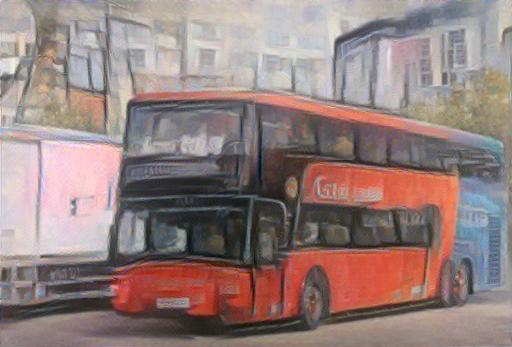}
    \includegraphics[width=0.153\linewidth]{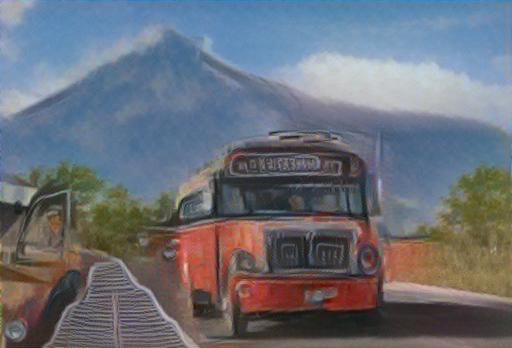} 
    \includegraphics[width=0.153\linewidth]{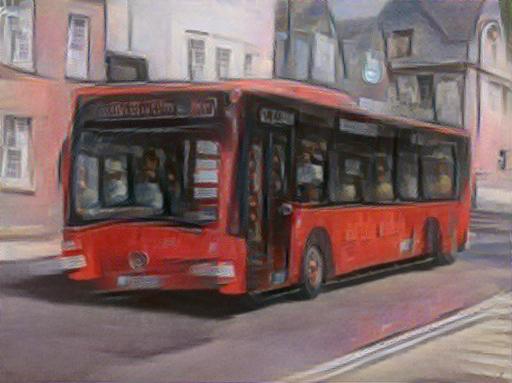}  
    \includegraphics[width=0.153\linewidth]{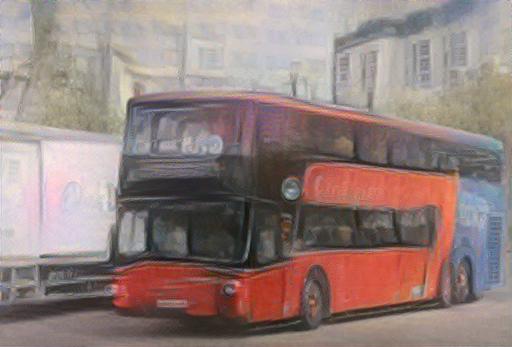}   
    \includegraphics[width=0.153\linewidth]{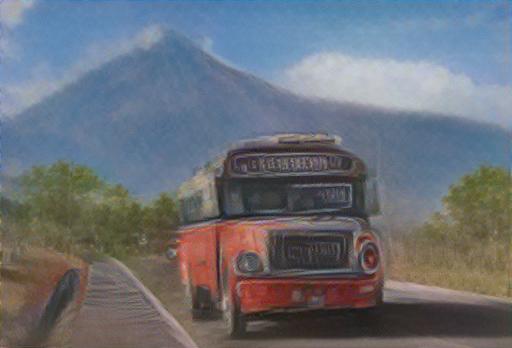}
    \includegraphics[width=0.153\linewidth]{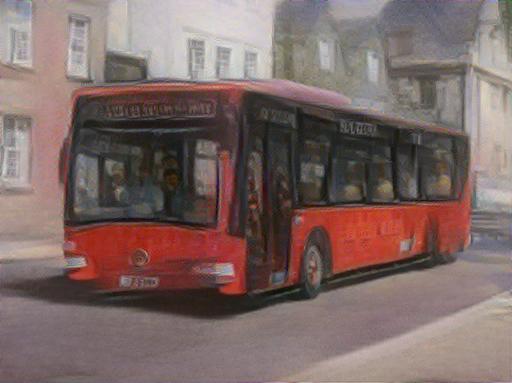}
    \\
    \includegraphics[width=0.153\linewidth]{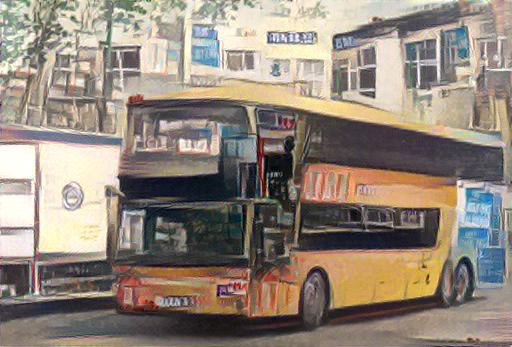}
    \includegraphics[width=0.153\linewidth]{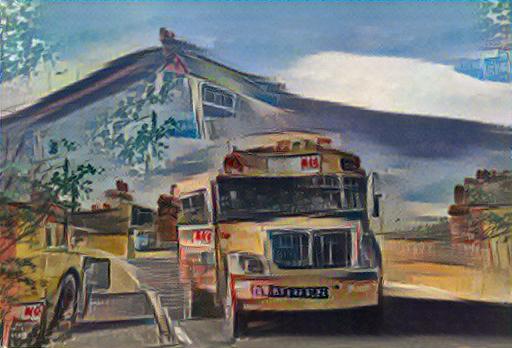} 
    \includegraphics[width=0.153\linewidth]{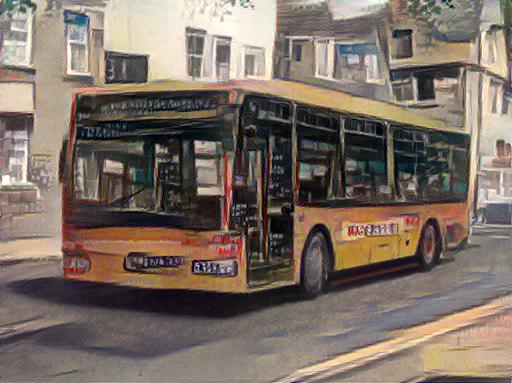}     
    \includegraphics[width=0.153\linewidth]{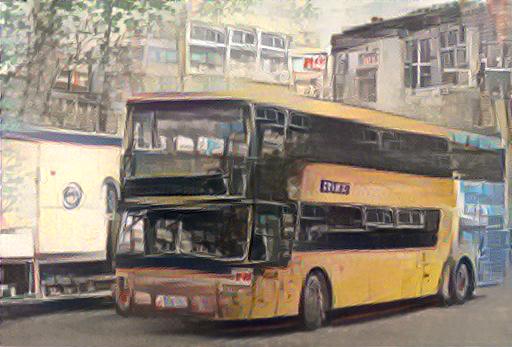}   
    \includegraphics[width=0.153\linewidth]{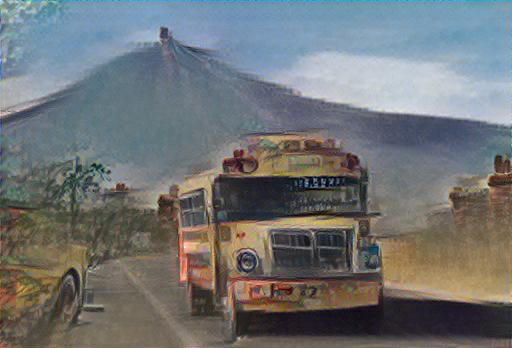}
    \includegraphics[width=0.153\linewidth]{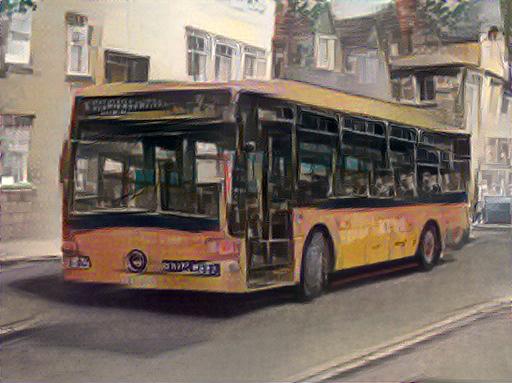}  
\caption{Style transfer comparison (columns 1--3 CNNMRF method, columns 4--6 our method).}
\label{fig:com_bus}       
\end{figure*}

\begin{figure*}
\centering
    \includegraphics[width=0.15 \linewidth]{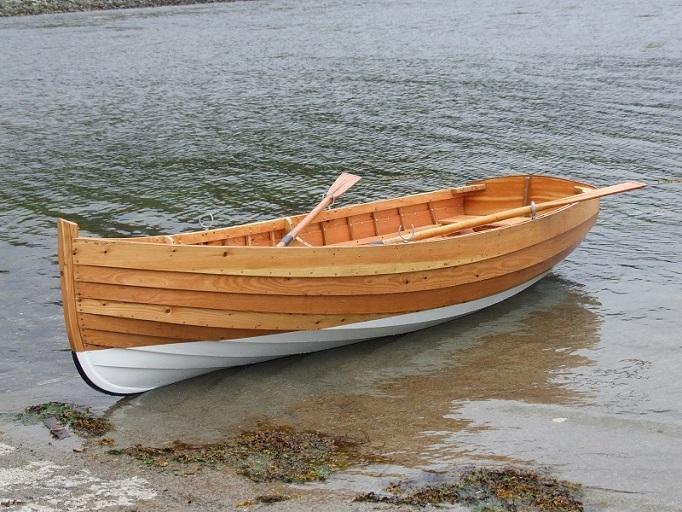} 
    \includegraphics[width=0.15 \linewidth]{boat_13.jpg} 
    \includegraphics[width=0.15 \linewidth]{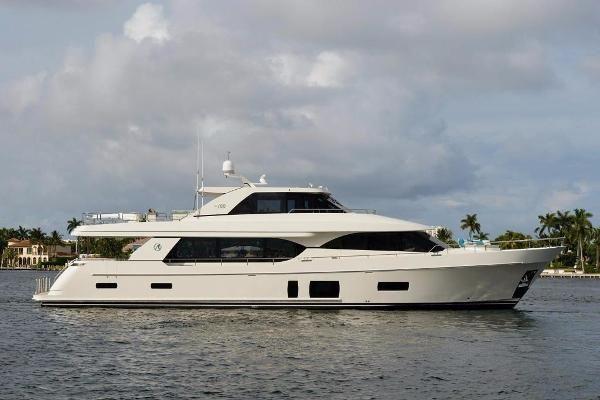} 
    \includegraphics[width=0.15 \linewidth]{style_boat_02.jpg}
    \includegraphics[width=0.15 \linewidth]{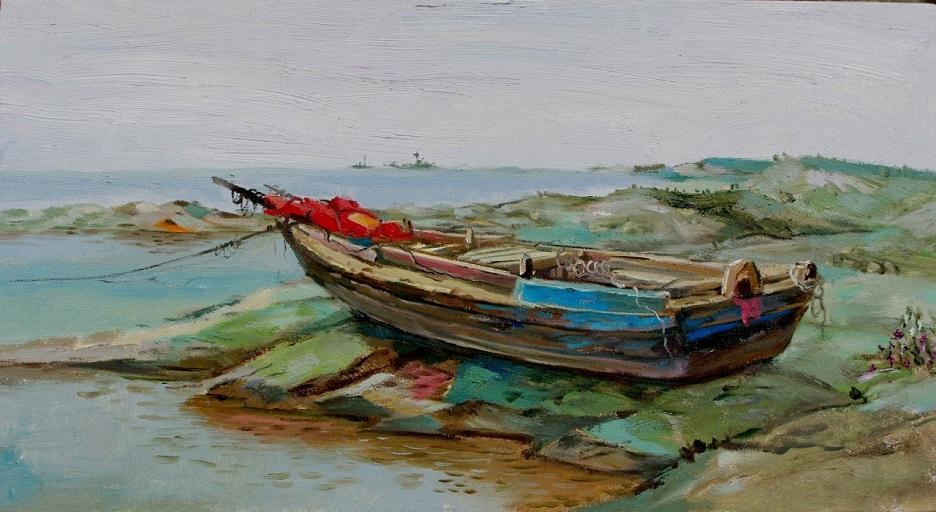}
     \includegraphics[width=0.15 \linewidth]{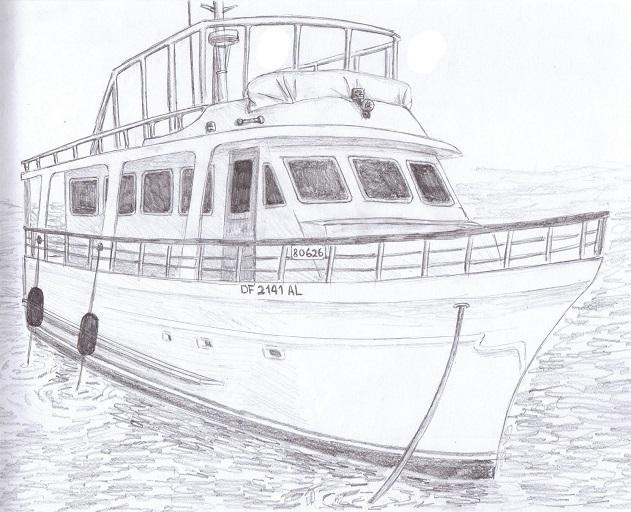} 
      \\
    \subfigure[]{\includegraphics[width=0.15 \linewidth]{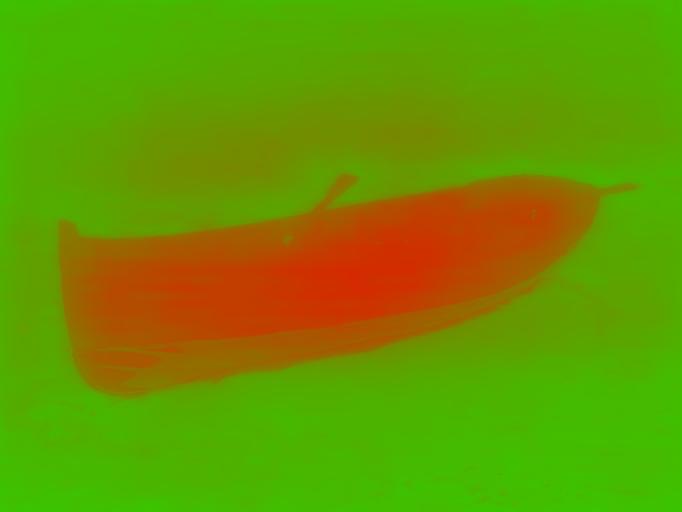}}  
    \subfigure[]{\includegraphics[width=0.15 \linewidth]{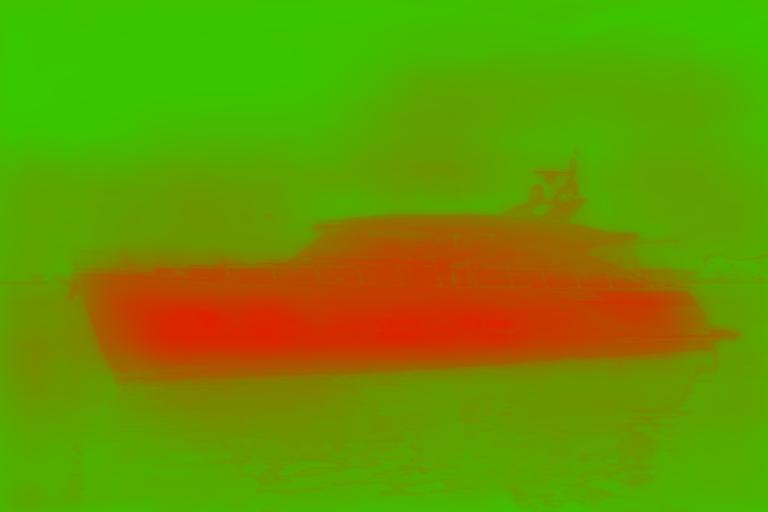}}      
    \subfigure[]{\includegraphics[width=0.15 \linewidth]{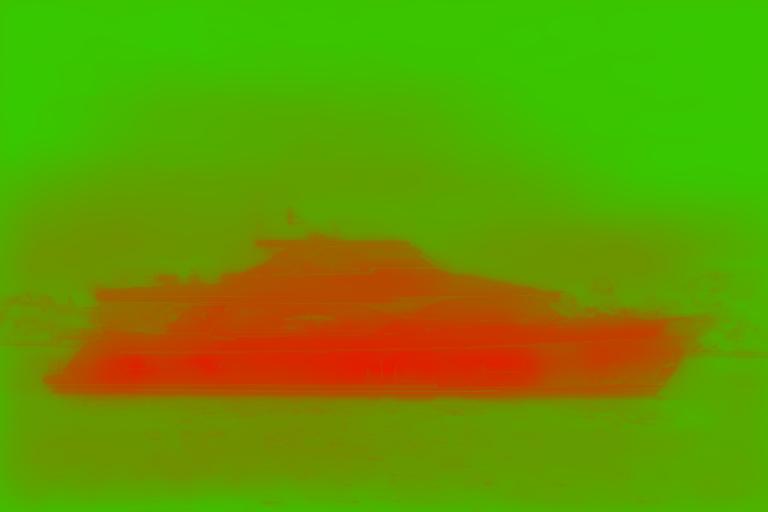}}   
    \subfigure[]{\includegraphics[width=0.15\linewidth]{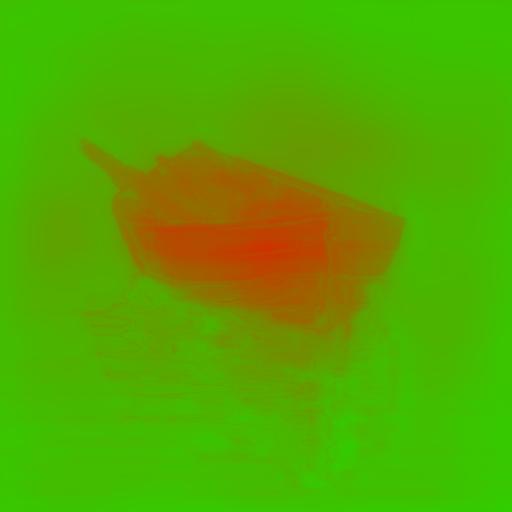}}
    \subfigure[]{\includegraphics[width=0.15\linewidth]{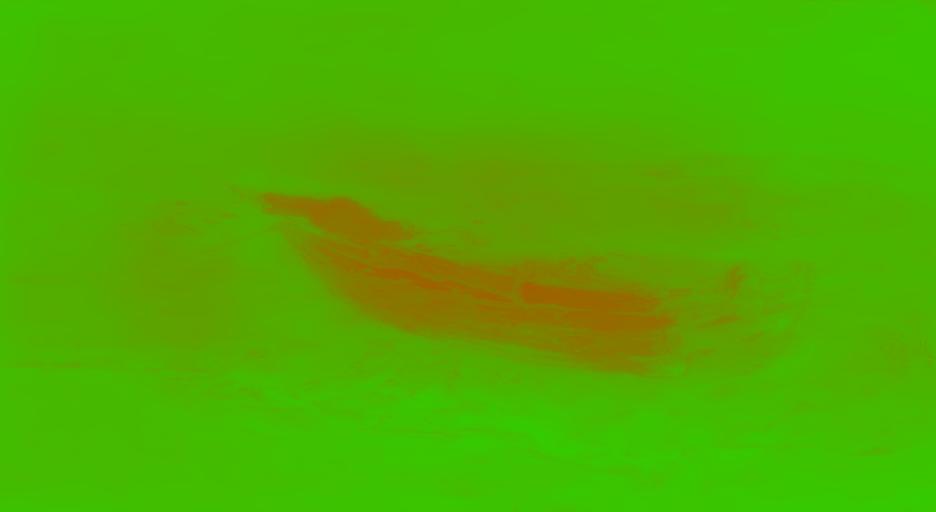}}
    \subfigure[]{\includegraphics[width=0.15\linewidth]{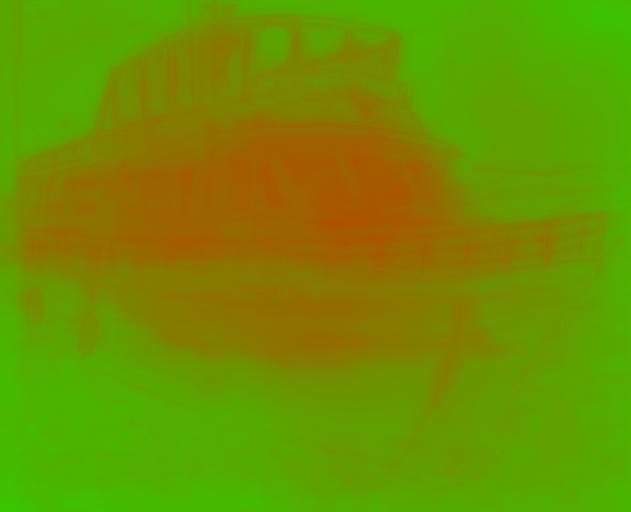}}
    
\caption{Content (a--c) and style (d--f) images and their soft masks.}  
\label{fig:content_boats}    
\end{figure*}

\begin{figure*}
\centering
    \includegraphics[width=0.152\linewidth]{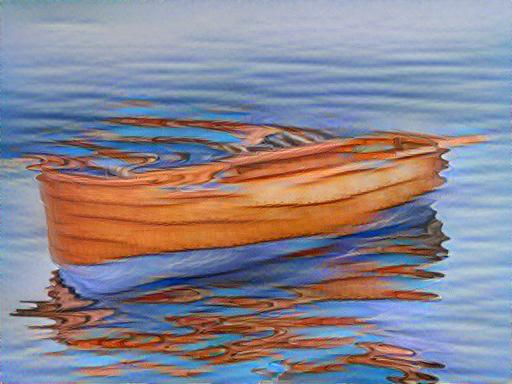}  
    \includegraphics[width=0.152\linewidth]{boat_13_TO_style_boat_02_res_3_100_cnnmrf.jpg}
    \includegraphics[width=0.152\linewidth]{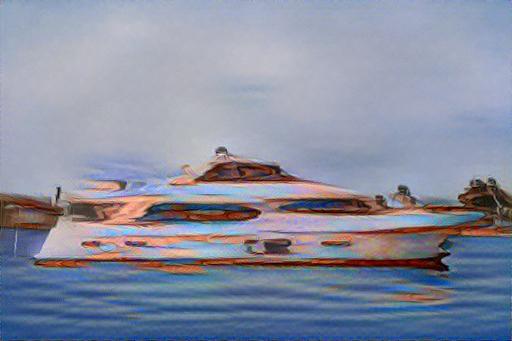}
    \includegraphics[width=0.152\linewidth]{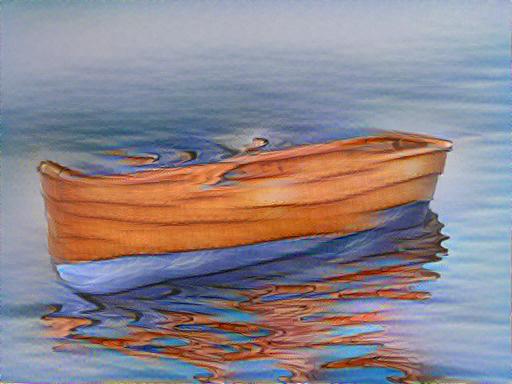}
    \includegraphics[width=0.152\linewidth]{boat_13_TO_style_boat_02_res_3_100_pb_hh_sem_20.jpg}   
    \includegraphics[width=0.152\linewidth]{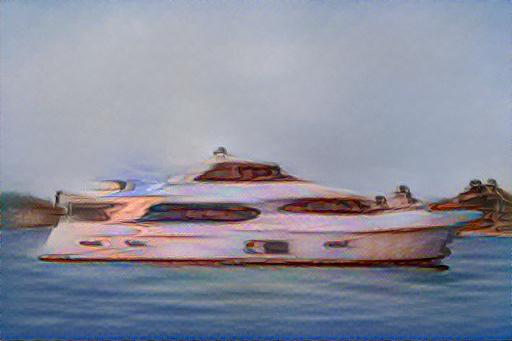}
\\
    \includegraphics[width=0.152\linewidth]{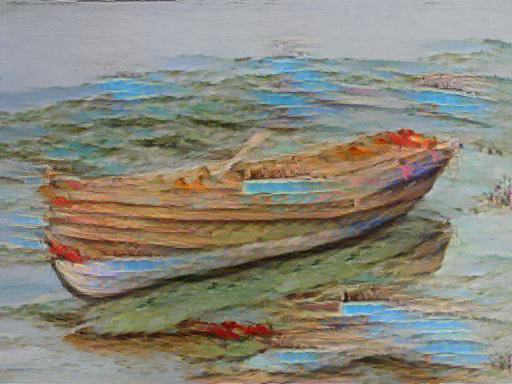}  
    \includegraphics[width=0.152\linewidth]{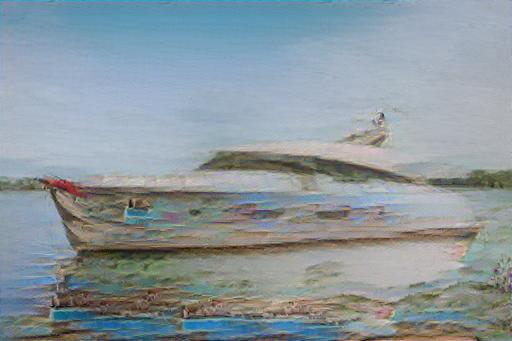}
    \includegraphics[width=0.152\linewidth]{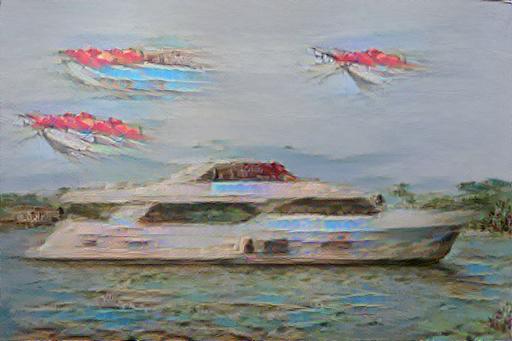} 
    \includegraphics[width=0.152\linewidth]{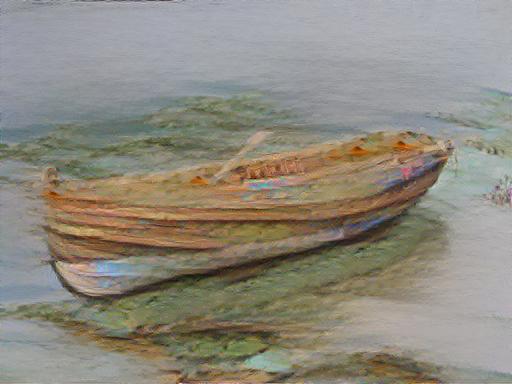}  
    \includegraphics[width=0.152\linewidth]{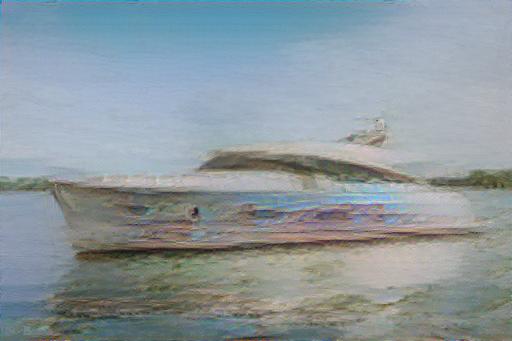}
    \includegraphics[width=0.152\linewidth]{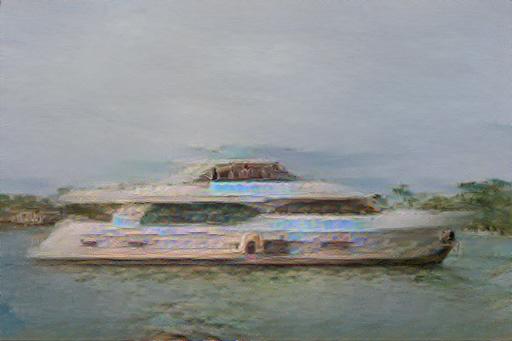} 
\\
    \includegraphics[width=0.152\linewidth]{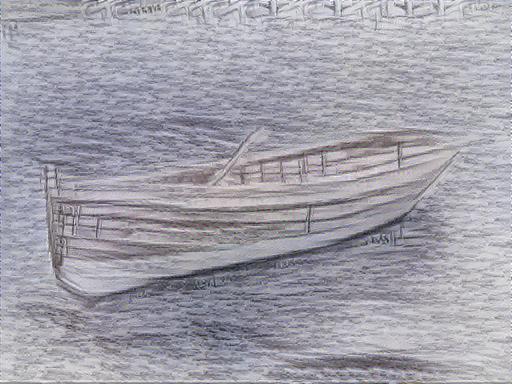}  
    \includegraphics[width=0.152\linewidth]{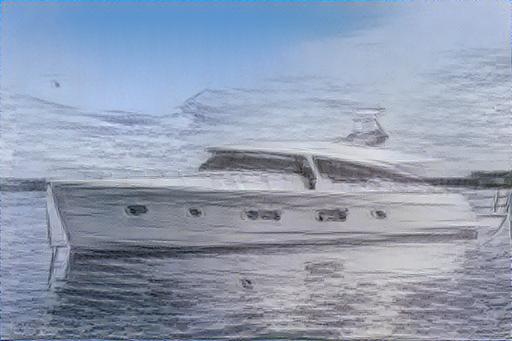}
    \includegraphics[width=0.152\linewidth]{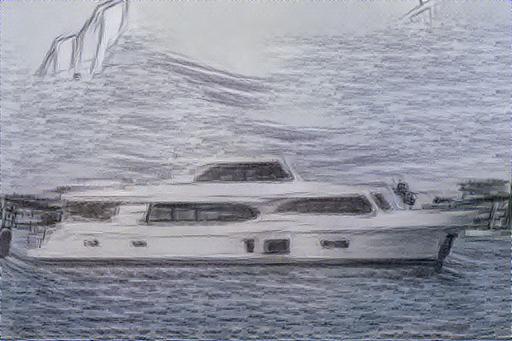} 
    \includegraphics[width=0.152\linewidth]{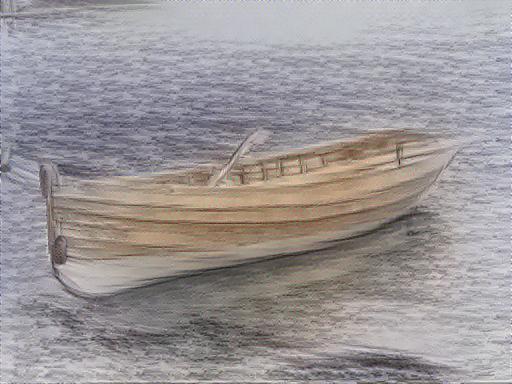}
    \includegraphics[width=0.152\linewidth]{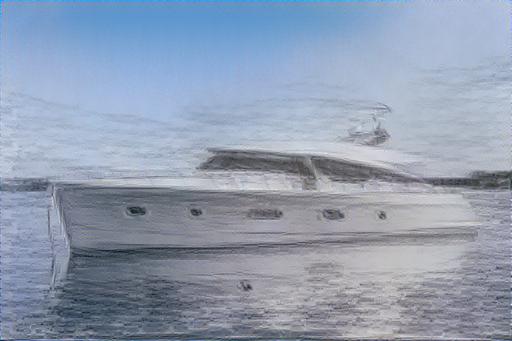}   
    \includegraphics[width=0.152\linewidth]{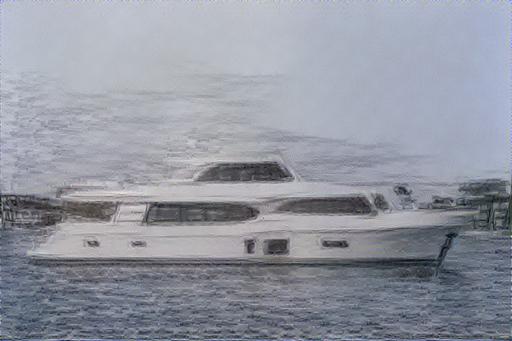}

\caption{Style transfer comparison (columns 1--3 CNNMRF method, columns 4--6 our method).}
\label{fig:com_boats}       
\end{figure*}

Figures \ref{fig:com_more02} and \ref{fig:com_more01} show style transfer applied separately to photographs of men and women. We transfer the style of each style image to each content image. We can see from figures~\ref{fig:com_more02} and~\ref{fig:com_more01} that our method can achieve better results than the CNNMRF method and avoid errors in applying style transfer to inappropriate parts. 
The style images contain a range of simple and more complicated textures. In both cases, our method achieves effective results, and preserves the content of the images. \cite{li2016combining} can also achieve interesting results, but only for simple texture images. 
For some examples such as figure \ref{fig:style_woman_02} (b) and (c) content woman image, the CNNMRF method achieves interesting results as well, but our method can achieve better results in specific parts, such as the eyes, nose, mouth and background area.
For style images that contain a mixture of textures -- figure \ref{fig:com_more01} and figure \ref{fig:com_more02} -- the results of ~\cite{li2016combining} have many errors in which styles are misapplied. In figure~\ref{fig:com_more01} last row column 1, our results contain artefacts due to the errors in the content semantic masks. 

More style transfer results for objects like train, car, bus and boat are shown in figures~\ref{fig:com_trains}, \ref{fig:com_cars}, \ref{fig:com_bus}, and~\ref{fig:content_boats}. In these examples, in the mask images the green part shows the background probability mask, and the red part shows the object probability mask. Our method produces better results in all these examples. 

\begin{figure*}
\centering

    \includegraphics[width=0.10\linewidth]{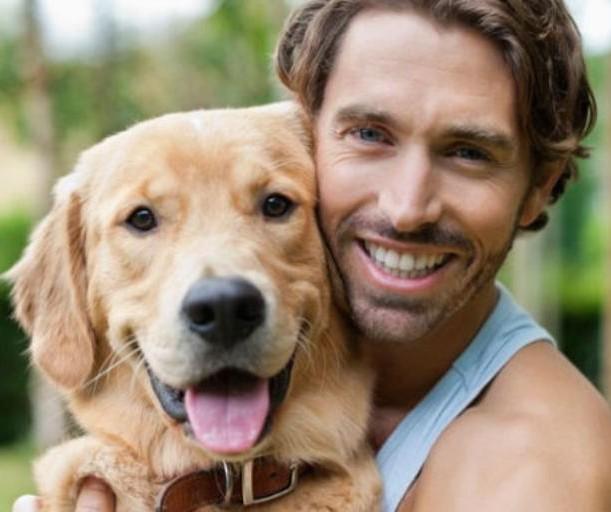}
    \includegraphics[width=0.10\linewidth]{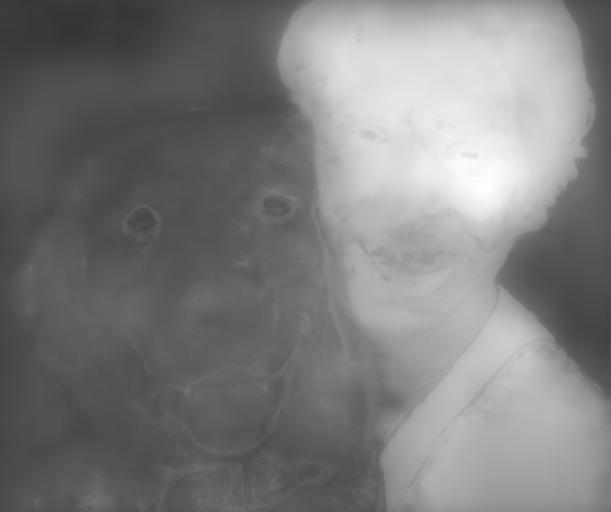}   
    \includegraphics[width=0.10\linewidth]{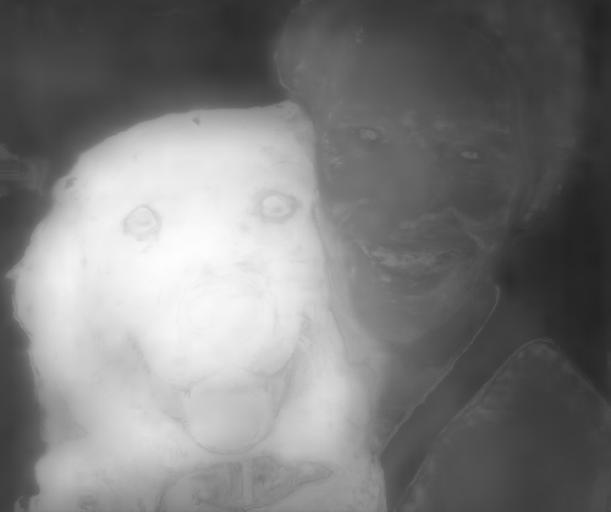}    
    \includegraphics[width=0.10\linewidth]{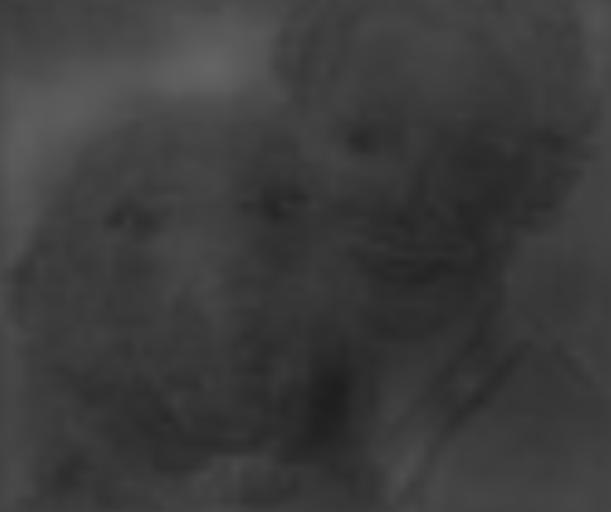} 
    \includegraphics[width=0.10\linewidth]{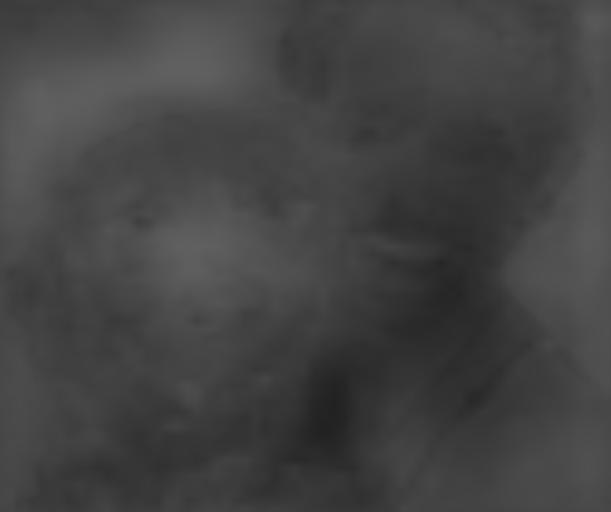}
    \includegraphics[width=0.10\linewidth]{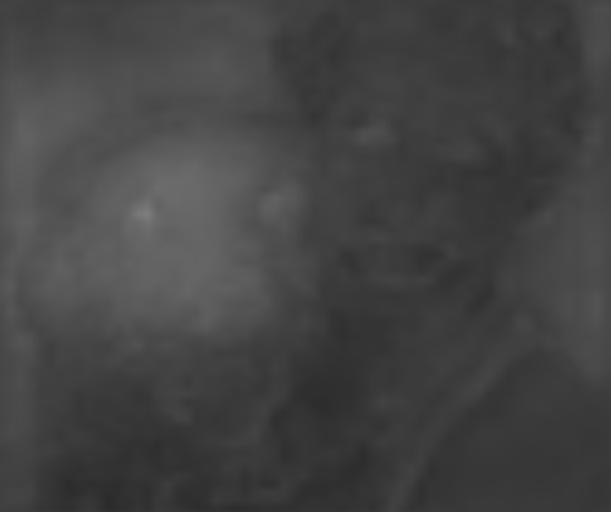}
    \includegraphics[width=0.10\linewidth]{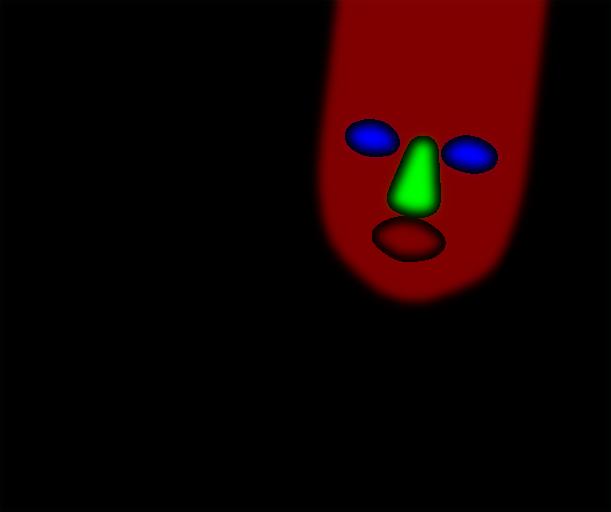} 
    \includegraphics[width=0.10\linewidth]{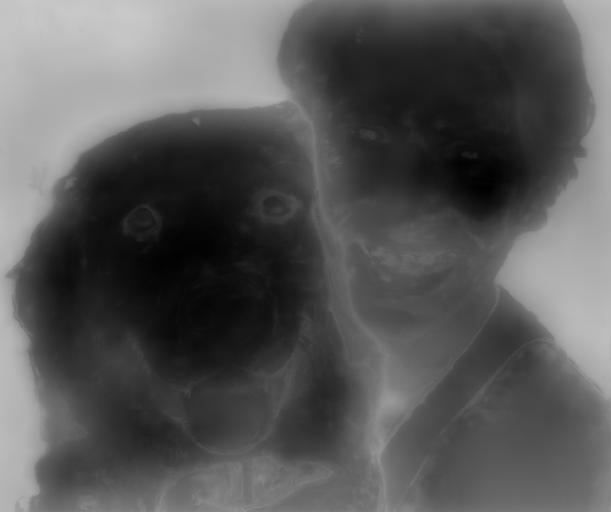} 
    \includegraphics[width=0.10\linewidth]{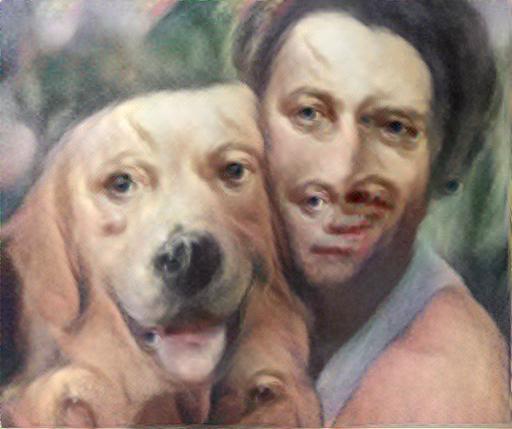}    
    \\
    
    \subfigure[image]{\includegraphics[width=0.10\linewidth]{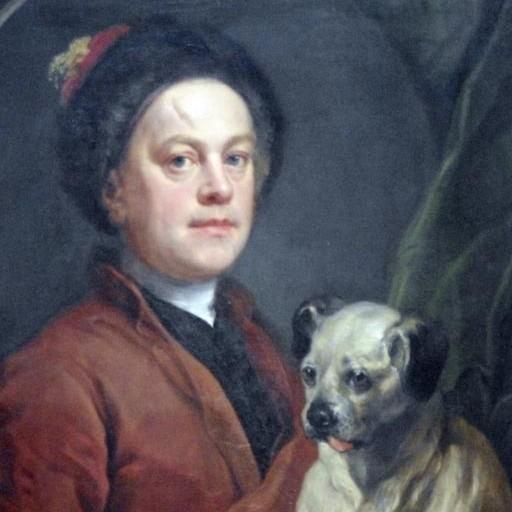}}
    \subfigure[1st mask]{\includegraphics[width=0.10\linewidth]{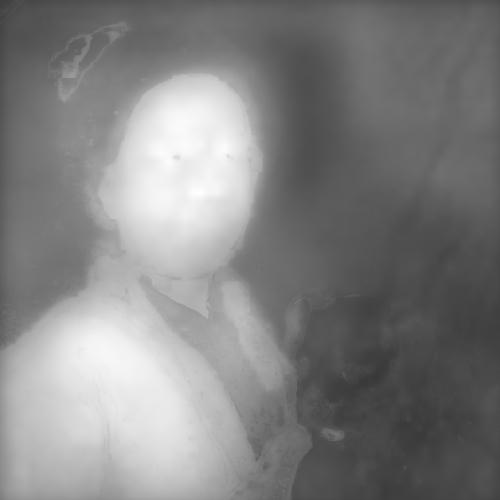}}   
    \subfigure[2nd mask]{\includegraphics[width=0.10\linewidth]{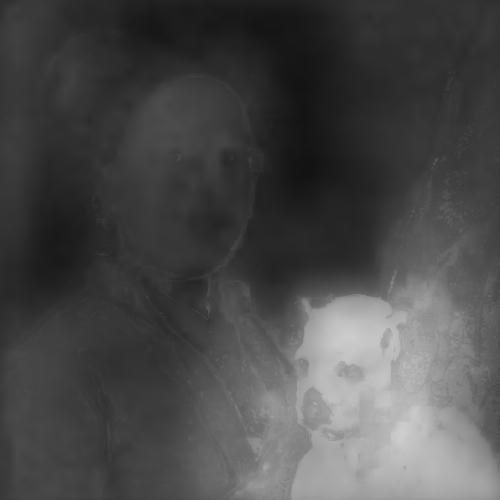}}    
    \subfigure[3rd mask]{\includegraphics[width=0.10\linewidth]{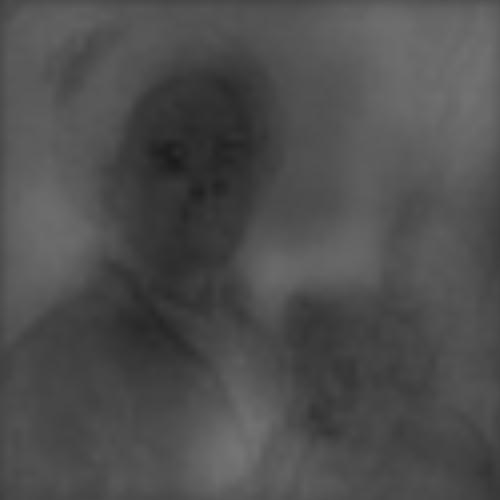}}   
    \subfigure[4th mask]{\includegraphics[width=0.10\linewidth]{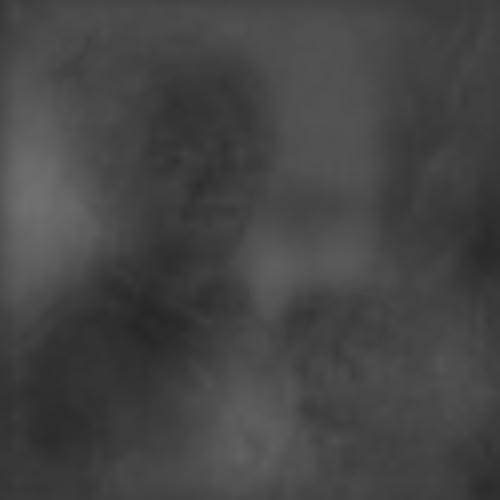}}
    \subfigure[5th mask]{\includegraphics[width=0.10\linewidth]{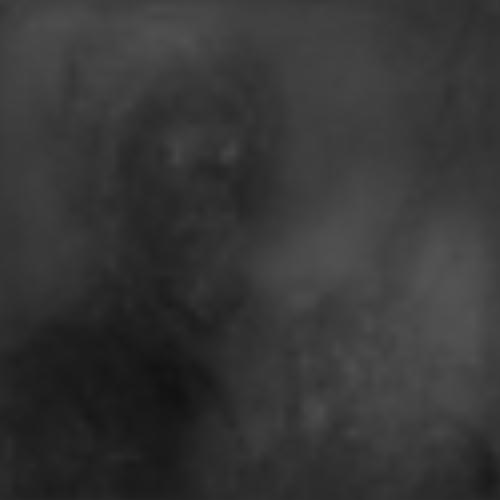}}
    \subfigure[head masks]{\includegraphics[width=0.10\linewidth]{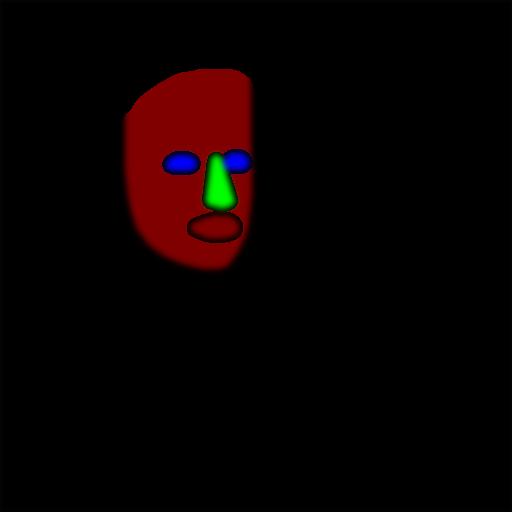}} 
    \subfigure[background]{\includegraphics[width=0.10\linewidth]{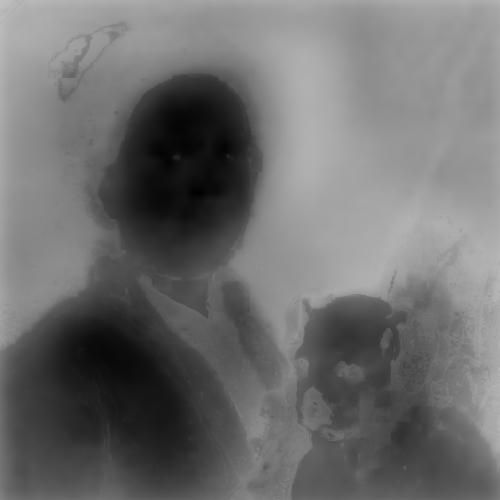}}  
    \subfigure[result]{\includegraphics[width=0.10\linewidth]{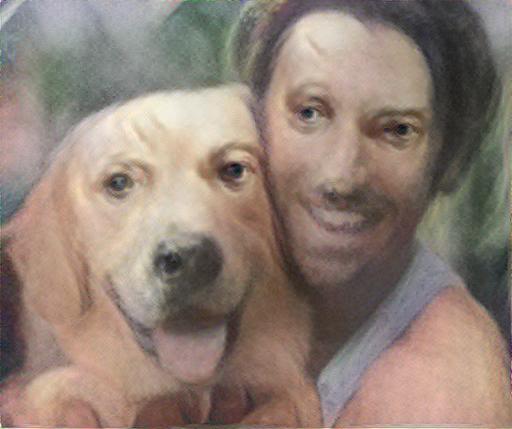}}\\   
\caption{Object style transfer with automatic probability map selection. (a) content and style images, (b)--(f) the automatically selected top 5 semantic masks, (g) head masks, (h) background mask, (i) top: CNNMRF result, bottom: our result.}
\label{fig:man_dog} 
\end{figure*}

\textbf{Automatic multi probability maps selection.}
Not only will probability maps provide a richer feature vector that will benefit the style transfer, but avoiding the need for thresholding or winner-take-all selection has the potential to improve robustness. 
Figure \ref{fig:man_dog} shows an example in which our automatic semantic mask selection effectively chooses relevant object types (person and dog). It demonstrates style transfer using our method when multiple object categories are present. Note that even though the irrelevant 3rd -- 5th masks contain very little response, it is not a problem to include them. 

\begin{figure}
\centering
  \subfigure[]{\includegraphics[width=2.3cm, height=1.6cm ]{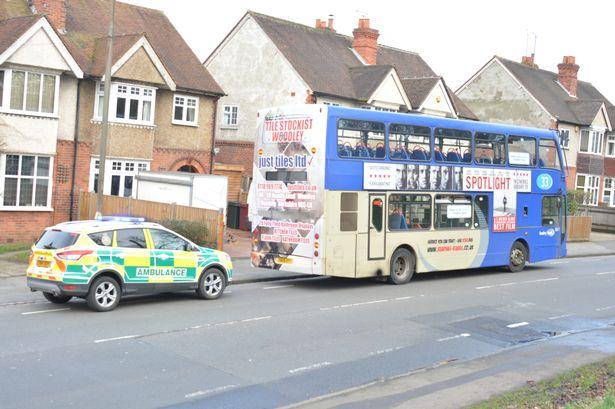}}
  \subfigure[]{\includegraphics[width=1.8cm, height=1.6cm ]{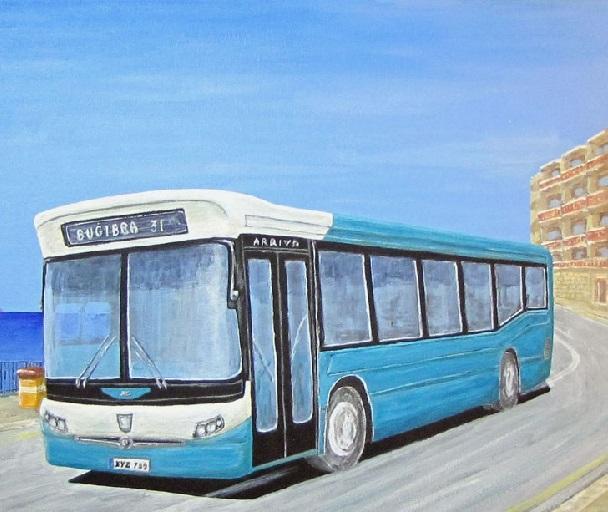}} 
  \subfigure[]{\includegraphics[width=1.8cm, height=1.6cm ]{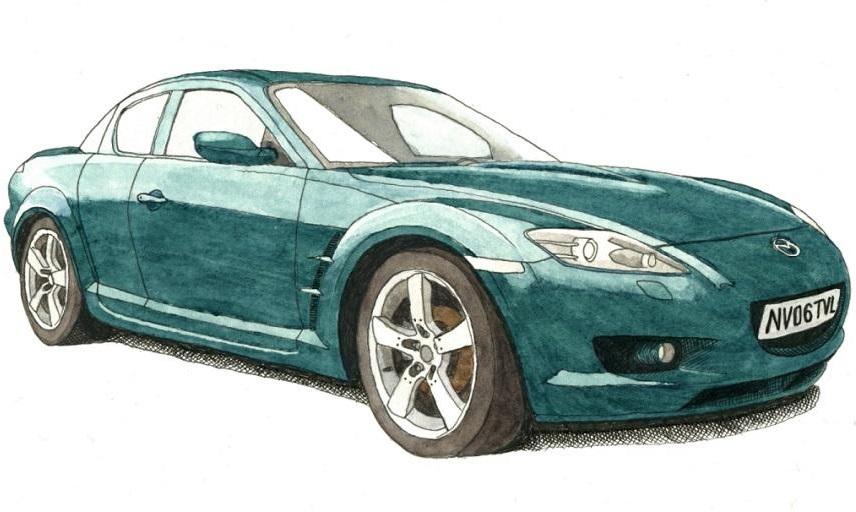}} 
  \subfigure[]{\includegraphics[width=2.3cm, height=1.6cm ]{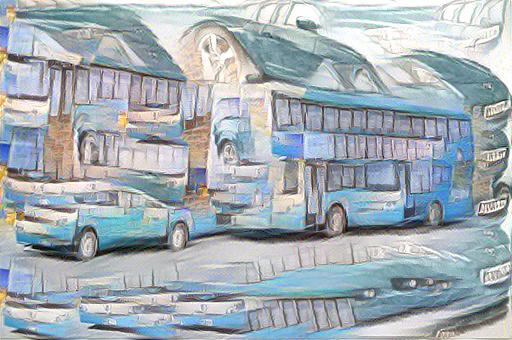}} \\
    \subfigure[]{\includegraphics[width=2.3cm, height=1.6cm ]{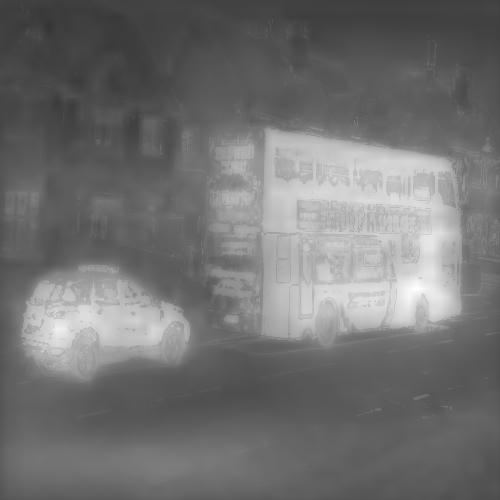}}
    \subfigure[]{\includegraphics[width=1.8cm, height=1.6cm ]{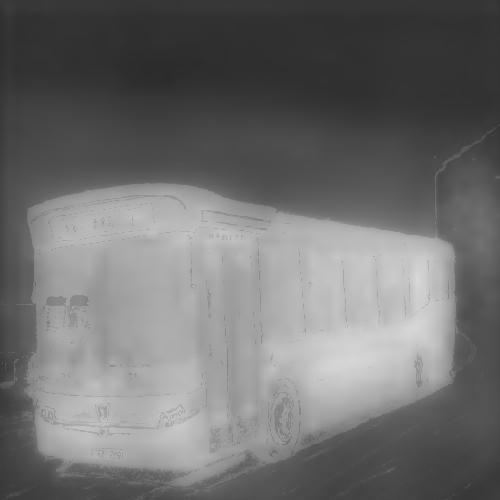}}
    \subfigure[]{\includegraphics[width=1.8cm, height=1.6cm ]{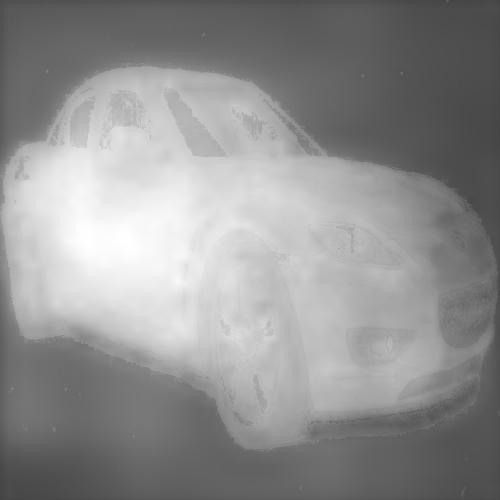}}
    \subfigure[]{\includegraphics[width=2.3cm, height=1.6cm ]{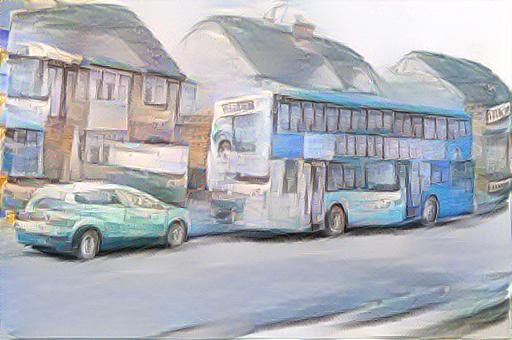}}
\caption{More style images transfer with soft masks, (a) content, (b--c) multiple styles, (e--g) probability maps, (d) CNNMRF result and (h) our result.}  
\label{fig:bus_car} 
\end{figure}

\begin{figure}
\centering
  \subfigure[]{\includegraphics[width=2.6cm, height=1.6cm ]{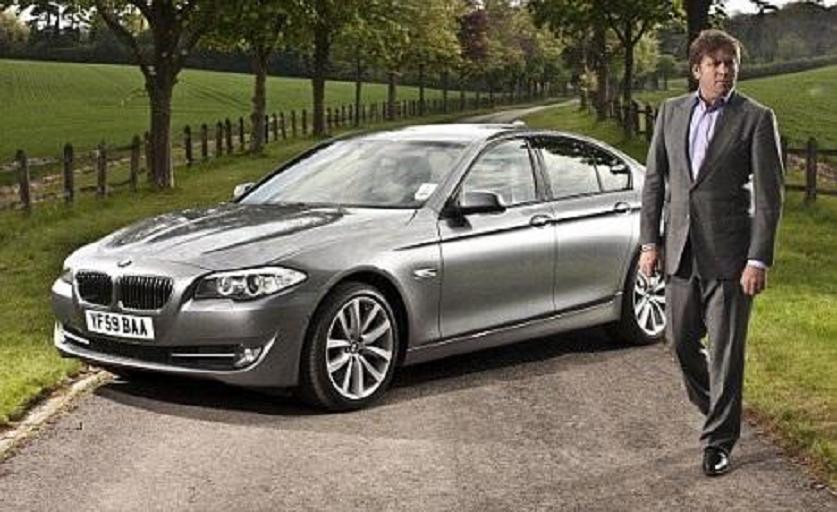}}
  \subfigure[]{\includegraphics[width=1.1cm, height=1.6cm ]{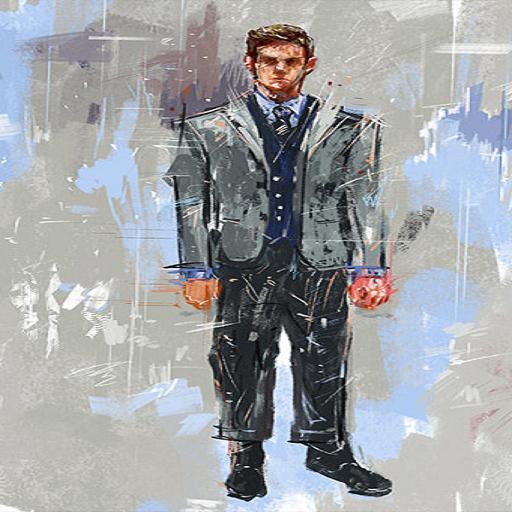}} 
  \subfigure[]{\includegraphics[width=1.6cm, height=1.6cm ]{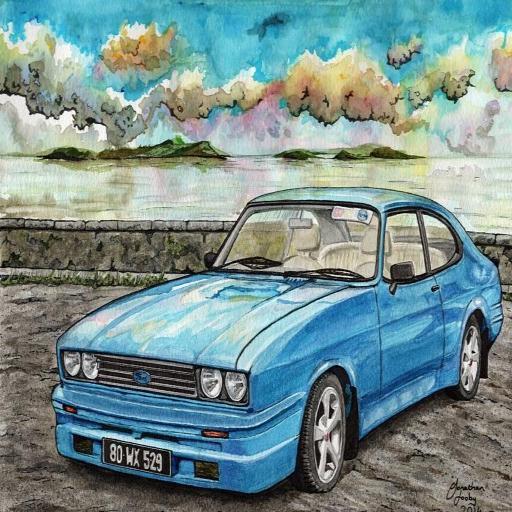}} 
  \subfigure[]{\includegraphics[width=2.6cm, height=1.6cm ]{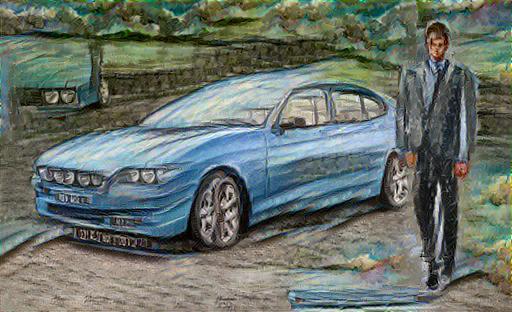}} \\
    \subfigure[]{\includegraphics[width=2.6cm, height=1.6cm ]{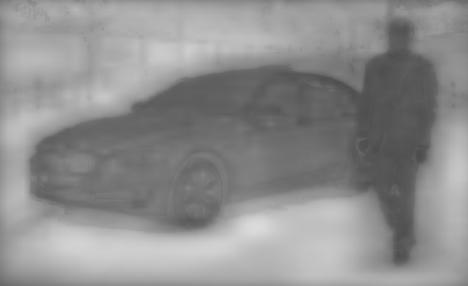}}
    \subfigure[]{\includegraphics[width=1.1cm, height=1.6cm ]{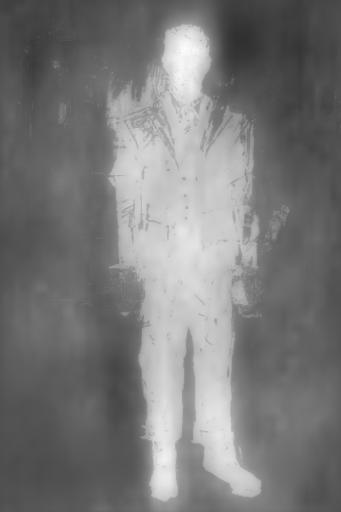}}   
    \subfigure[]{\includegraphics[width=1.6cm, height=1.6cm ]{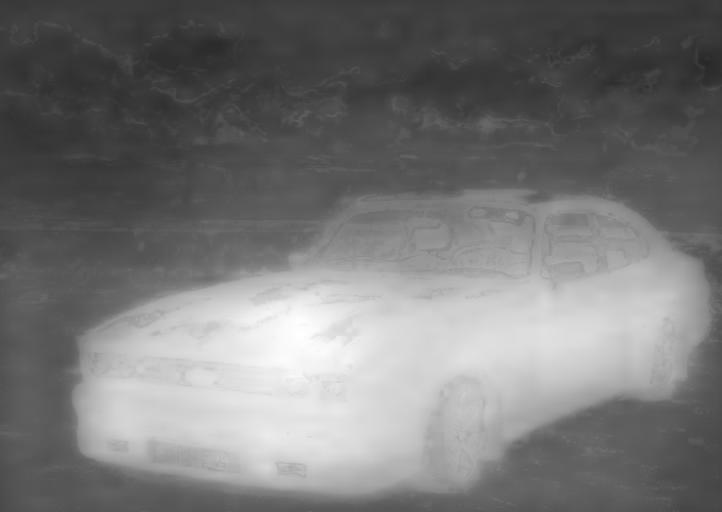}}  
    \subfigure[]{\includegraphics[width=2.6cm, height=1.6cm ]{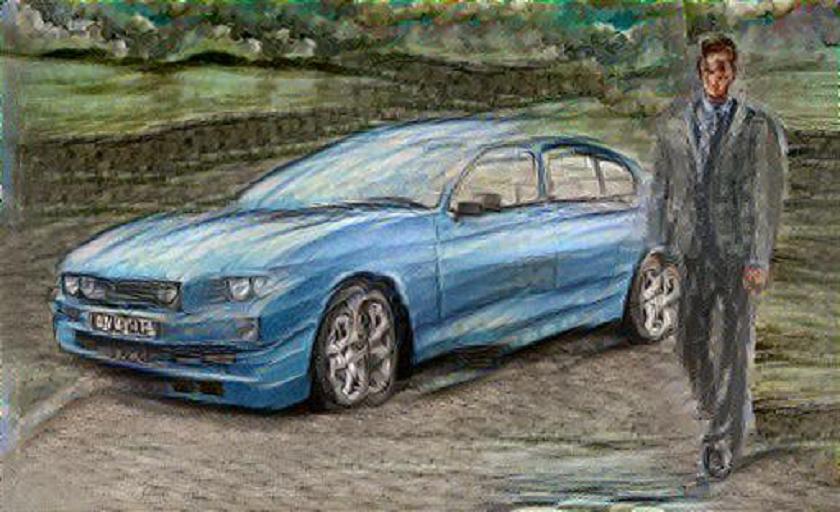}}  
\caption{More style images transfer with soft masks, (a) content, (b--c) multiple styles, (e--g) probability maps, (d) CNNMRF result and (h) our result.}  
\label{fig:man_car} 
\end{figure}

\textbf{Multiple style images.} Our method also allows styles to be transfered from multiple style images to a single content image. In this case, the semantic masks are essential to direct the method to choose suitable patches. Some interesting style transfer results are shown in figures~\ref{fig:bus_car} and~\ref{fig:man_car}.


\begin{figure*}
\centering
 \subfigure[]{\includegraphics[width=0.139 \linewidth]{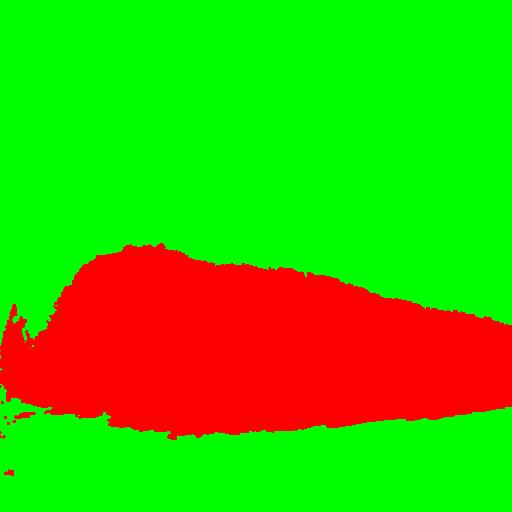}}
 \subfigure[]{\includegraphics[width=0.139 \linewidth]{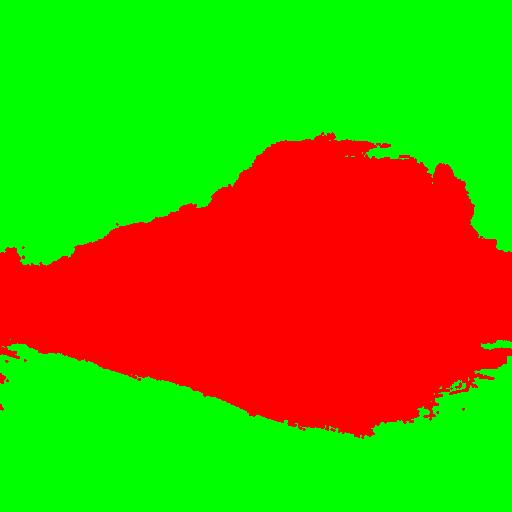}}
 \subfigure[]{\includegraphics[width=0.139 \linewidth]{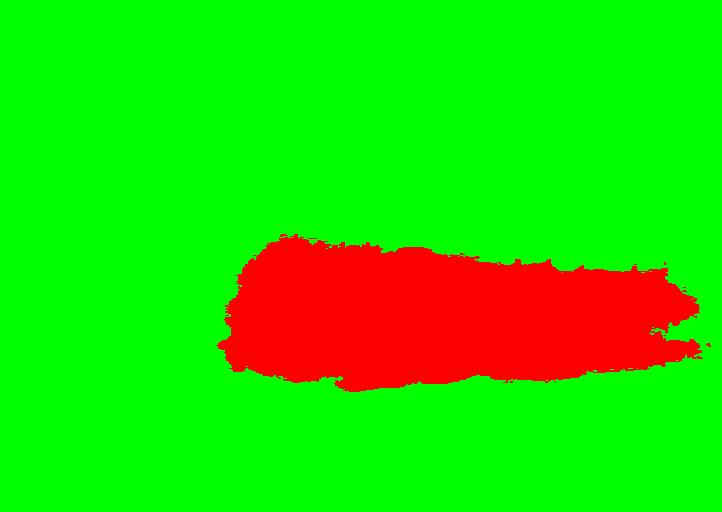}}
 \subfigure[]{\includegraphics[width=0.139 \linewidth]{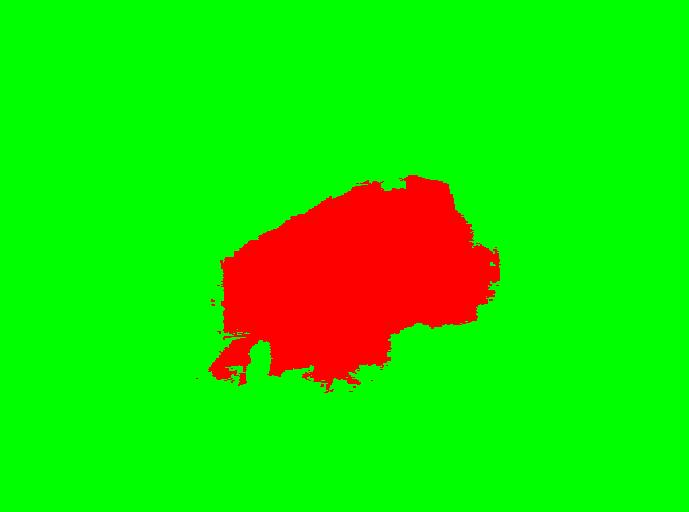}}
 \subfigure[]{\includegraphics[width=0.139 \linewidth]{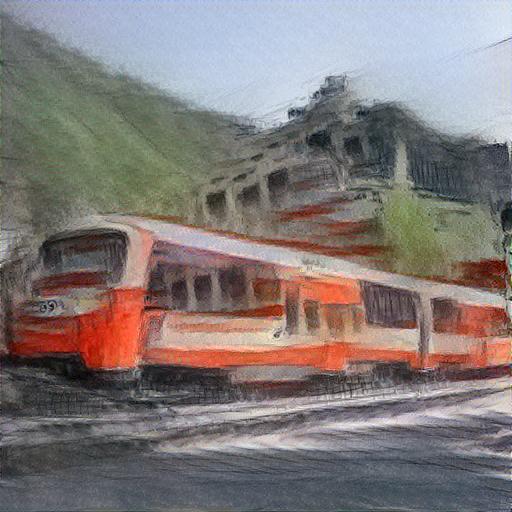}} 
 \subfigure[]{\includegraphics[width=0.139 \linewidth]{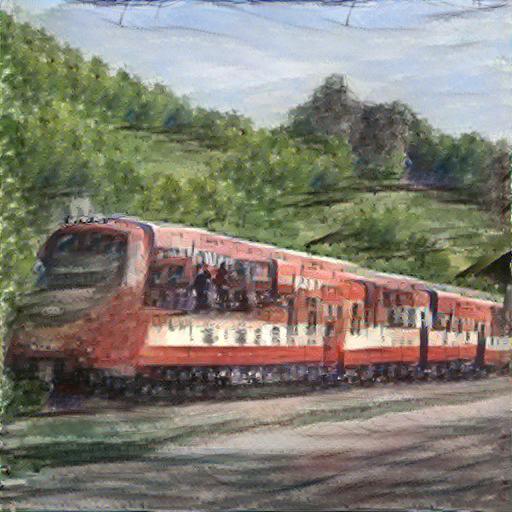}}   
 \subfigure[]{\includegraphics[width=0.139 \linewidth]{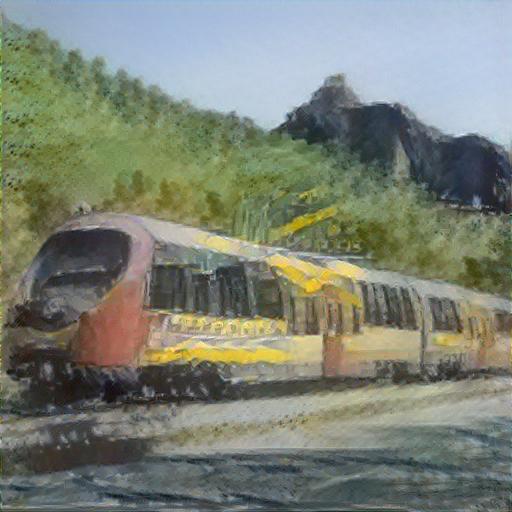}}  
    
\caption{Result based on binary masks, (a) content binary mask, (b--d) binary masks for three styles, (e--g) results }  
\label{fig:soft_com_binary} 
\end{figure*}

\textbf{Comparison of soft masks and binary masks.} We compare our method using soft masks with alternative binary masks. The results are shown in figure~\ref{fig:soft_com_binary}. In comparison, the results with the soft masks (the 2nd column in figure~\ref{fig:com_trains}) not only avoid choosing thresholds but also are visually better since more information is preserved.  


\begin{figure}
\centering
  \includegraphics[width=0.23 \linewidth]{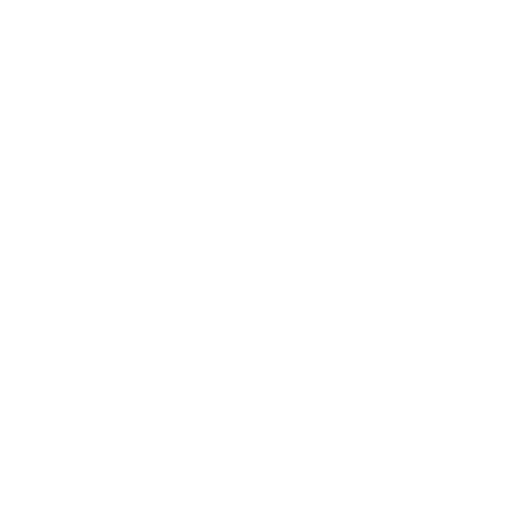} 
  \includegraphics[width=0.23 \linewidth]{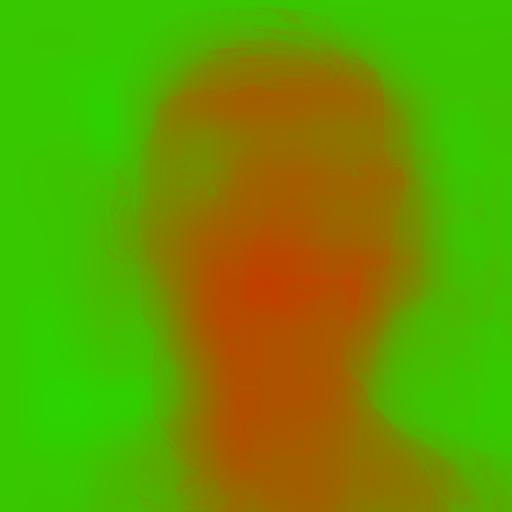} 
  \includegraphics[width=0.23 \linewidth]{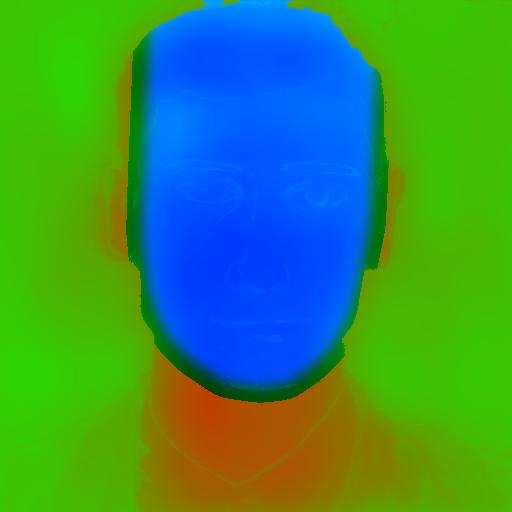}   
  \includegraphics[width=0.23 \linewidth]{style_man_09_pb_06sem_v5_masks_all.jpg}\\   

  \includegraphics[width=0.23 \linewidth]{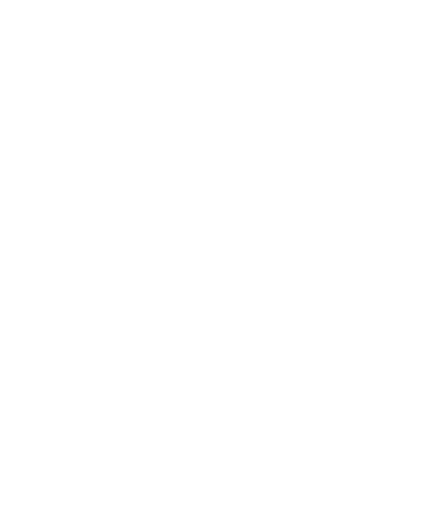} 
  \includegraphics[width=0.23 \linewidth]{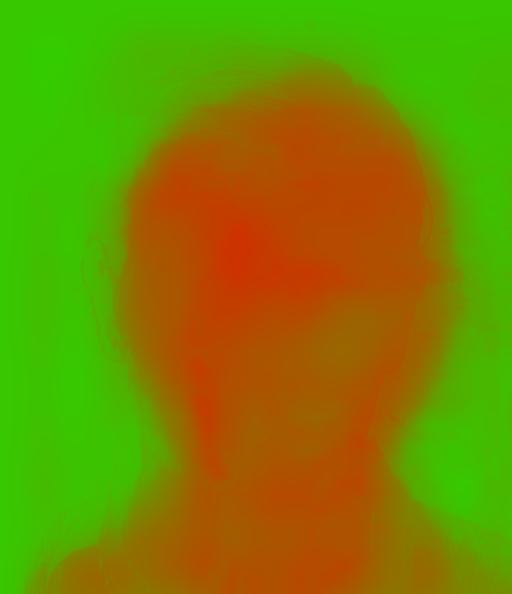} 
  \includegraphics[width=0.23 \linewidth]{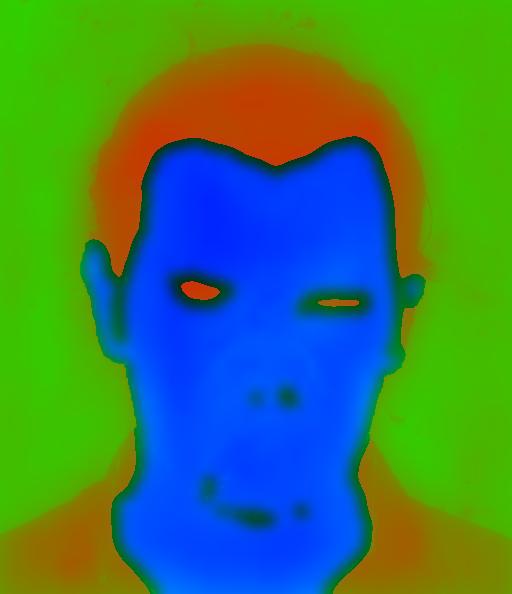} 
  \includegraphics[width=0.23 \linewidth]{man_12_pb_06sem_v5_masks_all.jpg}\\

\subfigure[no mask]{\includegraphics[width=0.23 \linewidth]{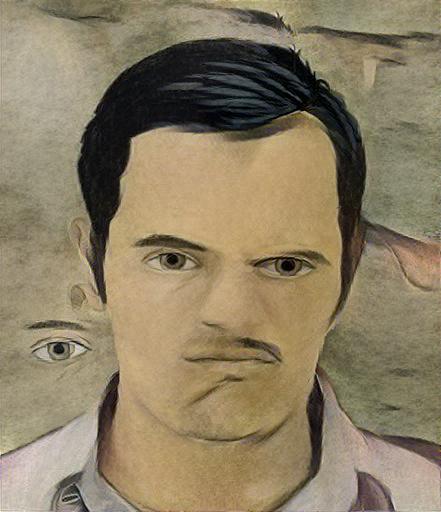}}
\subfigure[2 masks]{\includegraphics[width=0.23 \linewidth]{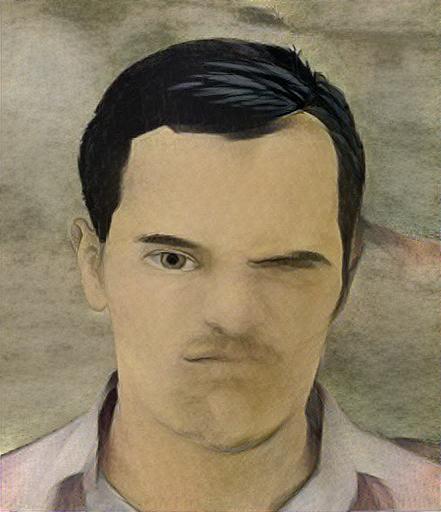}}
\subfigure[3 masks]{\includegraphics[width=0.23 \linewidth]{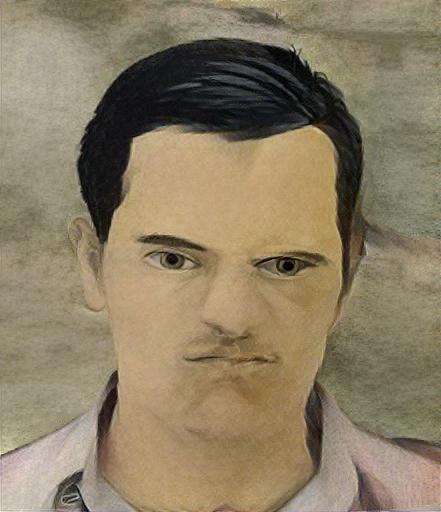} }
\subfigure[6 masks]{\includegraphics[width=0.23 \linewidth]{man_12_TO_style_man_09_pb_hh.jpg}}      
\caption{Result showing the effects on style transfer with an increasing number of labelled objects in the soft masks.}  
\label{fig:add_masks} 
\end{figure}

\textbf{Modifying the number of masks.}
The semantic segmentation significantly affects the style transfer results. For some style images, for example paintings of portraits, it is difficult to automatically segment the face, skin, month, eyes, etc., and to properly segment the background and foreground. Failures in the segmentation will cause some background texture to be embedded into the foreground elements in the synthesised image, thereby generating bad content, such as the jewellery in figure~\ref{fig:style_woman_02}(f) and our result in figure~\ref{fig:com_more01} row 4. The jewellery should be around the neck and not in the background. If the accuracy and reliability of the semantic segmentation can be improved this will lead to better style transfer results. Figure~\ref{fig:add_masks} shows an experiment in which the number of labels in the semantic masks is increased, and demonstrates the importance of separately labelling all the major components of the face.

\begin{figure}
\centering
    \subfigure[$\beta=0$]{\includegraphics[width=0.25\linewidth]{man_12_0_4_20_TO_style_man_09_res_3_100_hh.jpg}}
    \subfigure[$\beta=10$]{\includegraphics[width=0.25\linewidth]{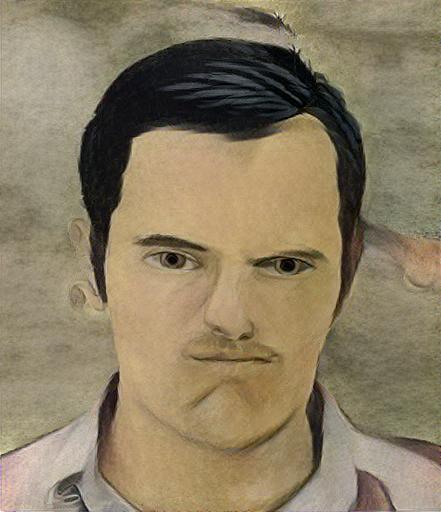}}
    \subfigure[$\beta=20$]{\includegraphics[width=0.25\linewidth]{man_12_TO_style_man_09_pb_hh.jpg}}\\
    \subfigure[$\beta=30$]{\includegraphics[width=0.25\linewidth]{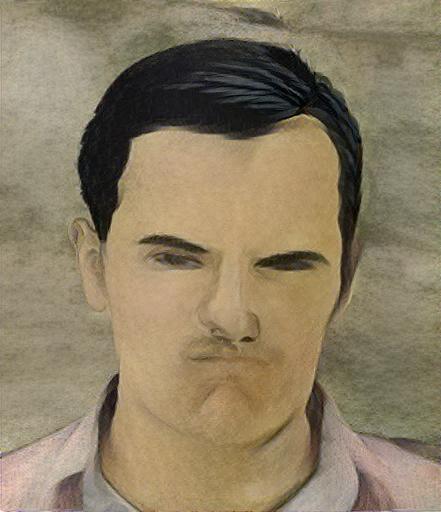}}   
    \subfigure[$\beta=40$]{\includegraphics[width=0.25\linewidth]{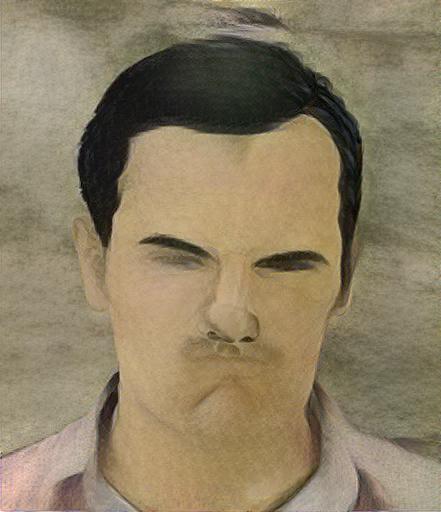}}
    \subfigure[$\beta=80$]{\includegraphics[width=0.25\linewidth]{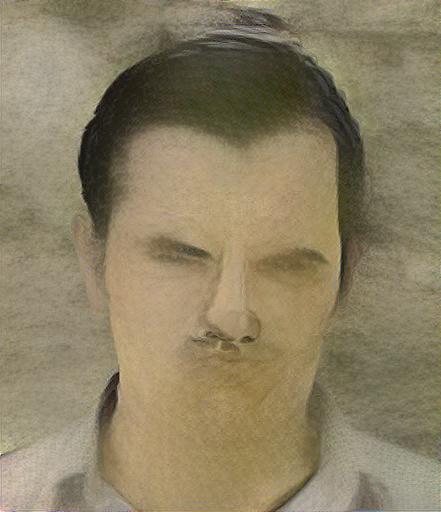}}
\caption{Result showing the effects of varying parameter $\beta$.}  
\label{fig:tune_values} 
\end{figure}

\textbf{Modifying the soft mask weight.}
There are three parameters in our style transfer model, $\alpha_1$, $ \alpha_2$ and $ \beta$ which are the weights for the style, content and semantic mask loss terms. Since the effect of $\alpha_1$ and and $\alpha_2$ is considered in~\cite{li2016combining}, we focus on studying the effect of $\beta$.
By default we set the soft mask weight $\beta=20$. 
This value can be adjusted to control the importance of semantic compliance.
Figure \ref{fig:tune_values} demonstrates the effect of modifying $\beta$ using the content image in figure \ref{fig:content_man_2}(b) and style image in figure \ref{fig:Semanticimg}(b), where $\alpha_1=10^{-4}$, $\alpha_2=20$.
When $\beta$ is too small, the result does not have sufficient semantic control and can produce semantically wrong matches. On the other hand, setting $\beta$ too large may result in matched patches having poor content/style consistency. 
According to our experiments, $\beta \in [15, 35]$ achieves best results.

\section{Conclusions}\label{sec:conc}


Semantic masks are very important for improving the style transfer results. They can achieve background texture and object texture style transfer separately, and prevent them from contaminating each other. 
We can also fine tune the weight of the semantic masks to achieve different results. In the future we will carry out more extensive experiments to determine which weights produce the best style transfer results.

In most cases, soft masks can achieve better results than binary masks, especially in uncertain areas. The probability maps show the likelihood of having specific objects in the image, and can help capture elements of the styles for objects in the style image and preserve the structure of the content image. Therefore, they are useful for finding better patches in the style image and improving the style transfer results. 

Our paper demonstrates the benefits of automatic semantic mask extraction by combining state-of-the-art methods for both semantic segmentation and facial features. The correctness and accuracy of the semantic masks are critical. Using soft masks helps mitigate this, but there is certainly scope to improve semantic segmentation, or to develop methods dedicated to generating soft semantic masks. 


\section*{Acknowledgements}
This work was supported by National Natural Science Foundation of China (61503128), Science and Technology Plan Project of Hunan Province (2016TP102), Scientific Research Fund of Hunan Provincial Education Department (14B025,16C0311), and Hunan Provincial Natural Science Foundation of China (2017JJ4001). We also would like to thank NVIDIA for the GPU donation.
\bibliographystyle{eg-alpha-doi}

\bibliography{egbibsample}

\end{document}